\documentclass[10pt,twocolumn,letterpaper]{article}

\usepackage[pagenumbers]{cvpr} 

\usepackage[dvipsnames]{xcolor}

\usepackage{graphicx}
\usepackage{amsmath}
\usepackage{amssymb}
\usepackage{booktabs}
\usepackage{todonotes}
\usepackage{tabularx}
\usepackage{multirow}
\usepackage{bm}
\usepackage{url}
\usepackage{balance}
\usepackage{enumitem}
\usepackage{color}
\usepackage{bbding}
\usepackage{nccmath}
\usepackage{adjustbox}
\usepackage{tikzinclude}
\usepackage{comment}
\usepackage{wrapfig}

\def\VEC#1{{\boldsymbol{#1}}}

\newcommand{\argmin}{\mathop{\rm arg~min}\limits}
\def\kdb{k_\mathrm{db}}
\def\ndb{n_\mathrm{db}}

\renewcommand{\arraystretch}{0.9}

\definecolor{cvprblue}{rgb}{0.21,0.49,0.74}
\usepackage[pagebackref,breaklinks,colorlinks,allcolors=cvprblue]{hyperref}

\def\paperID{XXXX} 
\def\confName{CVPR}
\def\confYear{2026}

\title{Spatial Polarization Multiplexing:\\
Single-Shot Invisible Shape and Reflectance Recovery}

\author{Tomoki Ichikawa \qquad
Ryo Kawahara \qquad
Ko Nishino \\
Graduate School of Informatics, Kyoto University\\
{\tt\small \url{https://vision.ist.i.kyoto-u.ac.jp/research/spm/}}
}

\begin{document}

\twocolumn[{
    \maketitle 
    \vspace{-2.0em}
    \begin{center}
        \captionsetup{type=figure}
        \begin{tikzpicture}[x=0.001\linewidth,y=0.001\linewidth,every node/.style={inner sep=0pt, text depth=0pt}]
{
    \newcount\vX
    \newcount\vY
    \newcount\vdX
    \newcount\vdXX
    \newcount\vdXObj
    \newcount\vdXhalf
    \newcount\vdY
    \vX = 0
    \vY = 0
    \vdX = 74
    \vdXX = 7
    \vdXObj = 50
    \vdXhalf = 37
    \def\mysz{0.075\linewidth}
    \def\myfont{\scriptsize}

    \draw[->] (-50,0)--(-50,-190);
    \node[inner sep=0pt, font=\myfont, rotate=90] (a) at (-65,-100) {Pushing};
    \draw[->] (450,0)--(450,-190);
    \node[inner sep=0pt, font=\myfont, rotate=90] (a) at (435,-100) {Expanding};

    \advance\vX by \vdXhalf
    \node[inner sep=0pt, font=\myfont] (a) at (\vX,\vY) {Single-shot sensing};
    \advance\vX by \vdX
    \advance\vX by \vdXX
    \advance\vX by \vdX
    \advance\vX by \vdX
    \node[inner sep=0pt, font=\myfont] (a) at (\vX,\vY) {Recovered deforming shape and reflectance};
    \advance\vX by \vdX
    \advance\vX by \vdX
    \advance\vX by \vdX
    \advance\vX by \vdXObj
    \node[inner sep=0pt, font=\myfont] (a) at (\vX,\vY) {Single-shot sensing};
    \advance\vX by \vdX
    \advance\vX by \vdXX
    \advance\vX by \vdX
    \advance\vX by \vdX
    \node[inner sep=0pt, font=\myfont] (a) at (\vX,\vY) {Recovered deforming shape and reflectance};

    \vdY = -40
    \vX = 0
    \advance\vY by \vdY
    \def\myincA#1{\resizebox{\mysz}{!}{\adjincludegraphics[Clip={{0.07\width} {0.19\width} {0.28\width} {0.17\width}}]{#1}}} 
    \def\myincB#1{\resizebox{\mysz}{!}{\adjincludegraphics[Clip={{0.08\width} {0.1\width} {0.14\width} {0.17\width}}]{#1}}} 
    \node[inner sep=0pt] (a) at (\vX,\vY) {\myincA{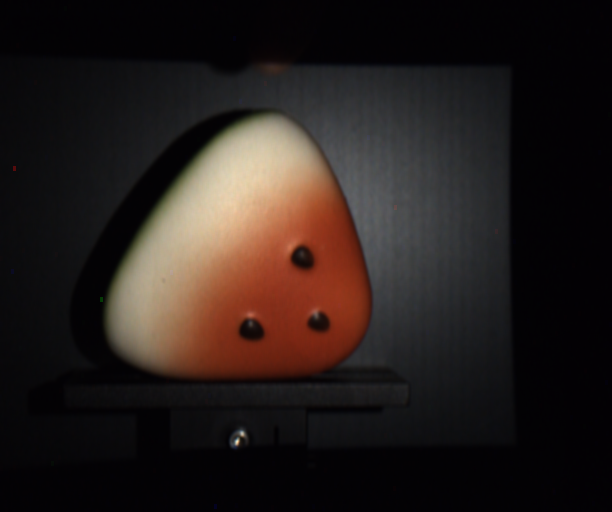}};
    \advance\vX by \vdX
    \node[inner sep=0pt] (a) at (\vX,\vY) {\myincA{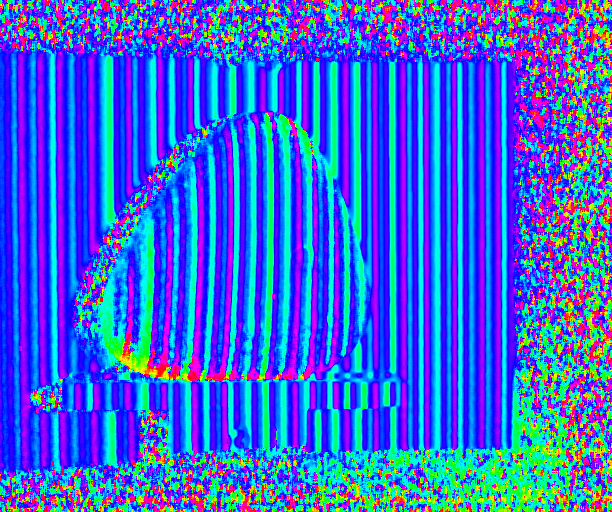}};
    \advance\vX by \vdX
    \advance\vX by \vdXX
    \node[inner sep=0pt] (a) at (\vX,\vY) {\myincA{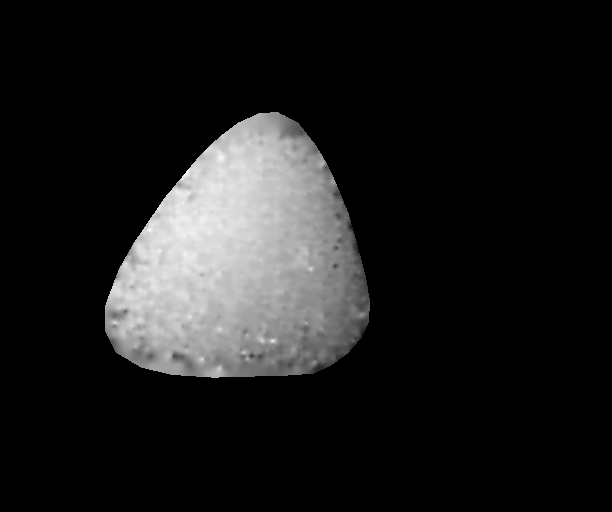}};
    \advance\vX by \vdX
    \node[inner sep=0pt] (a) at (\vX,\vY) {\myincA{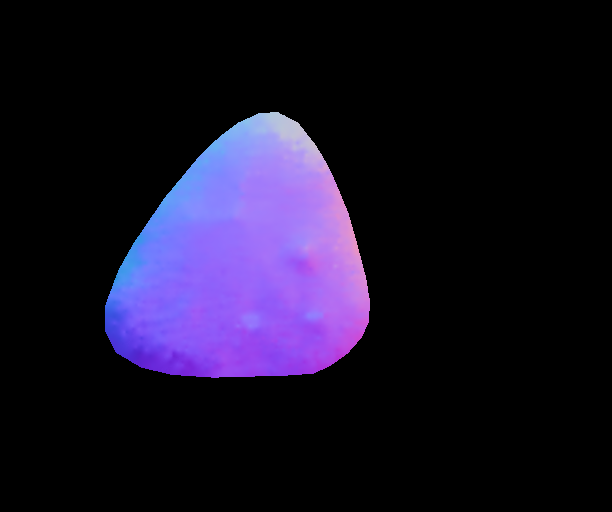}};
    \advance\vX by \vdX
    \node[inner sep=0pt] (a) at (\vX,\vY) {\myincA{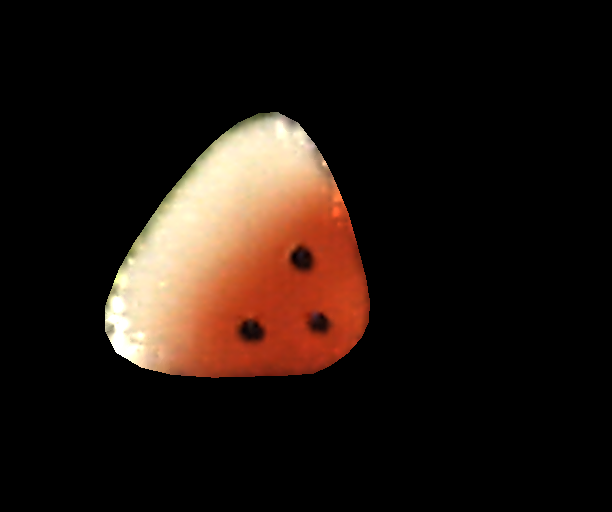}};
    \advance\vX by \vdX
    \node[inner sep=0pt] (a) at (\vX,\vY) {\myincA{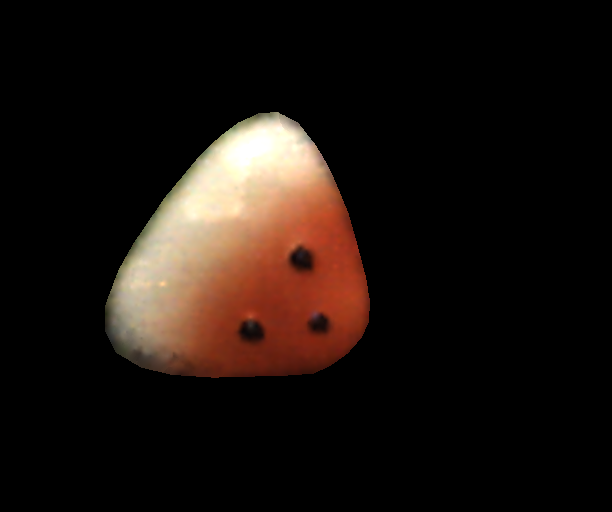}};
    \advance\vX by \vdX
    \advance\vX by \vdXObj
    \node[inner sep=0pt] (a) at (\vX,\vY) {\myincB{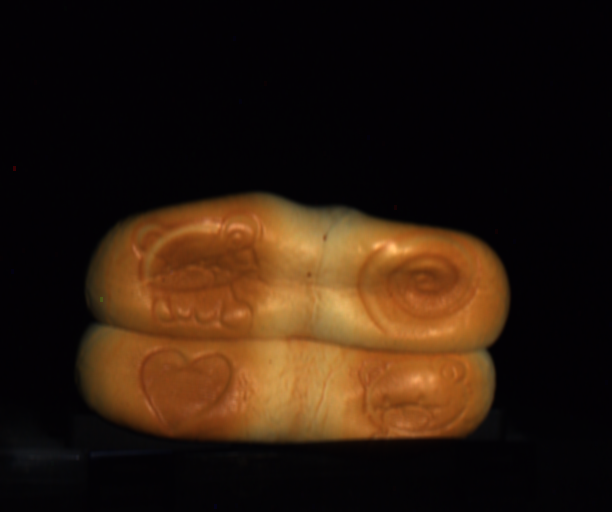}};
    \advance\vX by \vdX
    \node[inner sep=0pt] (a) at (\vX,\vY) {\myincB{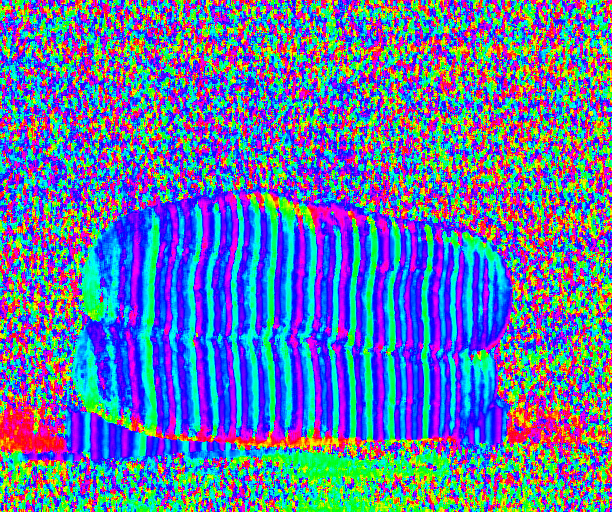}};
    \advance\vX by \vdX
    \advance\vX by \vdXX
    \node[inner sep=0pt] (a) at (\vX,\vY) {\myincB{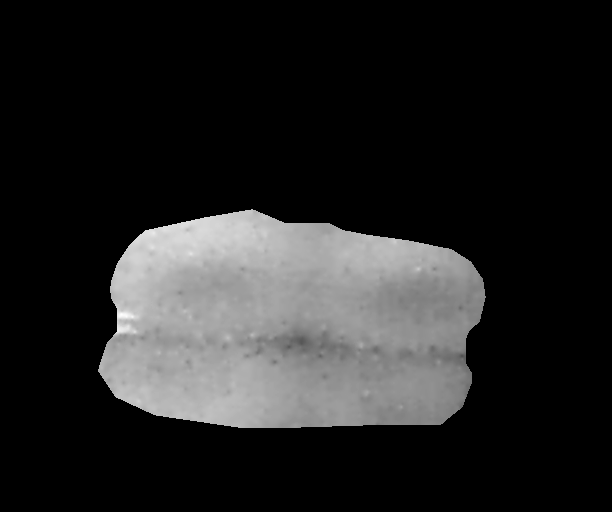}};
    \advance\vX by \vdX
    \node[inner sep=0pt] (a) at (\vX,\vY) {\myincB{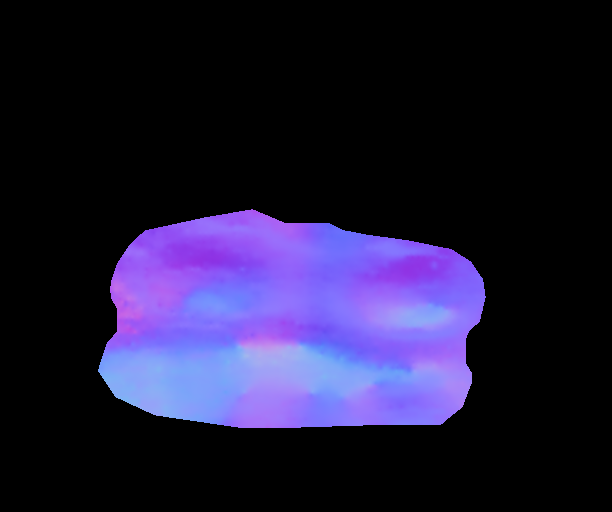}};
    \advance\vX by \vdX
    \node[inner sep=0pt] (a) at (\vX,\vY) {\myincB{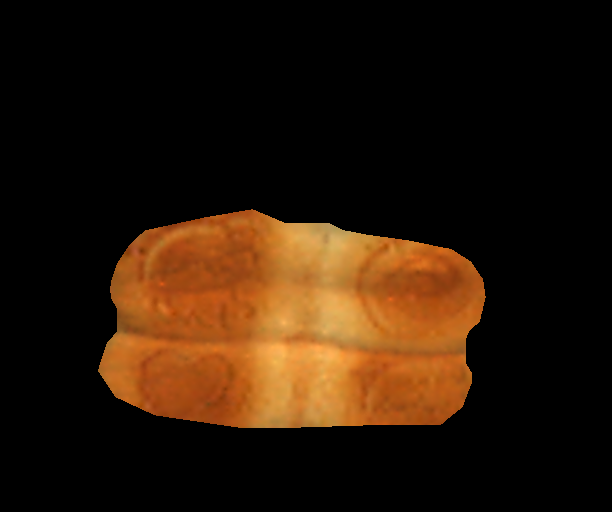}};
    \advance\vX by \vdX
    \node[inner sep=0pt] (a) at (\vX,\vY) {\myincB{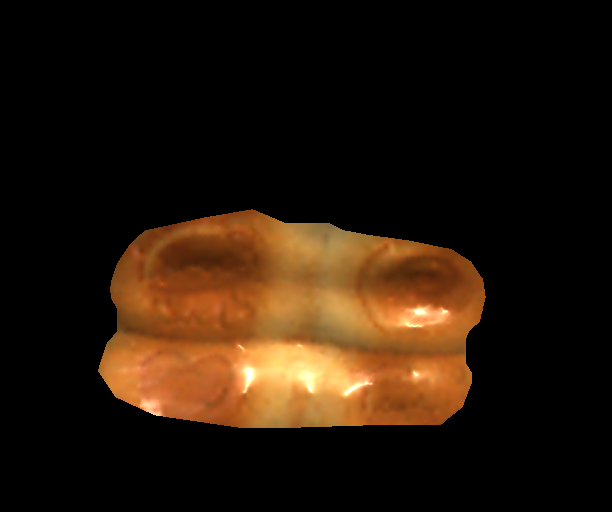}};

    \vdY = -53
    \vX = 0
    \advance\vY by \vdY
    \node[inner sep=0pt] (a) at (\vX,\vY) {\myincA{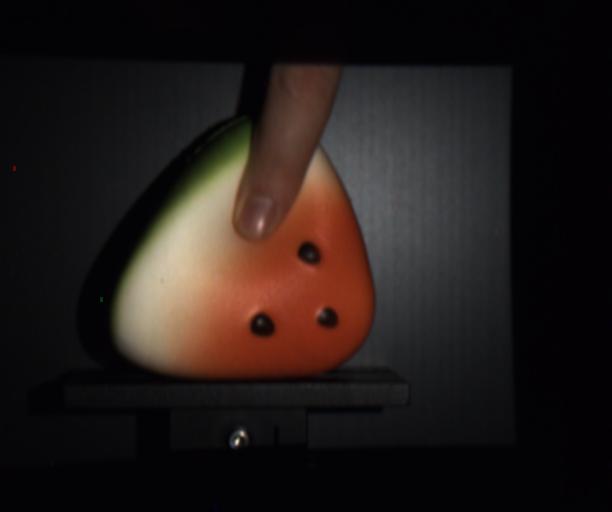}};
    \advance\vX by \vdX
    \node[inner sep=0pt] (a) at (\vX,\vY) {\myincA{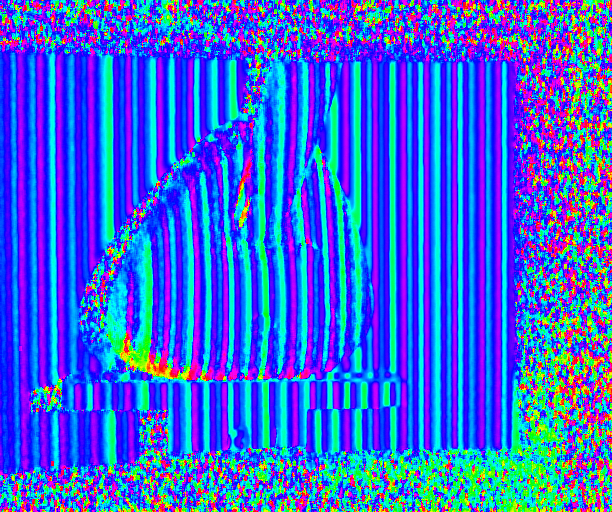}};
    \advance\vX by \vdX
    \advance\vX by \vdXX
    \node[inner sep=0pt] (a) at (\vX,\vY) {\myincA{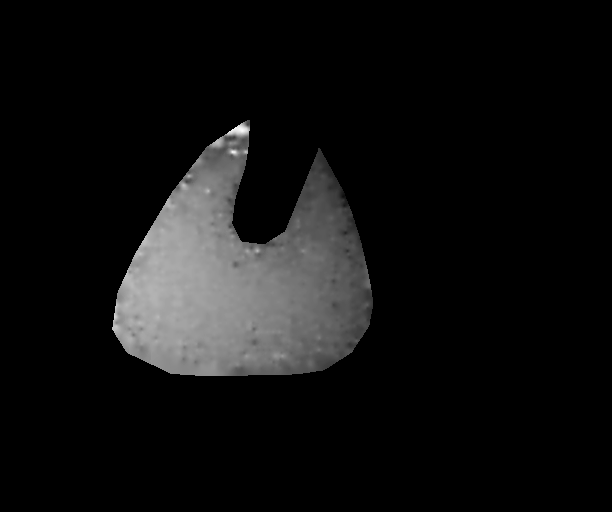}};
    \advance\vX by \vdX
    \node[inner sep=0pt] (a) at (\vX,\vY) {\myincA{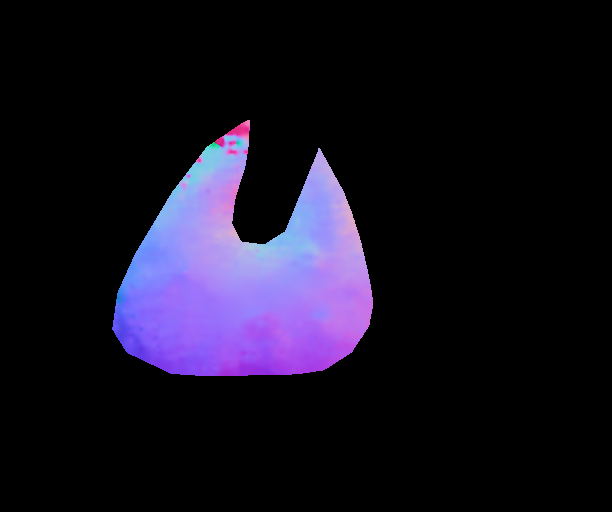}};
    \advance\vX by \vdX
    \node[inner sep=0pt] (a) at (\vX,\vY) {\myincA{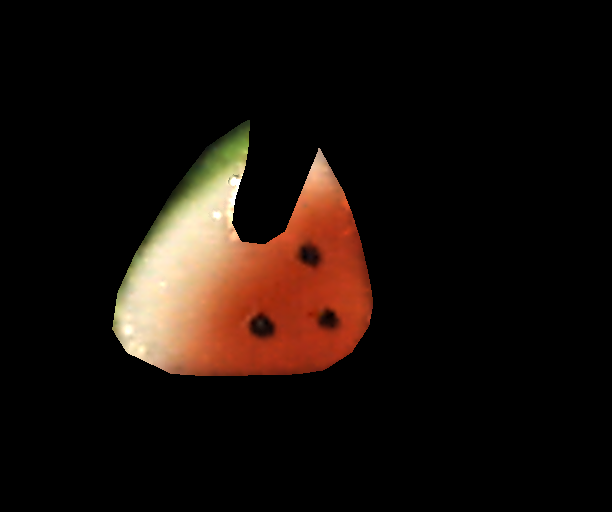}};
    \advance\vX by \vdX
    \node[inner sep=0pt] (a) at (\vX,\vY) {\myincA{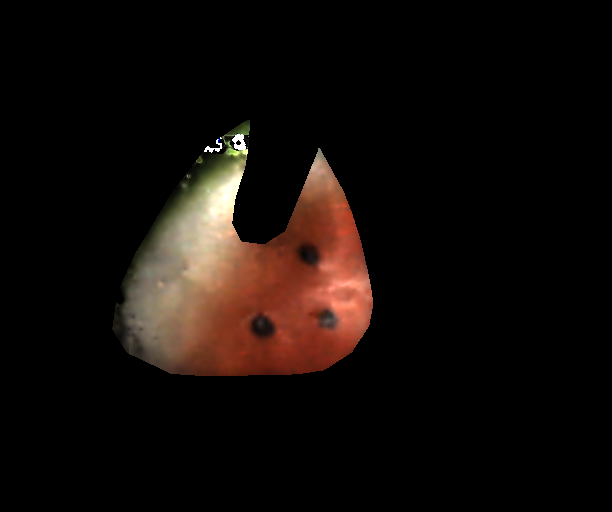}};
    \advance\vX by \vdX
    \advance\vX by \vdXObj
    \node[inner sep=0pt] (a) at (\vX,\vY) {\myincB{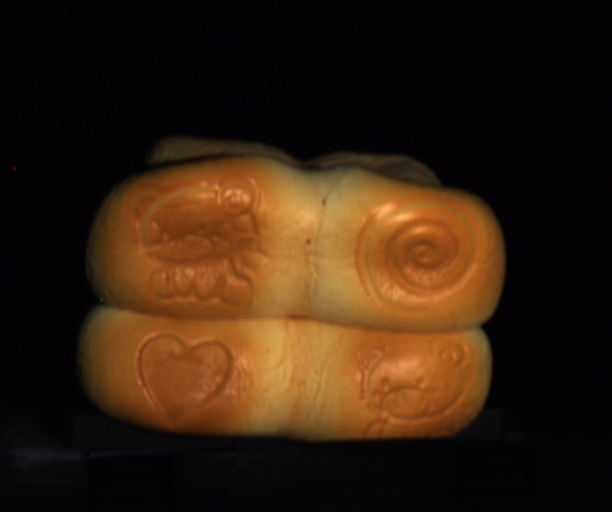}};
    \advance\vX by \vdX
    \node[inner sep=0pt] (a) at (\vX,\vY) {\myincB{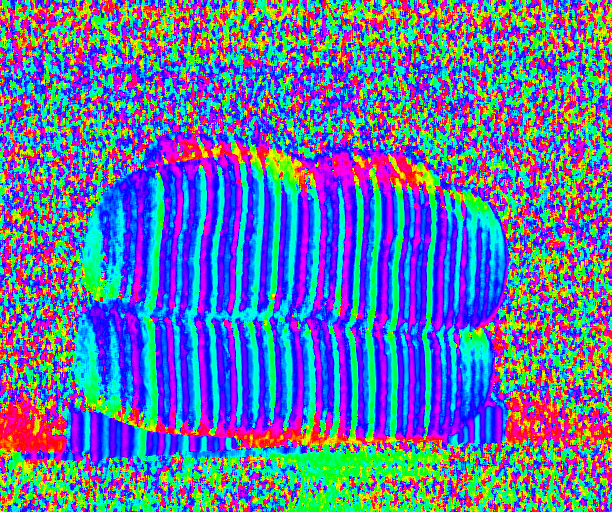}};
    \advance\vX by \vdX
    \advance\vX by \vdXX
    \node[inner sep=0pt] (a) at (\vX,\vY) {\myincB{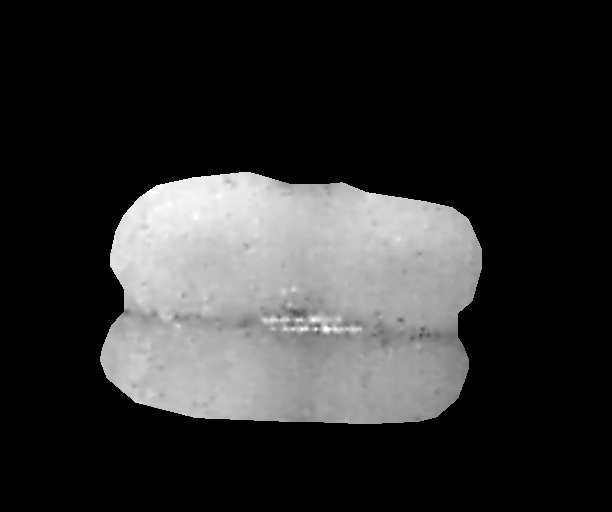}};
    \advance\vX by \vdX
    \node[inner sep=0pt] (a) at (\vX,\vY) {\myincB{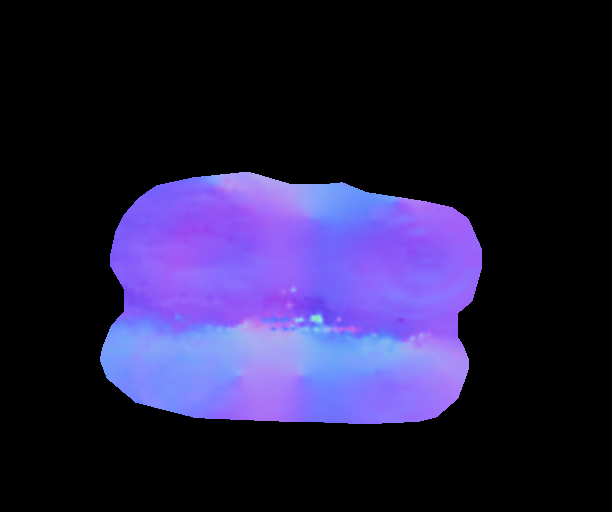}};
    \advance\vX by \vdX
    \node[inner sep=0pt] (a) at (\vX,\vY) {\myincB{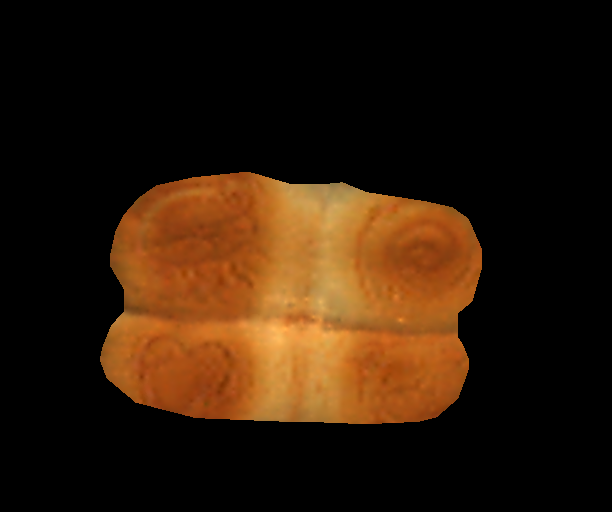}};
    \advance\vX by \vdX
    \node[inner sep=0pt] (a) at (\vX,\vY) {\myincB{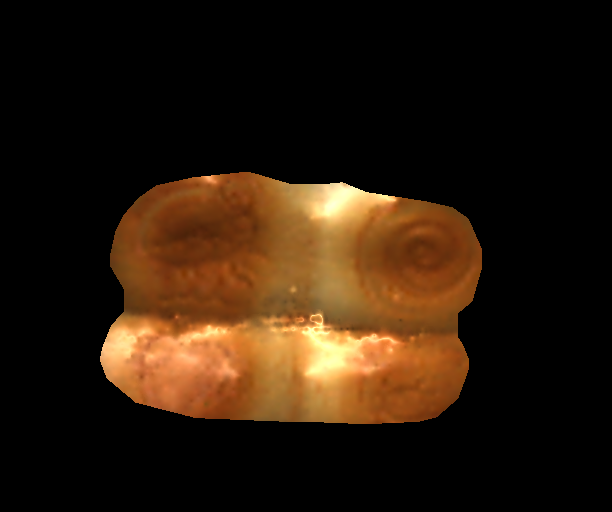}};

    \vdY = -53
    \vX = 0
    \advance\vY by \vdY
    \node[inner sep=0pt] (a) at (\vX,\vY) {\myincA{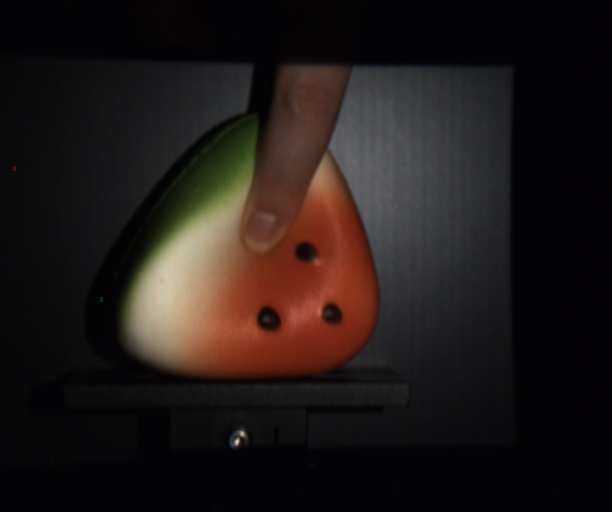}};
    \advance\vX by \vdX
    \node[inner sep=0pt] (a) at (\vX,\vY) {\myincA{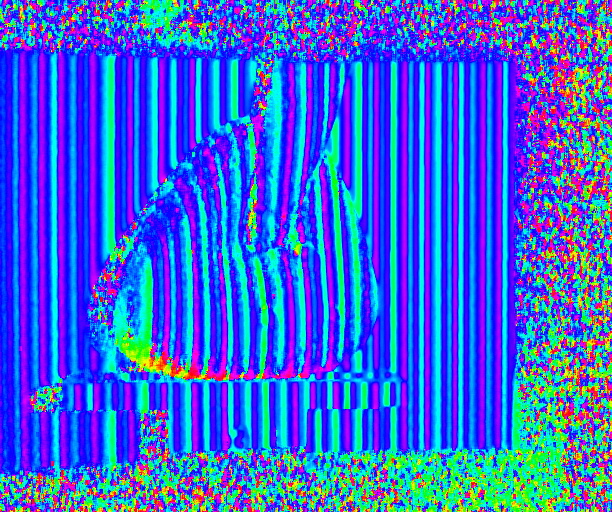}};
    \advance\vX by \vdX
    \advance\vX by \vdXX
    \node[inner sep=0pt] (a) at (\vX,\vY) {\myincA{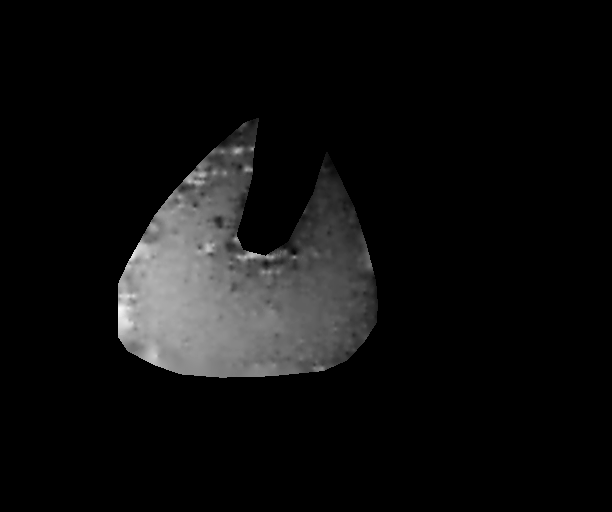}};
    \advance\vX by \vdX
    \node[inner sep=0pt] (a) at (\vX,\vY) {\myincA{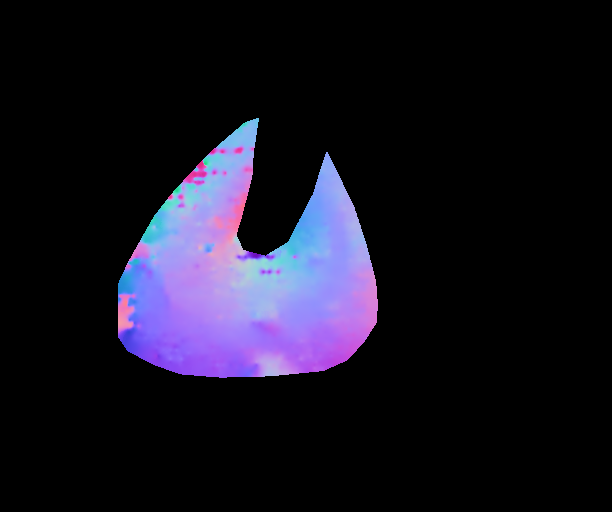}};
    \advance\vX by \vdX
    \node[inner sep=0pt] (a) at (\vX,\vY) {\myincA{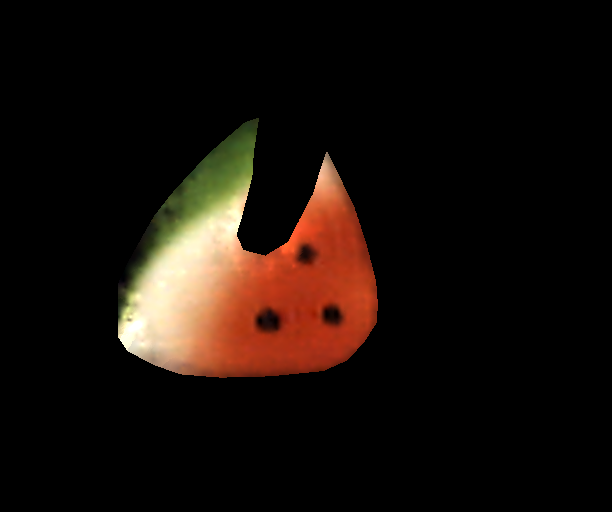}};
    \advance\vX by \vdX
    \node[inner sep=0pt] (a) at (\vX,\vY) {\myincA{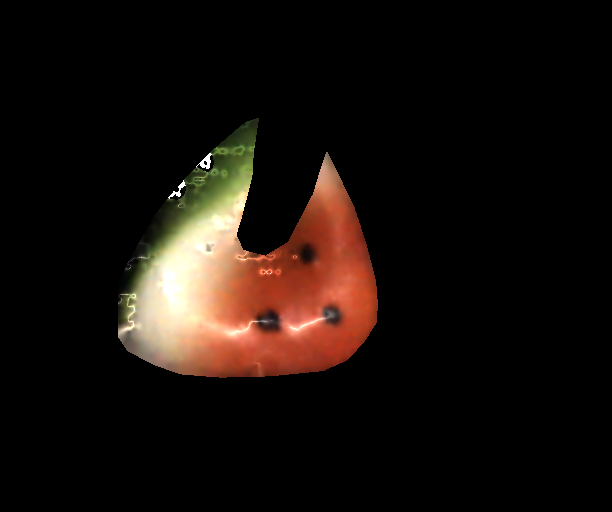}};
    \advance\vX by \vdX
    \advance\vX by \vdXObj
    \node[inner sep=0pt] (a) at (\vX,\vY) {\myincB{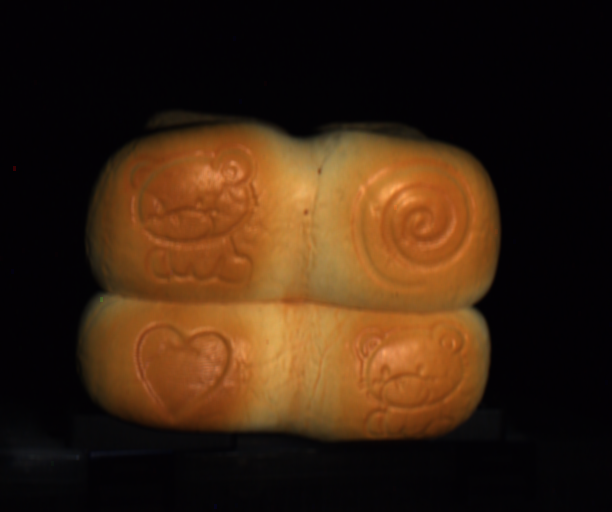}};
    \advance\vX by \vdX
    \node[inner sep=0pt] (a) at (\vX,\vY) {\myincB{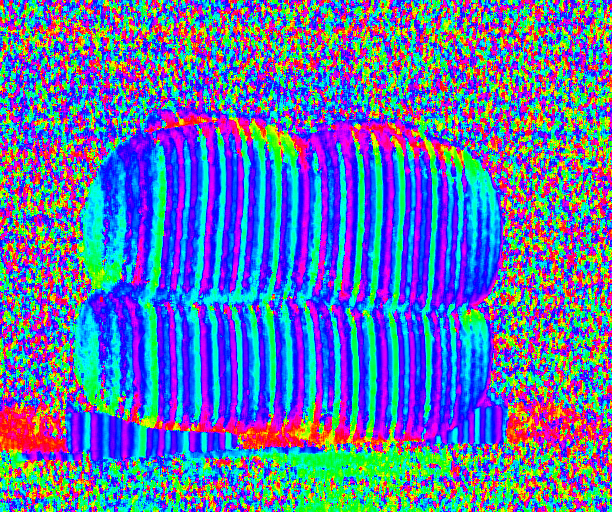}};
    \advance\vX by \vdX
    \advance\vX by \vdXX
    \node[inner sep=0pt] (a) at (\vX,\vY) {\myincB{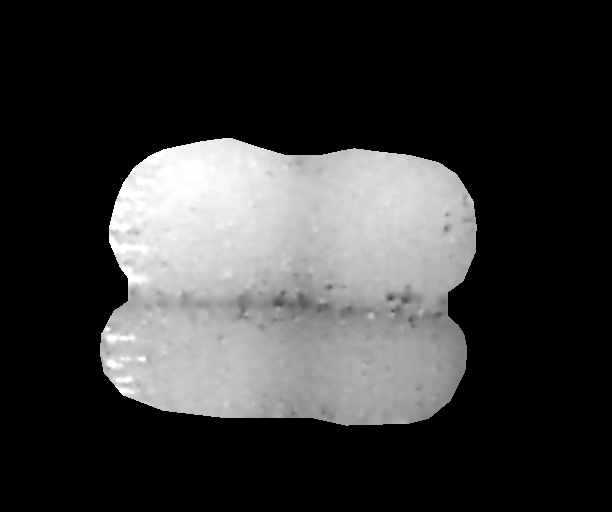}};
    \advance\vX by \vdX
    \node[inner sep=0pt] (a) at (\vX,\vY) {\myincB{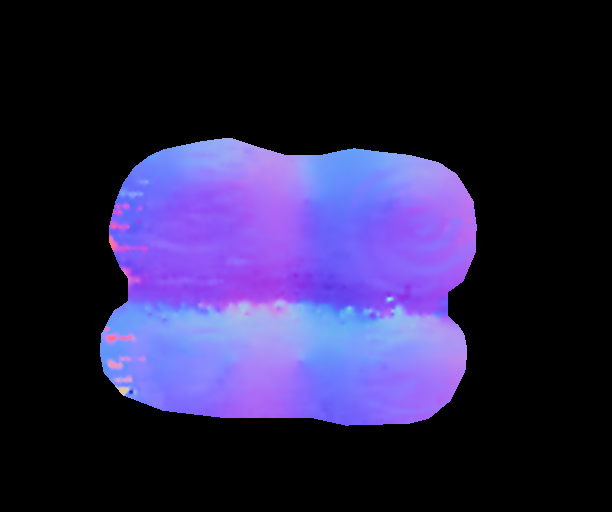}};
    \advance\vX by \vdX
    \node[inner sep=0pt] (a) at (\vX,\vY) {\myincB{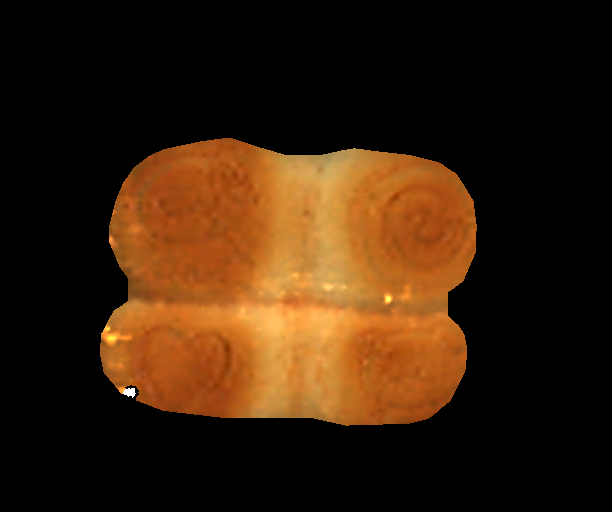}};
    \advance\vX by \vdX
    \node[inner sep=0pt] (a) at (\vX,\vY) {\myincB{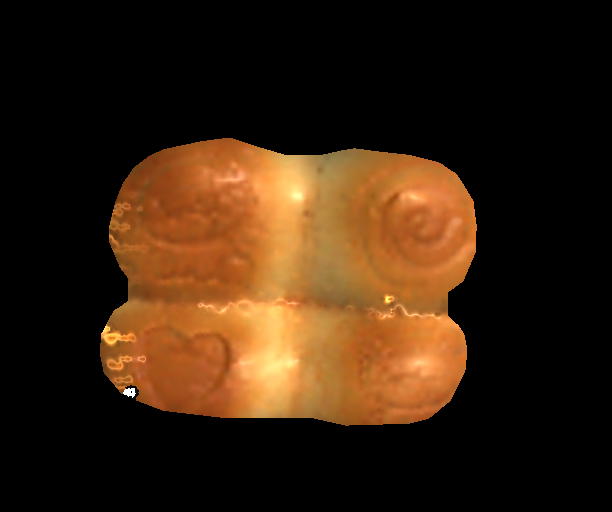}};

    \vdY = -41
    \vX = 0
    \advance\vY by \vdY
    \node[inner sep=0pt, font=\myfont] (a) at (\vX,\vY) {Image};
    \advance\vX by \vdX
    \node[inner sep=0pt, font=\myfont] (a) at (\vX,\vY) {AoLP};
    \advance\vX by \vdX
    \advance\vX by \vdXX
    \node[inner sep=0pt, font=\myfont] (a) at (\vX,\vY) {Depth};
    \advance\vX by \vdX
    \node[inner sep=0pt, font=\myfont] (a) at (\vX,\vY) {Normal};
    \advance\vX by \vdX
    \node[inner sep=0pt, font=\myfont] (a) at (\vX,\vY) {Diffuse albedo};
    \advance\vX by \vdX
    \node[inner sep=0pt, font=\myfont] (a) at (\vX,\vY) {Relighting};
    \advance\vX by \vdX
    \advance\vX by \vdXObj
    \node[inner sep=0pt, font=\myfont] (a) at (\vX,\vY) {Image};
    \advance\vX by \vdX
    \node[inner sep=0pt, font=\myfont] (a) at (\vX,\vY) {AoLP};
    \advance\vX by \vdX
    \advance\vX by \vdXX
    \node[inner sep=0pt, font=\myfont] (a) at (\vX,\vY) {Depth};
    \advance\vX by \vdX
    \node[inner sep=0pt, font=\myfont] (a) at (\vX,\vY) {Normal};
    \advance\vX by \vdX
    \node[inner sep=0pt, font=\myfont] (a) at (\vX,\vY) {Diffuse albedo};
    \advance\vX by \vdX
    \node[inner sep=0pt, font=\myfont] (a) at (\vX,\vY) {Relighting};
}
\end{tikzpicture}
        \vspace{-0.5em}
        
        \captionof{figure}{
            Spatial Polarization Multiplexing (SPM) recovers the shape and reflectance of a target in a single shot with a novel polarimetric structured light pattern that enables simultaneous decoding for shape reconstruction and decomposition of polarimetric diffuse and specular reflections. SPM enables these joint sensing per-frame of deforming objects without altering the appearance (invisible to the naked eyes). Left: a plush toy pushed down with a finger. Right: a soft loaf expanding after compression.
        }
        \vspace{-0.5em}
        \label{fig: opening figure}

    \end{center}
}]

\begin{abstract}

We propose spatial polarization multiplexing (SPM) for joint sensing of shape and reflectance of a static or dynamic deformable object, which is also invisible to the naked eye. Past structured-light methods are limited to shape acquisition and cannot recover reflectance as they alter scene appearance. Our key idea is to spatially multiplex a polarization pattern to encode the incident ray and also densely sample the reflected light. We derive a quantized polarized light pattern that can be robustly and uniquely decoded from the reflected Angle of Linear Polarization (AoLP) values. It also enables single-shot disentanglement of polarimetric diffuse and specular reflections for accurate BRDF estimation. We achieve this spatial polarization multiplexing (SPM) with a constrained de Bruijn sequence. We validate this novel invisible single-shot shape and reflectance  method with real static and dynamic objects. The results demonstrate the effectiveness of SPM for accurate shape and BRDF measurement which opens new avenues of application for 3D sensing thanks to its invisibility and ability to jointly recover the radiometric properties.

\end{abstract}

\section{Introduction}
\label{sec:intro}

The appearance of an object is dictated by the its shape and reflectance. Recovering both from observations, \ie, images, is essential for a multitude of applications including real-world scene understanding, rendering, and virtual content creation.
For this, inverse rendering has been widely studied in the computer vision field. Multi-view inverse rendering requires a calibrated multi-view capture setup, such as a capture dome~\cite{oxholm2015shape,yamashita2023nlmvs,zhang2021physg,cheng2021multi,liang2024gs,jiang2024gaussianshader,bi2020deep}. Single-view inverse rendering is handy but challenging due to the ill-posedness caused by the entanglement of shape and reflectance, even when using additional cues like polarization.
Learning-based single-view inverse rendering methods exploit learned priors to overcome this ill-posedness~\cite{barron2014shape,yu2019inverserendernet,oxholm2015shape,deschaintre2021deep,li2018learning}. These methods, however, are fundamentally limited by the learned prior (\eg, object types in training).

Structured light sensing does not suffer from these fundamental limitations as they directly measure the actual surface by encoding geometric and photometric cues into designed illuminations. 
A projector-camera system realizing structured light is easy to integrate into a single compact device.
Single-shot structured light methods leverage a single spatial illumination (projector) pattern for shape acquisition. These methods cannot measure the reflectance as radiometric sensing inherently requires multiple observations under different lighting conditions~\cite{baek2022all,xu2023unified,ichikawa2024spiders}.
The light pattern for shape sensing also alters the object appearance, making joint sensing of shape and reflectance challenging.

Can we design a structured light pattern that enables joint shape and reflectance sensing in a single shot?
Such a single-shot shape and reflectance sensing can open new possibilities for 3D applications. 
Many natural and man-made objects alter their shape and appearance as we interact with them. For instance, imagine picking up a loaf, a pillow, or a plush toy. Phase transitions and chemical reactions, \eg, in baking or scientific experiments, also incur complex dynamics of the shape and appearance including swelling, shrinking, and material changes.

In this paper, we realize single-shot shape and reflectance sensing by fully leveraging polarization of light. We derive a novel single structured pattern with spatial polarization multiplexing, which we refer to as SPM, that enables robust decoding from its reflection by an object surface. Inspired by a recent polarization projector-camera system method (SPIDeRS) ~\cite{ichikawa2024spiders}, SPM uses per-pixel Angle of Linear Polarization (AoLP) control for encoding and decoding polarization, instead of visible intensity or color. The method simultaneously encodes the incident ray and samples multiple polarimetric reflections.
Since the polarization pattern is invisible to the naked eye, SPM can measure the shape and reflectance of the target without disrupting human experiences with them, \eg, enable unnoticed sensing as people interact with their surroundings. 

We design a stripe pattern with quantized AoLP so that their decoding after reflection can be achieved robustly and uniquely for shape and BRDF recovery.
For this, we devise a de Bruijn sequence that is constrained to have different neighboring symbols and assign quantized AoLP values to these symbols.
Adjacent stripes are designed to project different polarization states, which enables disentanglement of captured polarimetric image into polarimetric diffuse and specular reflections for single-shot reflectance recovery.
Additional shifted patterns can be used to recover finer shape and reflectance when the object happens to be static. 
By continuously projecting these shifted patterns, the target can be measured at adaptive resolutions depending on the motion and deformation of the objects in the scene.

We validate the accuracy of structured polarization multiplexing on real data with quantitative and qualitative evaluation of measured shape and reflectance for complex static and deforming dynamic objects. We show the effectiveness of SPM for shape reconstruction, decomposition of polarimetric reflections, and relighting from a single shot as well as adaptive reconstruction of dynamic objects with opportunistic shifted SPM patterns. 
We believe our method establishes a novel foundation of instant spatio-temporal 3D and material capture and opens a new avenue of applications of real-world 3D sensing and appearance modeling.

\section{Related Work}
\label{sec:related}
\Cref{tab:methods comparison} compares our method with past single-view shape acquisition methods using active illumination.

\newcolumntype{C}{>{\centering\arraybackslash}X}
\begin{table}[t]
    \centering
    \small
    \caption{
    Unlike classic structured light, SPM only uses a single shot for recovering both the target geometry and radiometry and retains the natural object appearance as polarization is invisible to the naked eye. 
    }
    \begin{tabularx}{\linewidth}{cCCC}
        \hline
        Method & Single shot & BRDF acquisition & Invisible pattern \\ \hline
        Single patt. SL & \checkmark & & \\
        ToF & \checkmark  & & \checkmark \\
        pToF~\cite{baek2022all}& & \checkmark & \checkmark \\
        Unified SL~\cite{xu2023unified} & & \checkmark & \\
        SPIDeRS~\cite{ichikawa2024spiders} & & \checkmark & \checkmark \\
        Ours & \checkmark & \checkmark & \checkmark  \\ \hline
    \end{tabularx}
    \vspace{-1.5em}
    
    \label{tab:methods comparison}
\end{table}

\vspace{-8pt}
\paragraph{Shape Sensing}
Structured light sensing measures 3D shape by decoding the reflected intensity or color patterns projected onto the target object. 
Several patterns have been proposed for different purposes~\cite{salvi2010state}.
Multi-pattern structured light~\cite{posdamer1982surface,caspi1998range,srinivasan1985automated, gupta2012micro, moreno2015embedded, gupta2018geometric} enables robust and accurate shape measurement but is fundamentally limited to static scenes. 

Single-pattern structured light~\cite{cong2014accurate,kawasaki2008dynamic,sagawa2009dense,sagawa2011dense,nguyen2022different,li2023self,song2022super} realizes shape reconstruction of dynamic scenes.
Fourier transform profilometry methods estimate the height from a reference plane as the phase of a sinusoidal pattern~\cite{takeda1983fourier,su2010dynamic,yang2017single}.
Spatially multiplexed patterns use continuous intensity or wavelength gradation~\cite{carrihill1985experiments,tajima19903} and discrete codes represented with different colors and geometric features~\cite{zhang2002rapid,pages2005optimised,zhou2023pattern}.
Other methods avoid directly using correspondences for dynamic scene reconstruction~\cite{furukawa2017depth,mirdehghan2024turbosl}.
These methods, however, only focus on shape acquisition, and cannot recover the reflectance which is essential for analyzing, viewing, and editing the object appearance.




\vspace{-8pt}
\paragraph{Joint Shape and Reflectance Sensing}

Several inverse rendering methods reconstruct shape and reflectance from multi-view images~\cite{oxholm2015shape,yamashita2023nlmvs,zhang2021physg,cheng2021multi,liang2024gs,jiang2024gaussianshader,bi2020deep}.
These multi-view methods, however, need multiple shots or cameras.
Multiple shots prevent their application to dynamic scenes, and using multiple cameras requires a calibrated setup which can be cumbersome to realize.
Although some methods address inverse rendering from only a single image~\cite{barron2014shape,yu2019inverserendernet,oxholm2015shape,deschaintre2021deep,li2018learning,lombardi2012single,lombardi2015reflectance,enyo2024diffusion}, they lack geometric constraints to reconstruct the 3D shape, \ie depth, and have to heavily rely on learned priors. Learning-based methods are inherently limited by the training data, which is hard to come by, especially for joint shape and reflectance estimation, and are biased towards what the model has learned. Our interest is in actual sensing of the physical instance observed in the real world. That is, our goal is to achieve direct measurement of the physical surface both in geometry and radiometry, which can also serve as ground truth for any learning-based methods. 

Controlled illumination helps disentangle shape and reflectance for accurate inverse rendering.
A combination of multi-view capture and controlled illumination has been used to acquire shape and spatially varying reflectance~\cite{zhou2013multi,tunwattanapong2013acquiring,kang2019learning,holroyd2010coaxial,nam2018practical,hwang2022sparse}.
Single-view methods~\cite{ma2007rapid,ghosh2010circularly,Baek2018SimultaneousAO,nam2016simultaneous} reconstruct the surface normal and reflectance but cannot obtain depth.
Recent works combine inverse rendering and active depth sensing, such as structured light and ToF, by using a dedicated projector-camera system~\cite{baek2022all, xu2023unified}.
Ichikawa \etal~\cite{ichikawa2024spiders} propose structured polarization for depth sensing with encoded AoLP patterns and also normal and reflectance recovery from polarimetric reflection.
These methods, however, require multiple shots under different lighting conditions and are limited to a static scene.

In contrast, our method captures depth and reflectance in a single shot with a projector-camera system, which can directly be applied to dynamic scenes.
Additionally, as the naked eye cannot perceive the polarization pattern, unlike structured light methods, our method retains the natural appearance of an object, which enables a wider range of applications (\eg, 3D and appearance sensing without being noticed for robot and human object interactions for HCI, robotics, and UI).

\section{Prerequisites}
Let us first review key properties of polarized light~\cite{Hecht}. For projection, we use the polarization projector introduced in~\cite{ichikawa2024spiders,li2024fooling} which can control the angle of linear polarization orientation (AoLP) at each pixel.
Please see the supplementary material for more details of its hardware implementation with a liquid crystal spatial light modulator.

\vspace{-8pt}
\paragraph{Polarization}
Light is a transverse electromagnetic wave and the orientation of its oscillation plane varies over time.
In contrast to unpolarized light which randomly changes its oscillating orientation, polarized light has a constant or periodically rotating oscillation plane. The former light is called linearly polarized and the latter is referred to as circularly polarized. Light can also be partially polarized, in-between polarized and unpolarized. 

A polarization filter lets through only light with a specific oscillating orientation. Light intensity of partially polarized light after a polarization filter at angle $\phi_c$ becomes
\begin{equation}
    I(\phi_c) = \overline{I}(1 + \rho_L\cos(2\phi_c - 2\phi_L)) \,,
    \label{eq:polarization filter}
\end{equation}
where $\overline{I}$ is the average intensity over the filter angle, $\rho_L$ is the Degree of Linear Polarization (DoLP), and $\phi_L$ is the Angle of Linear Polarization (AoLP).
We can reconstruct a linear polarization state $\overline{I}$, $\rho_L$, and $\phi_L$ by capturing the light intensity passing through different filter angles.
A polarimetric camera, \eg, Lucid TRI050S-PC with polarization image sensor Sony IMX250MZR, has on-chip four polarization filters of different angles, $0^\circ$, $45^\circ$, $90^\circ$, and $135^\circ$, for each $2\times2$ matrix of the sensor array.
It enables us to capture linear polarization states in a single shot.
It cannot distinguish unpolarized and circularly polarized light.

Mueller calculus with Stokes vectors is a useful mathematical representation to formulate polarimetric behavior on the surface.
Stokes vector represents partially polarized light, defined as
\begin{equation}
    \VEC{s} = \begin{bmatrix} s_0 \\ s_1 \\ s_2 \\ s_3 \end{bmatrix} =  \begin{bmatrix}
    I(0) + I(\frac{\pi}{2}) \\ I(0) - I(\frac{\pi}{2}) \\ I(\frac{\pi}{4}) - I(\frac{3}{4}\pi) \\ s_3
    \end{bmatrix} = \begin{bmatrix} 2\overline{I} \\ 2\overline{I}\rho_L\cos(2\phi_L) \\ 2\overline{I}\rho_L\sin(2\phi_L) \\ s_3 \end{bmatrix}\,,
    \label{eq:Stokes vector}
\end{equation}
where $s_0$ represents the light intensity, $s_1$ and $s_2$ represent the linearly polarized light, and $s_3$ represents the circularly polarized light.
From \cref{eq:Stokes vector}, the DoLP and AoLP can be recovered as
\begin{equation}
    \rho_L = \frac{\sqrt{s_1^2 + s_2^2}}{s_0} \,,\quad \phi_L = \frac{1}{2}\tan^{-1}\left( \frac{s_2}{s_1} \right) \,,
\end{equation}
respectively.
The Mueller matrix represents the change of the polarization state as
\begin{equation}
    \VEC{s}^o = \VEC{M}\VEC{s}^i \,,
    \label{eq:Mueller matrix}
\end{equation}
where $\VEC{s}^i$ and $\VEC{s}^o$ are Stokes vectors of input and output light, respectively.
We use Mueller calculus to analyze the polarimetric reflection on an object surface.

\vspace{-8pt}
\paragraph{Polarimetric Reflection of Polarization Pattern}
\label{sec:polarimetric reflection}
The polarized light pattern emitted from the polarization projector gets reflected by the target object surface.
A polarimetric reflectance model expresses the modulation of the polarization states by the surface with the Mueller matrix~\cite{Baek2018SimultaneousAO,kondo2020accurate,hwang2022sparse,ichikawa2023fresnel}.
Baek \etal~\cite{Baek2018SimultaneousAO} approximate the polarimetric specular and diffuse reflections, $\VEC{M}^s$ and $\VEC{M}^d$, by assuming a co-axial setup 
\begin{equation}
\scalebox{0.9}{$
\begin{aligned}
    \VEC{M} &= \VEC{M}^s + \VEC{M}^d \\
    &\approx c^s {\footnotesize \begin{bmatrix}
        1 & 0 & 0 & 0 \\ 0 & 1 & 0 & 0 \\ 0 & 0 & -1 & 0 \\ 0 & 0 & 0 & -1
    \end{bmatrix}}
    +
    c^d {\footnotesize\begin{bmatrix}
        1 & m^d_{01} & m^d_{02} & 0 \\ m^d_{10} & 0 & 0 & 0 \\ m^d_{20} & 0 & 0 & 0 \\ 0 & 0 & 0 & 0
    \end{bmatrix}}
    \,,
\end{aligned}
$}
\label{eq: Mueller matrix of reflection}
\end{equation}
where $c^s$ and $c^d$ are radiometric specular and diffuse terms representing the shading and specular BRDFs, respectively, and $m^d_{ij}$ consists of Fresnel transmittance on the surface.

SPIDeRS~\cite{ichikawa2024spiders} analyzes these approximated polarimetric reflections for reflectance recovery from the reflected polarization pattern.
This method, however, requires two additional uniform polarization patterns for robust decoding and BRDF reconstruction.
In contrast, we design a novel polarization pattern by spatially multiplexing the AoLP and achieve joint recovery of shape and reflectance from purely a single polarimetric image capture.

\begin{figure}[t]
    \centering

    \input{fig_tex/overview}
    
    \vspace{-0.5em}
    \caption{We design an SPM pattern with quantized stripes of AoLP values. Per-pixel polarization control enables robust decoding based on neighboring pixels and decomposition of captured polarimetric image into polarimetric diffuse and specular reflections for BRDF reconstruction.}
    \vspace{-1em}
    \label{fig: overview}

\end{figure}

\section{Spatial Polarization Multiplexing}
Many spatially-multiplexed structured light patterns have been proposed in the past~\cite{boyer1987color,zhang2002rapid,pages2005optimised} which  are all used for intensity or color encoding and are also limited to shape acquisition.
In sharp contrast, our focus is spatial multiplexing of polarized light for joint shape and reflectance acquisition. 

For hardware configuration, we follow the projector-camera proposed in SPIDeRS~\cite{ichikawa2024spiders}. To balance the co-axial assumption of polarimetric reflection for decoding and to ensure sufficient parallax, we locate the projector and camera with enough baseline at a distance from the target object.
We also assume that there is no ambient light.

\subsection{Quantization of AoLPs}
\label{sec: quantization of AoLPs}
Our method encodes and decodes structured polarization with AoLP values.
As shown in \cref{fig: AoLP modulation image}, the diffuse reflection represented by $c^dm^d_{10}$ and $c^dm^d_{20}$ modulates the projected AoLP pattern. SPIDeRS~\cite{ichikawa2024spiders} addressed this by capturing additional shots.
To address this corruption in a single shot, the AoLP values used in SPM are quantized for robust decoding. 

Let us show how quantization allows us to deal with the AoLP corruption by analyzing the effect of diffuse polarization represented as \cref{eq: Mueller matrix of reflection}.
We reparameterize $m^d_{10}$ and $m^d_{20}$ with the diffuse DoLP and AoLP
\begin{equation}
    \rho^d_L = \sqrt{\left( m^d_{10} \right)^2 + \left( m^d_{20} \right)^2} \,,\quad \phi^d_L = \frac{1}{2}\tan^{-1}\left( \frac{m^d_{20}}{m^d_{10}} \right) \,.
    \label{eq:diffuse polarization}
\end{equation}
From \cref{eq:Mueller matrix,eq: Mueller matrix of reflection,eq:diffuse polarization}, the AoLP of the reflected light $\phi^o_L$ becomes
\begin{equation}
    \phi^o_L = \frac{1}{2}\tan^{-1}\frac{c^d \rho^d_L\sin \left(2\phi^d_L \right) - c^s\rho^i_L\sin \left( 2\phi^i_L \right)}{c^d \rho^d_L\cos \left(2\phi^d_L \right) + c^s\rho^i_L\cos \left( 2\phi^i_L \right)}\,,
    \label{eq:reflected AoLP rewrite}
\end{equation}
where $\rho^i_L$ and $\phi^i_L$ are the DoLP and AoLP of the projected pattern, respectively.
While the DoLP of the polarization projector $\rho^i_L$ can be 
set to $\approx 1$~\cite{ichikawa2024spiders},
the DoLP of diffuse reflection is small.
The diffuse DoLP $\rho^d_{L}$ modeled as Fresnel transmission~\cite{Baek2018SimultaneousAO,ichikawa2023fresnel} is much lower than $1.0$ for all zenith angles.
In particular, when the refractive index is $1.5$ and the zenith angle is less than $45^\circ$, the diffuse DoLP is $\rho^d_L < 0.05$.

We compute the AoLP modulation for $\frac{c^d}{c^s}$ when $\frac{\rho^d_L}{\rho^i_L} = 0.05$.
When $\frac{c^d}{c^s} = 2$, $\frac{c^d}{c^s} = 3$, $\frac{c^d}{c^s} = 5$, the AoLP modulations are less than $2.9^\circ$, $4.3^\circ$, and $7.3^\circ$ for all diffuse AoLP $\phi^d_L$, respectively.
Even when the diffuse reflection $c^d$ is dominant, the AoLP change is within only a few degrees.
In the supplementary material, we show the diffuse DoLP and AoLP modulation derivation in detail.
When this AoLP corruption is smaller than the quantization width, the projected AoLP is correctly detected. 
Otherwise, decoding using dynamic programming~\cite{zhang2002rapid} can correct mislabeling by leveraging the global pattern as described in \cref{sec: spatial polarization multiplexing pattern}.

Parallax also causes AoLP modulation by non-diagonal elements of polarimetric specular reflection in \cref{eq: Mueller matrix of reflection}. When the angle between the viewing and lighting directions is less than $20^\circ$, AoLP change by Fresnel reflection is less than $1.2^\circ$~\cite{ichikawa2024spiders}. As such, we can handle small parallax by AoLP quantization.

\begin{figure}[t]
    \centering
    \input{fig_tex/AoLP_modulation_image}
    \vspace{-0.6em}
    \caption{Pattern detection and decoding using quantized AoLP values. As shown in the inset, the observed AoLPs are corruptlll due to diffuse reflection. ``Detected'': the quantized AoLPs are incorrect where the corruption exceeds tolerance. ``Decode'': our decoding corrects these mislabelings.}
    \vspace{-1.2em}
    \label{fig: AoLP modulation image}
\end{figure}

\subsection{Spatial Polarization Multiplexing Pattern}
\label{sec: spatial polarization multiplexing pattern}

We design an SPM pattern that enables simultaneous shape and BRDF recovery based on a de Bruijn sequence~\cite{zhang2002rapid,pages2005optimised} with quantized AoLP values. 
The substrings of length-$\ndb$ in a de Bruijn sequence are all unique, enabling unique decoding for shape reconstruction.
An example is 
``012022110''. This sequence contains all nine substrings of $\ndb=2$ consisting of 0, 1, and 2, where 00 wraps around.

The polarization pattern is constructed by assigning the quantized AoLP values to the sequence symbols.
As shown in \cref{fig: de Bruijn example}, instead of directly using the de Bruijn sequence, two constraints are imposed on it.
First, AoLP values of three sequential elements must be different for polarimetric diffuse and specular decomposition using neighboring pixels.
Second, the differences between AoLP values of adjacent elements must be more than one quantization step for robust stripe detection.
The constrained de Bruijn sequence can be constructed as an Eulerian path of a graph. 
In the supplementary material, we show how to modify an original graph by deleting the paths and nodes that violate our constraints.

\begin{figure}[t]
    \centering

    \input{fig_tex/de_Bruijn_example}
    \vspace{-0.8em}
    \caption{Example of our polarization pattern for $\kdb=6$, $\ndb=3$, and AoLP range $[0^\circ,80^\circ]$. We impose constraints on the de Bruijn sequence to prevent adjacent stripes with identical or near AoLP values for polarimetric decomposition and robust detection of the pattern.}

    \vspace{-1.5em}
    \label{fig: de Bruijn example}
\end{figure}

Reflected stripe pattern along each horizontal scanline in the captured polarimetric image is first detected.
In the local window, each pixel votes for the symbols by quantizing the observed AoLP.
The symbols that gather enough votes are selected and the center position of each is assigned the detected symbols.
By sliding the local window, the symbols of the stripe pattern and their positions across the scanline can be detected.

Inspired by past structured scanline color decoding~\cite{zhang2002rapid,pages2005optimised}, we use dynamic programming to decode the sequence of the detected symbols.
The matching score is 
\begin{equation}
    \mathrm{score}(d_i, p_j) = \cos\left(2\phi_{d_i} - 2\phi_{p_j}\right) - \cos\left( 2\phi_\mathrm{th} \right)\,,
\end{equation}
where $\phi_{d_i}$ and $\phi_{p_j}$ are the AoLP values corresponding to the symbols $d_i$ and $p_j$ at the $i$-th detected line and the $j$-th projected line, respectively.
The threshold for matching is set as $\phi_\mathrm{th} = 30^\circ$.
Global matching using dynamic programming can handle mislabeling, misdetection, and occlusion.

The 3D points of decoded pixels can be reconstructed by triangulation--they are the intersection points of the camera ray and the plane projected as the stripe pattern.
The decoding also provides a pair of incident and observed Stokes vectors for each point.
Adjacent matched points on the scanline are used to separate polarimetric diffuse and specular reflections as described in \cref{sec: polarimetric diffuse and specular separation}.

\subsection{Polarimetric Diffuse--Specular Decomposition}
\label{sec: polarimetric diffuse and specular separation}
The decomposition of the Mueller matrix into polarimetric diffuse and specular reflections provides rich surface properties~\cite{Baek2018SimultaneousAO}.
Measuring the Mueller matrix, however, requires multiple polarimetric images while rotating the polarization orientation of incident light.

As shown in \cref{fig: overview}, neighboring projector pixels of SPM with different polarization states can be used for single shot separation of polarimetric diffuse and specular reflections. 
Although diffuse polarization is ignored during AoLP quantization for robust decoding, it can be reconstructed by directly using the observed Stokes vectors of each correspondence.

For robust polarimetric decomposition, the Mueller matrix elements from \cref{eq: Mueller matrix of reflection} can be reduced by exploiting the near co-axial projector-camera setting as
\begin{equation}
    \begin{split}        
    & m_{12} = m_{21} = 0\,, \qquad  m_{33} = m_{22} = -m_{11}\,, \\
    & m_{30} = m_{31} = m_{32} = m_{03} = m_{13} = m_{23} = 0 \,,
    \label{eq: coaxial assumption}
    \end{split}
\end{equation}
where $m_{ij}$ is the $(i,j)$ element of the Mueller matrix $\VEC{M}$ of the target surface.
A transpositional reciprocity relationship of the Mueller matrix~\cite{ding2024polarimetric} can also be leveraged, where the interchanging of incident light and viewing directions results in transposition
\begin{equation}
    m_{10} = m_{01}\,,\ \  m_{20} = -m_{02}\,.
    \label{eq: transpositional reciprocity}
\end{equation}
From \cref{eq: coaxial assumption,eq: transpositional reciprocity}, the Mueller matrix has four remaining elements: $m_{00}$, $m_{11}$, $m_{10}$, and $m_{20}$.
Since the Stokes vector of linear polarization has three elements, at least two pairs of projected and observed Stokes vectors are required.
This indicates that our SPM is essential to reconstruct these Mueller matrix elements.

By using the pairs of observed and incident Stokes vectors at the matched point and its adjacent matched points, $m_{00}, m_{10}, m_{20}$ and $m_{11}$ are reconstructed.
First, the second and third rows of the Mueller matrix are computed as 
\begin{equation}
\scalebox{0.95}{$
    \begin{bmatrix}
        s^o_{1,1} \\ s^o_{1,2} \\ \vdots \\ s^o_{N,1} \\ s^o_{N,2}
    \end{bmatrix}
    =
    \begin{bmatrix}
        s^i_{1,0} & 0 & s^i_{1,1} \\
        0 & s^i_{1,0} & -s^i_{1,2} \\
        \vdots & \vdots & \vdots \\
        s^i_{N,0} & 0 & s^i_{N,1} \\
        0 & s^i_{N,0} & -s^i_{N,2}        
    \end{bmatrix}
    \begin{bmatrix} m_{10} \\ m_{20} \\ m_{11} \end{bmatrix}
    $}
    \,,
    \label{eq:Mueller12}
\end{equation}
where $s^{i/o}_{k,j}$ is the $j$-th element of the $k$-th observed Stokes vector $\VEC{s}^{i/o}_{k}$ and $N$ is the number of used Stokes vector pairs.
Then, $m_{00}$ is computed from a single Stokes vector pair for each correspondence as
\begin{equation}
    m_{00} = \frac{s^o_0 - s^i_1 m_{10} + s^i_2 m_{20}}{s^i_0}\,.
    \label{eq:Mueller0}
\end{equation}
Polarimetric diffuse and specular reflections are then extracted:
\begin{equation}
    \begin{split}
    & c^s = m_{11}\,,\ \ c^d = m_{00} - m_{11}\,, \\
    & m^d_{10}=m^d_{01}=\frac{m_{10}}{c^d}\,,\ \ m^d_{20}=-m^d_{02}=\frac{m_{20}}{c^d}\,.
    \label{eq: specular and diffuse separation}
    \end{split}
\end{equation}

\subsection{BRDF Estimation}
SPM enables BRDF recovery using the measured 3D shape and decomposed polarimetric reflections of the target.
We adopt the Fresnel microfacet BRDF model~\cite{ichikawa2023fresnel}, which describes polarimetric and radiometric surface (specular) and body (diffuse) reflections in a  unified model.
Initial surface normals of the reconstructed point cloud can be computed with Principal Component Analysis (PCA) of neighboring points.

We assume that the material parameters, refractive index $\mu$, surface reflection albedo $k_s$, and microfacet distribution $\alpha$ and $\beta$, are uniform on the surface.
For body reflection, we assume mesoscopic roughness $\kappa$ is fixed at $0$ and body reflection albedo $\VEC{K}_b$ is spatially varying.
First, the surface normals and material parameters are estimated from the specular reflection and the diffuse polarization as
\begin{equation}
    \begin{split}
    \argmin_{\substack{\mu, k_s, \alpha, \\ \beta, \VEC{N}}} & \sum_i^L \left| c^s_i - \overline{c}^s_i \right|^2 + \lambda_d |m^d_{i, 10} - \overline{m}^d_{i,10}|^2 \\
    & + \lambda_d|m^d_{i,20} - \overline{m}^d_{i,20}|^2 + \lambda_n\|\VEC{n}_i - \VEC{n}_i^\mathrm{init}\| \,,
    \end{split}
    \label{eq:specular optimization}
\end{equation}
where $\overline{\cdot}_i$ is the rendered component at the $i$-th correspondence, $L$ is the number of correspondences, $\VEC{N}$ is a collection of surface normals $\VEC{n}_i$, and $\VEC{n}_i^\mathrm{init}$ is the initial normals.
We set $\lambda_d = 0.01$ and $\lambda_n = 0.001$.
Then, the diffuse reflection albedo values are recovered as
\begin{equation}
    \argmin_{\VEC{K}_b} \sum_i^L \left| c^d_i - \overline{c}^d_i \right|^2\,.
    \label{eq:diffuse optimization}
\end{equation}
Since the body reflection albedo $k_{b,i} \in \VEC{K}_b$ represents the scale of $\overline{c}^d_i$, \cref{eq:diffuse optimization} admits an analytical solution.

\section{Shifted Spatial Polarization Multiplexing}
\begin{figure}[t]
    \centering

    \input{fig_tex/shifted_pattern}
    \vspace{-0.8em}
    \caption{We smooth and shift the SPM pattern for high-resolution reconstruction of a static scene. We also change the AoLP assignment for each pattern to ensure sufficient variations of projected polarization states. The projected Stokes vectors are scaled and translated.
    The matching between the observed and projected Stokes vectors provides pixel-wise correspondences.
    }
    \vspace{-1.2em}
    \label{fig: shifted pattern}
\end{figure}
The resolution of the sensed surface is limited by the stripe width of the SPM pattern. 
For a static scene, multiple shifted SPM patterns can be leveraged for high-resolution reconstruction.
Furthermore, by continuously projecting the shifted patterns, the reconstruction resolution can be adapted to the scene--higher resolution for static regions and lower resolution for dynamic regions. 

\subsection{High-Resolution Recovery for Static Scene}
\label{sec: high-resolution recovery for static scene}
As shown in \cref{fig: shifted pattern}, to obtain dense, sub-pixel correspondences, we smooth the edge of the SPM pattern by applying a Gaussian filter. 
For matching between smoothed patterns and polarimetric images, we directly use Stokes vectors, instead of the non-linearly modulated reflected AoLPs.

The Stokes vectors are linearly transformed by the Mueller matrix, as represented with \cref{eq:Mueller matrix,eq: Mueller matrix of reflection}.
The observed Stokes vectors of linear polarization become
\begin{equation}
    s^o_{1} = c^d m^d_{10}+ c^s s^i_{1}\,,\ \ s^o_{2} = c^d m^d_{20} - c^s s^i_{2}\,.
    \label{eq: stokes vector modulation}
\end{equation}
\Cref{eq: stokes vector modulation} indicates $s^o_{1}$ and $s^o_{2}$ are scaled by $c^s$ and translated by $c^d m^d_{10}$ and $c^d m^d_{20}$.

When multiple polarimetric images are available, pixel-wise correspondences can be established by defining the matching cost to simultaneously estimate the scale and translation. 
While correct matching yields small residuals between the observed and incident Stokes vectors with the optimal scale and translation, incorrect matching yields larger residuals.
The matching cost function reflects this with 
\begin{equation}
    \begin{split}
        C(\VEC{S}_k^o, \VEC{S}_j^i) & = \min_{a, b_1, b_2} \| a \VEC{S}_j^i + b_1 \VEC{1}_\mathrm{o} + b_2\VEC{1}_\mathrm{e} - \VEC{S}_k^o \|^2 \\
        & + \min_{a, b_1, b_2} \| a \VEC{S}_k^o + b_1 \VEC{1}_\mathrm{o} + b_2\VEC{1}_\mathrm{e} - \VEC{S}_j^i \|^2 \,,
        \end{split}
\end{equation}
where $\VEC{S}_k^o$ and $\VEC{S}_j^i$ are the stacks of observed and projected Stokes vectors of the $k$-th and the $j$-th pixels on the scanline, respectively, $\VEC{1}_\mathrm{o}$ are vectors whose odd-numbered elements are 1 and else 0, and $\VEC{1}_\mathrm{e}$ are vectors whose even-numbered elements are 1 and else 0. Note that the signs of $s^i_2$ in $\VEC{S}_j^i$ are inverted. The symmetric cost function stabilizes matching even when the norm of $\VEC{S}^o_k$ is close to zero~\cite{zhang2002rapid}.

Following~\cite{zhang2002rapid}, by inverting the matching cost, the matching score function is defined as 
\begin{equation}
    \mathrm{score}(\VEC{S}_k^o, \VEC{S}_j^i) = C_0 - C(\VEC{S}_k^o, \VEC{S}_j^i)\,,
\end{equation}
where the matching threshold $C_0$ is
\begin{equation}
    C_0 = (1-t)\min_{k,j}C(\VEC{S}_k^o,\VEC{S}_j^i) + t\max_{k,j}C(\VEC{S}_k^o,\VEC{S}_j^i)\,,
\end{equation}
with $t=0.2$.
Sub-pixel correspondences is computed by refining the pixel-wise correspondences with local maxima of the score function with respect to projector pixels. 
We fit the parabola to the points $(j-1, \mathrm{score}(\VEC{S}_k^o, \VEC{S}^i_{j-1})$, $(j, \mathrm{score}(\VEC{S}_k^o, \VEC{S}^i_{j})$, and $(j+1, \mathrm{score}(\VEC{S}_k^o, \VEC{S}^i_{j+1})$, and obtain its peak as the sub-pixel correspondences.

After establishing the correspondences, multiple pairs of the projected and observed Stokes vectors are obtained for each pixel.
From \cref{eq:Mueller12,eq:Mueller0,eq: specular and diffuse separation,eq:specular optimization,eq:diffuse optimization}, polarimetric diffuse and specular reflections are extracted and then the BRDF is recovered. 
To ensure sufficient variations of projected polarization states for the polarimetric reflectance decomposition, the AoLP assignment for each shifted pattern is changed.

\subsection{Adaptive Resolution for Dynamic Scene}
When capturing the continuously shifted spatial polarization multiplexing, the target surface may be static for some regions or frames and dynamic for others.
To fully leverage our shifted patterns even in dynamic scenes, we adapt the number of polarimetric images used for reconstruction, depending on whether each region is dynamic or static.

As shown in \cref{fig: adaptive resolution}, each pixel in frame is labeled as ``dynamic'' or ``static.''
Since the polarization pattern has little effect on the object appearance, the differences in light intensity $s^o_0$ between frames can be used for this labeling.
When the intensity does not change between frames, the pixel is static. Otherwise, the pixel is dynamic.
We use each polarimetric image at the dynamic pixels and multiple polarimetric images of continuous frames at the static pixels for decoding and reconstruction.
To obtain the correspondences by dynamic programming, the detected stripe of the dynamic regions and all pixels of the static regions are concatenated to compute the matching score by switching the score functions.

\begin{figure}[t]
    \centering
    \resizebox{1.0\linewidth}{!}{
\begin{tikzpicture}[x=0.001\linewidth,y=0.001\linewidth,every node/.style={inner sep=0pt, text depth=0pt}]
{
    \newcount\vX
    \newcount\vY
    \newcount\vdX
    \newcount\vdY
    \vX = 0
    \vY = 0
    \vdX = 140
    \def\mysz{0.143\linewidth}

    \newcount\vdXerror
    \newcount\vdYerror
    \def\myerror#1{\tiny \textcolor{white}{#1}}
    \def\showerror#1{
        \advance\vX by \vdXerror
        \advance\vY by \vdYerror
        \node[inner sep=0pt] (a) at (\vX,\vY) {\myerror{#1}};
        \advance\vX by -\vdXerror
        \advance\vY by -\vdYerror
    }

    \def\myfont{\scriptsize}
    \node[inner sep=0pt, font=\myfont] (a) at (\vX,\vY) {GT depth};
    \advance\vX by \vdX
    \node[inner sep=0pt, font=\myfont] (a) at (\vX,\vY) {Ours};
    \advance\vX by \vdX
    \node[inner sep=0pt, font=\myfont] (a) at (\vX,\vY) {DPI~\cite{deschaintre2021deep}};
    \advance\vX by \vdX
    \node[inner sep=0pt, font=\myfont] (a) at (\vX,\vY) {GT normal};
    \advance\vX by \vdX
    \node[inner sep=0pt, font=\myfont] (a) at (\vX,\vY) {Ours};
    \advance\vX by \vdX
    \node[inner sep=0pt, font=\myfont] (a) at (\vX,\vY) {DPI~\cite{deschaintre2021deep}};
    \advance\vX by \vdX
    \node[inner sep=0pt, font=\myfont] (a) at (\vX,\vY) {SfPDLE~\cite{lyu2023shape}};

    \vdY = -80
    \vX = 0
    \vdXerror = 0
    \vdYerror = -47
    \advance\vY by \vdY

    \def\myincA#1{\resizebox{\mysz}{!}{\adjincludegraphics[Clip={{0.2\width} {0.18\width} {0.3\width} {0.2\width}}]{#1}}}
    \node[inner sep=0pt] (a) at (\vX,\vY) {\myincA{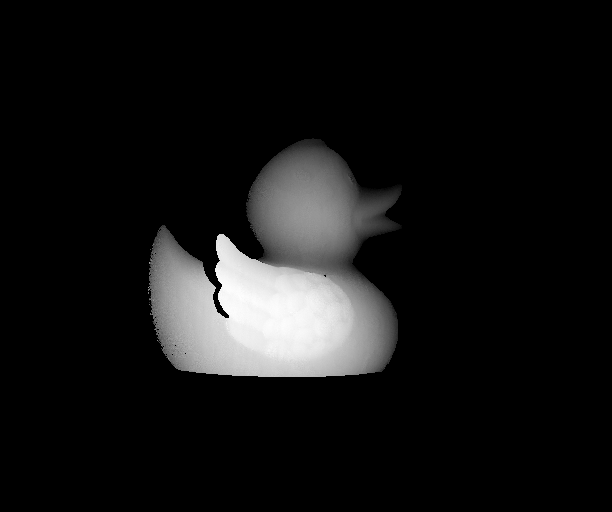}};
    \advance\vX by \vdX
    \node[inner sep=0pt] (a) at (\vX,\vY) {\myincA{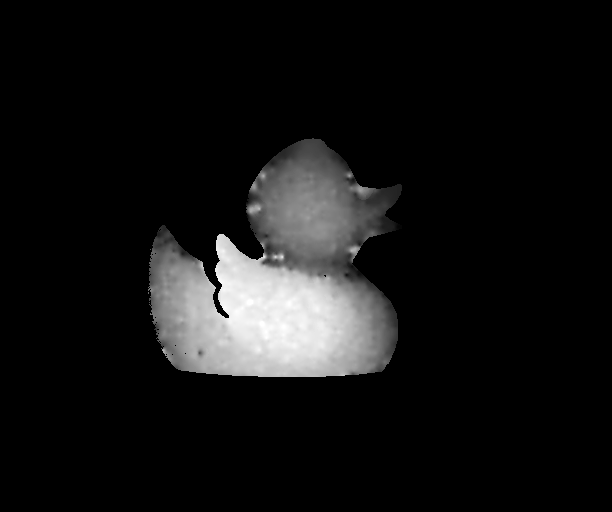}};
    \showerror{$\mathbf{1.23}$/$\mathbf{0.90}$}
    \advance\vX by \vdX
    \node[inner sep=0pt] (a) at (\vX,\vY) {\myincA{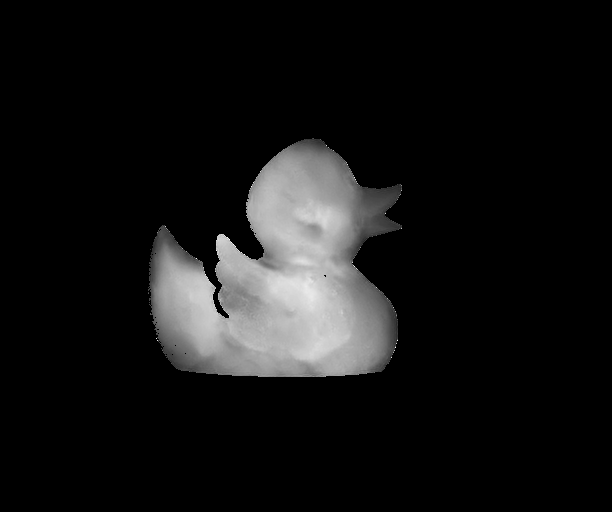}};
    \showerror{$4.68$/$4.44$}
    \advance\vX by \vdX
    \node[inner sep=0pt] (a) at (\vX,\vY) {\myincA{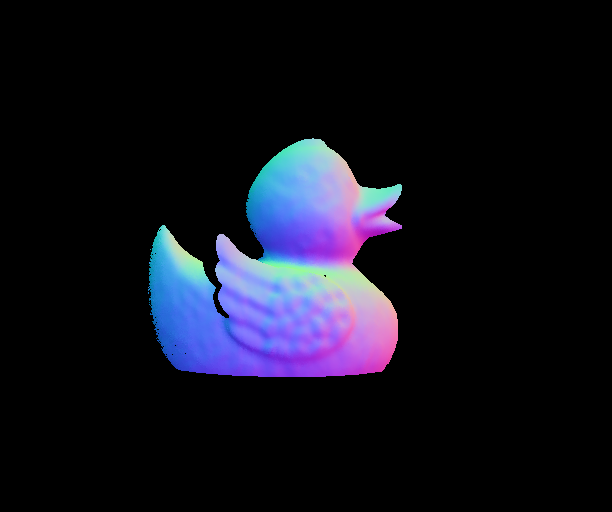}};
    \advance\vX by \vdX
    \node[inner sep=0pt] (a) at (\vX,\vY) {\myincA{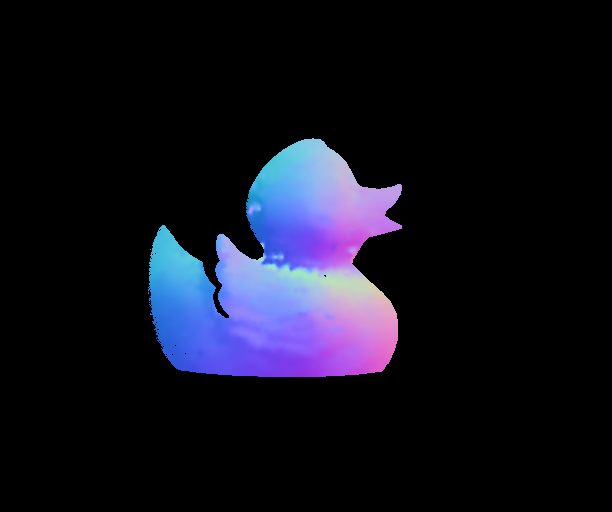}};
    \advance\vX by -5
    \showerror{$\mathbf{10.9^\circ}$/$\mathbf{7.51^\circ}$}
    \advance\vX by 5
    \advance\vX by \vdX
    \node[inner sep=0pt] (a) at (\vX,\vY) {\myincA{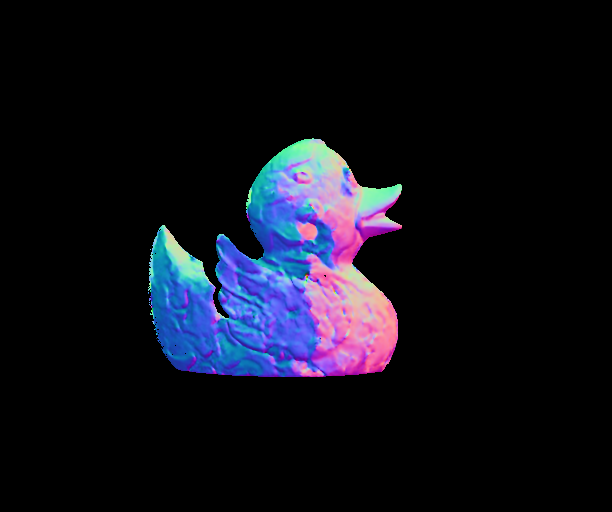}};
    \showerror{$30.7^\circ$/$25.3^\circ$}
    \advance\vX by \vdX
    \node[inner sep=0pt] (a) at (\vX,\vY) {\myincA{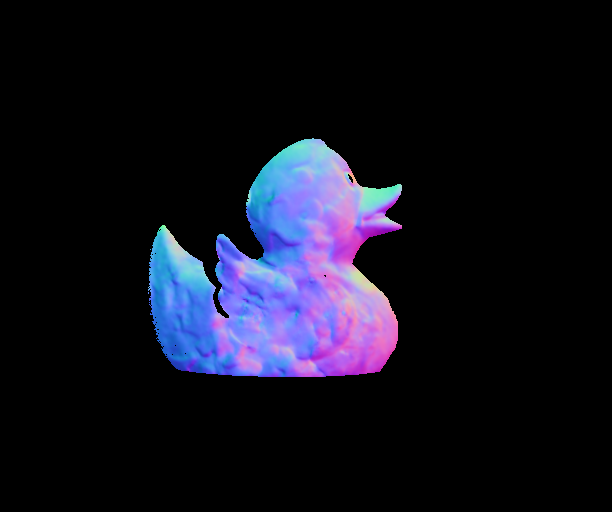}};
    \showerror{$18.0^\circ$/$13.1^\circ$}

    \vdY = -108
    \vX = 0
    \vdXerror = 0
    \vdYerror = -32
    \advance\vY by \vdY

    \def\myincA#1{\resizebox{\mysz}{!}{\adjincludegraphics[Clip={{0.1\width} {0.12\width} {0.1\width} {0.17\width}}]{#1}}}
    \node[inner sep=0pt] (a) at (\vX,\vY) {\myincA{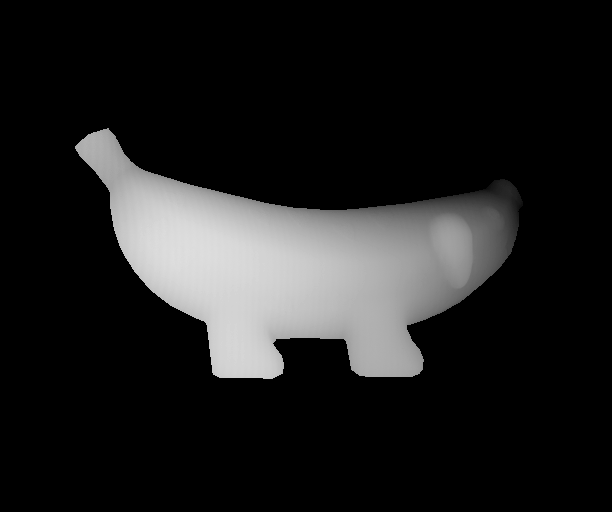}};
    \advance\vX by \vdX
    \node[inner sep=0pt] (a) at (\vX,\vY) {\myincA{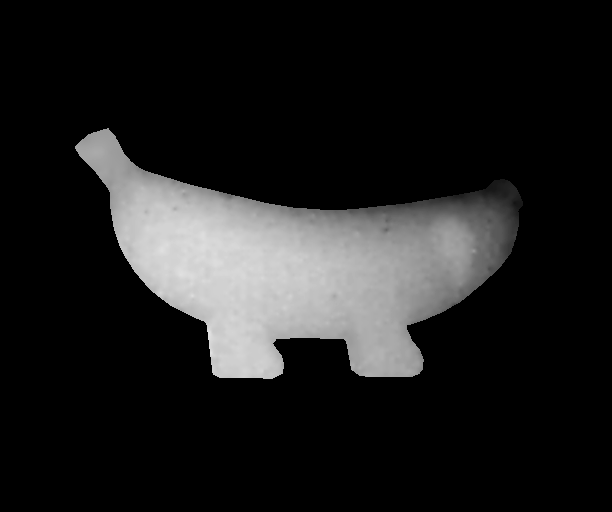}};
    \showerror{$\mathbf{0.97}$/$\mathbf{0.85}$}
    \advance\vX by \vdX
    \node[inner sep=0pt] (a) at (\vX,\vY) {\myincA{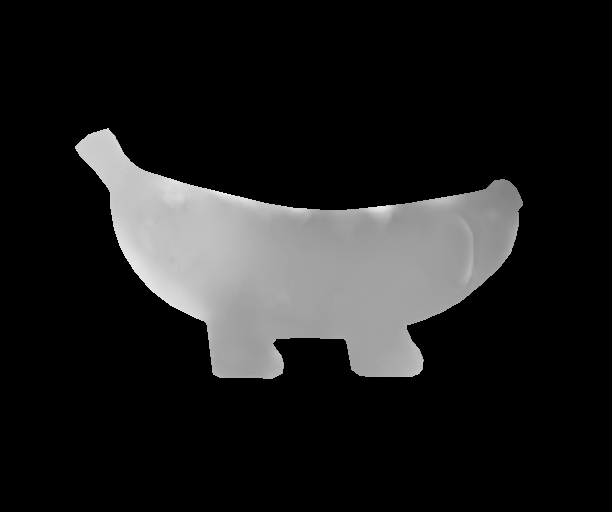}};
    \showerror{$5.76$/$5.16$}
    \advance\vX by \vdX
    \node[inner sep=0pt] (a) at (\vX,\vY) {\myincA{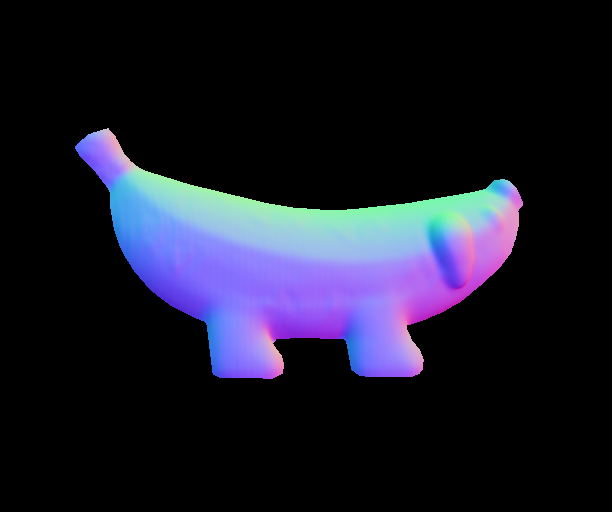}};
    \advance\vX by \vdX
    \node[inner sep=0pt] (a) at (\vX,\vY) {\myincA{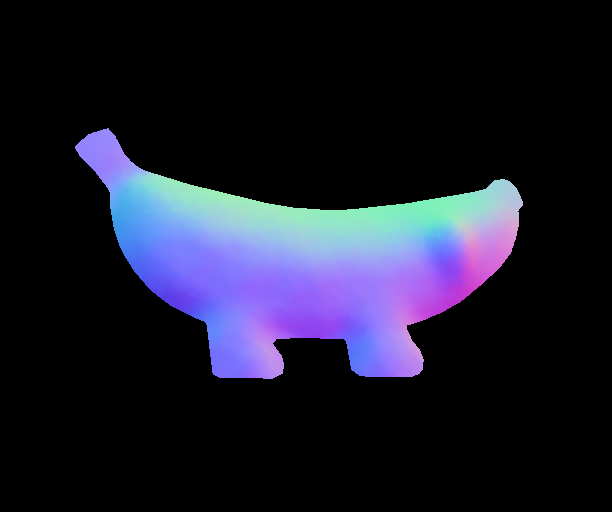}};
    \advance\vX by -5
    \showerror{$\mathbf{6.97^\circ}$/$\mathbf{4.32^\circ}$}
    \advance\vX by 5
    \advance\vX by \vdX
    \node[inner sep=0pt] (a) at (\vX,\vY) {\myincA{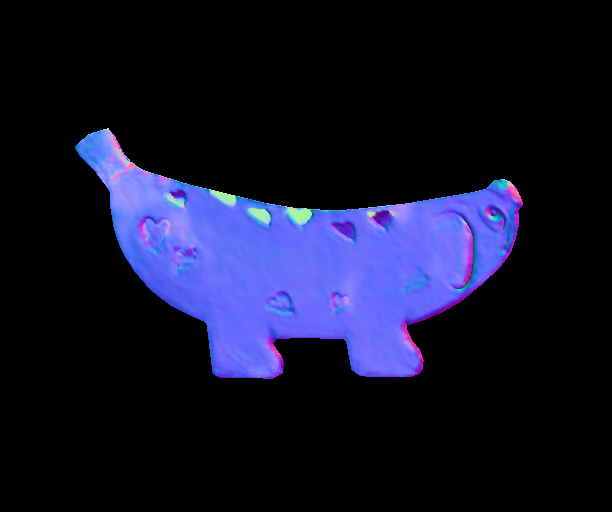}};
    \showerror{$35.2^\circ$/$30.2^\circ$}
    \advance\vX by \vdX
    \node[inner sep=0pt] (a) at (\vX,\vY) {\myincA{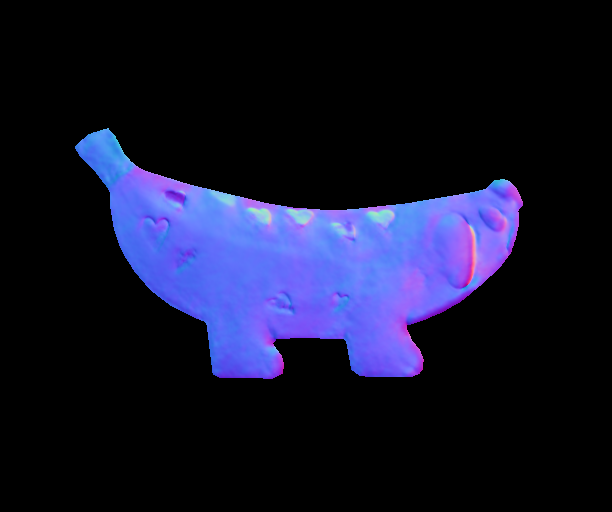}};
    \showerror{$30.6^\circ$/$27.0^\circ$}

    \vdY = -122
    \vX = 0
    \vdXerror = 0
    \vdYerror = -60
    \advance\vY by \vdY

    \def\myincA#1{\resizebox{\mysz}{!}{\adjincludegraphics[Clip={{0.07\width} {0.08\width} {0.3\width} {0.1\width}}]{#1}}}
    \node[inner sep=0pt] (a) at (\vX,\vY) {\myincA{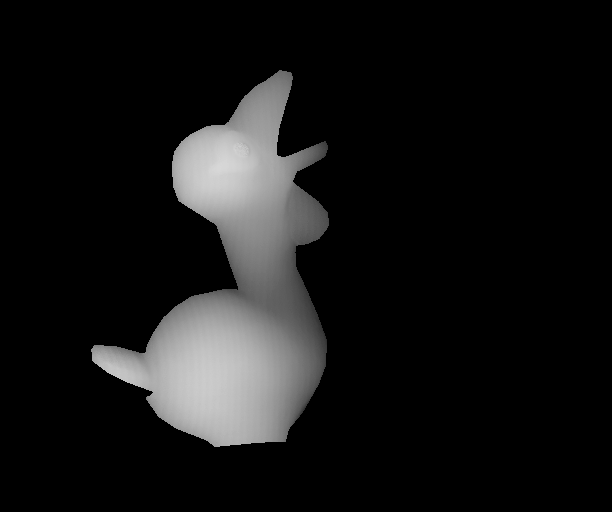}};
    \advance\vX by \vdX
    \node[inner sep=0pt] (a) at (\vX,\vY) {\myincA{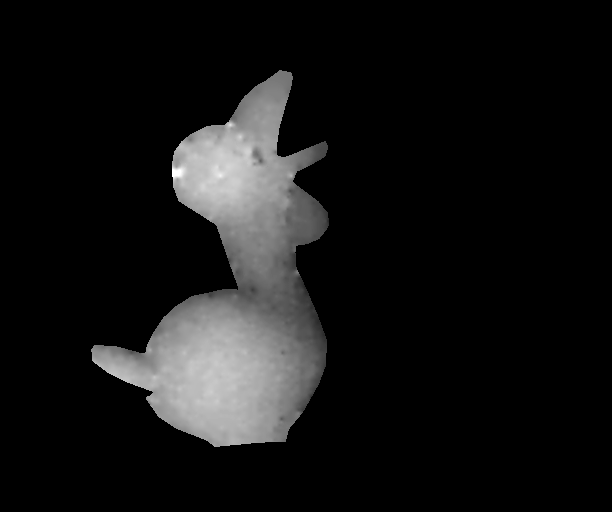}};
    \showerror{$\mathbf{1.19}$/$\mathbf{1.01}$}
    \advance\vX by \vdX
    \node[inner sep=0pt] (a) at (\vX,\vY) {\myincA{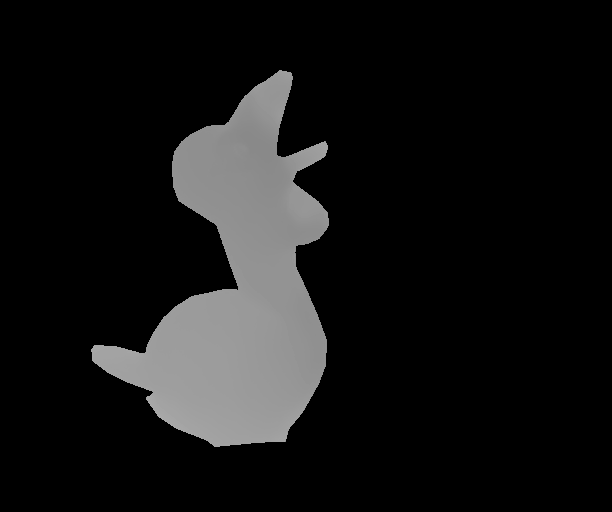}};
    \showerror{$4.81$/$4.52$}
    \advance\vX by \vdX
    \node[inner sep=0pt] (a) at (\vX,\vY) {\myincA{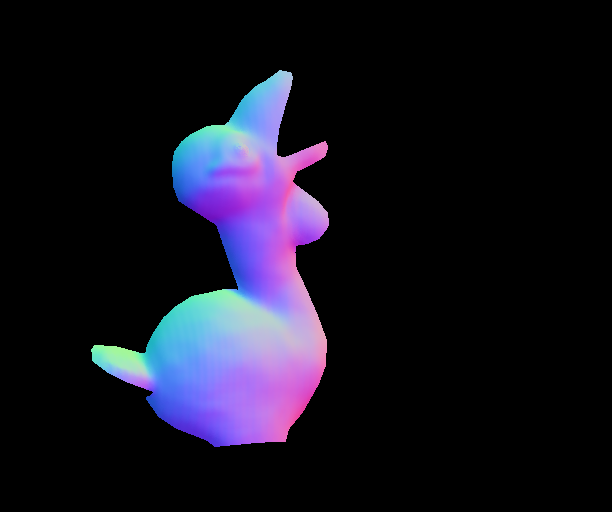}};
    \advance\vX by \vdX
    \node[inner sep=0pt] (a) at (\vX,\vY) {\myincA{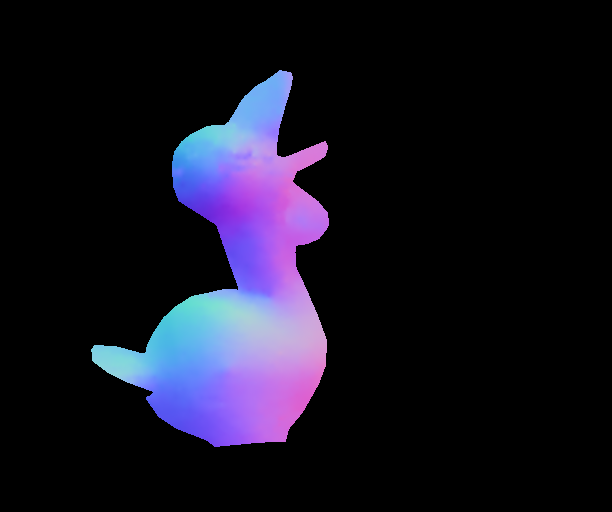}};
    \advance\vX by -5
    \showerror{$\mathbf{9.43^\circ}$/$\mathbf{6.55^\circ}$}
    \advance\vX by 5
    \advance\vX by \vdX
    \node[inner sep=0pt] (a) at (\vX,\vY) {\myincA{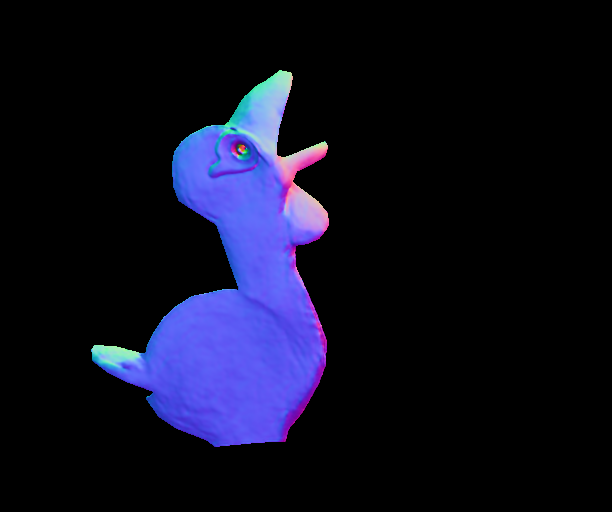}};
    \showerror{$38.2^\circ$/$37.0^\circ$}
    \advance\vX by \vdX
    \node[inner sep=0pt] (a) at (\vX,\vY) {\myincA{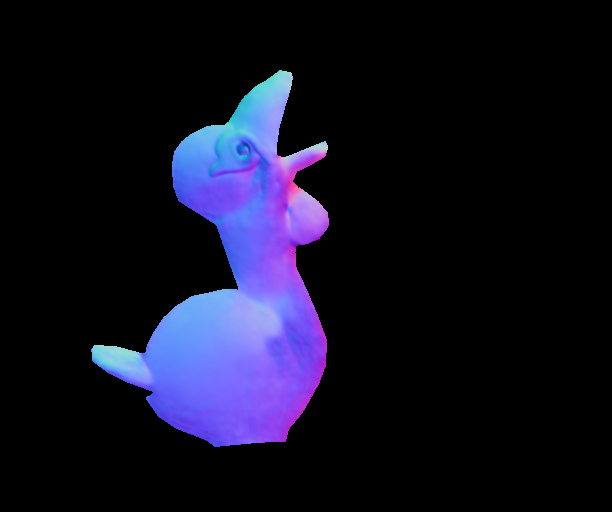}};
    \showerror{$23.8^\circ$/$22.0^\circ$}

}
\end{tikzpicture}
}
    \vspace{-1.7em}
    \caption{Shape reconstruction results of real-world objects and comparisons with learning-based methods using polarization. The numbers show the mean and median of depth errors in millimeters and surface normal angular errors in degrees. Our method robustly reconstructs shapes regardless of surface textures and materials in a single shot.}
    \vspace{-1.4em}
    \label{fig: reconstructed shape RGB}
\end{figure}

\section{Experimental Results}
We experimentally validate the effectiveness of our method on real data.
Our polarization projector consists of a white LED telecentric illuminator with a polarizer, a spatial light modulator with a resolution of 1024×768 (HOLOEYE LC2012), and convex lenses.
We capture polarimetric images with a commercial RGB polarimetric camera (Lucid TRI050S-QC).

For RGB reconstruction, the green channel is used for shape reconstruction and the Mueller matrix elements in \cref{eq:Mueller12,eq:Mueller0} are computed for each channel.
All color channels are used in \cref{eq:specular optimization} to optimize color-independent material parameters.
Diffuse albedo is obtained from \cref{eq:diffuse optimization} for each color channel.

Line widths of the stripe patterns are set to $12$ pixels on the polarization projector for the single pattern and 12 shifted patterns.
We set $n_\mathrm{db} = 4$ and $k_\mathrm{db} = 7$, and quantize the AoLP values at even intervals from $0^\circ$ to $80^\circ$.
For dense visualization of the results, the depths and the optimized BRDF parameters of the sparse single-shot reconstruction are interpolated by a radial basis function.

\begin{figure}
    \centering
    \begin{tikzpicture}[x=0.001\linewidth,y=0.001\linewidth,every node/.style={inner sep=0pt, text depth=0pt}]
{
    \newcount\vX
    \newcount\vY
    \newcount\vdX
    \newcount\vdXX
    \newcount\vdXXX
    \newcount\vdY
    \vX = 0
    \vY = 0
    \vdX = 130
    \vdXX = 85
    \vdXXX = 88
    \def\mysz{0.132\linewidth}

    \def\myfont{\footnotesize}
    \def\myfonth{\scriptsize}
    \advance\vX by \vdXXX
    \node[inner sep=0pt, font=\myfont] (a) at (\vX,\vY) {Pol. Image};
    \advance\vX by \vdXX
    \advance\vX by \vdXXX
    \node[inner sep=0pt, font=\myfont] (a) at (\vX,\vY) {Specular};
    \advance\vX by \vdX
    \node[inner sep=0pt, font=\myfont] (a) at (\vX,\vY) {Diffuse};
    \advance\vX by \vdX
    \node[inner sep=0pt, font=\myfont] (a) at (\vX,\vY) {$m^d_{10}$};
    \advance\vX by \vdX
    \node[inner sep=0pt, font=\myfont] (a) at (\vX,\vY) {$m^d_{20}$};
    \advance\vX by \vdX
    \node[inner sep=0pt, font=\myfont] (a) at (\vX,\vY) {$\rho^d_L$};
    \advance\vX by \vdX
    \node[inner sep=0pt, font=\myfont] (a) at (\vX,\vY) {$\phi^d_L$};

    \vdY = -94
    \vX = 0
    \advance\vY by \vdY
    \def\myincA#1{\resizebox{\mysz}{!}{\adjincludegraphics[Clip={{0.23\width} {0.18\width} {0.28\width} {0.18\width}}]{#1}}}
    \node[inner sep=0pt, font=\myfonth, rotate=90] (a) at (\vX,\vY) {Intensity};
    \advance\vX by \vdXXX
    \node[inner sep=0pt] (a) at (\vX,\vY) {\myincA{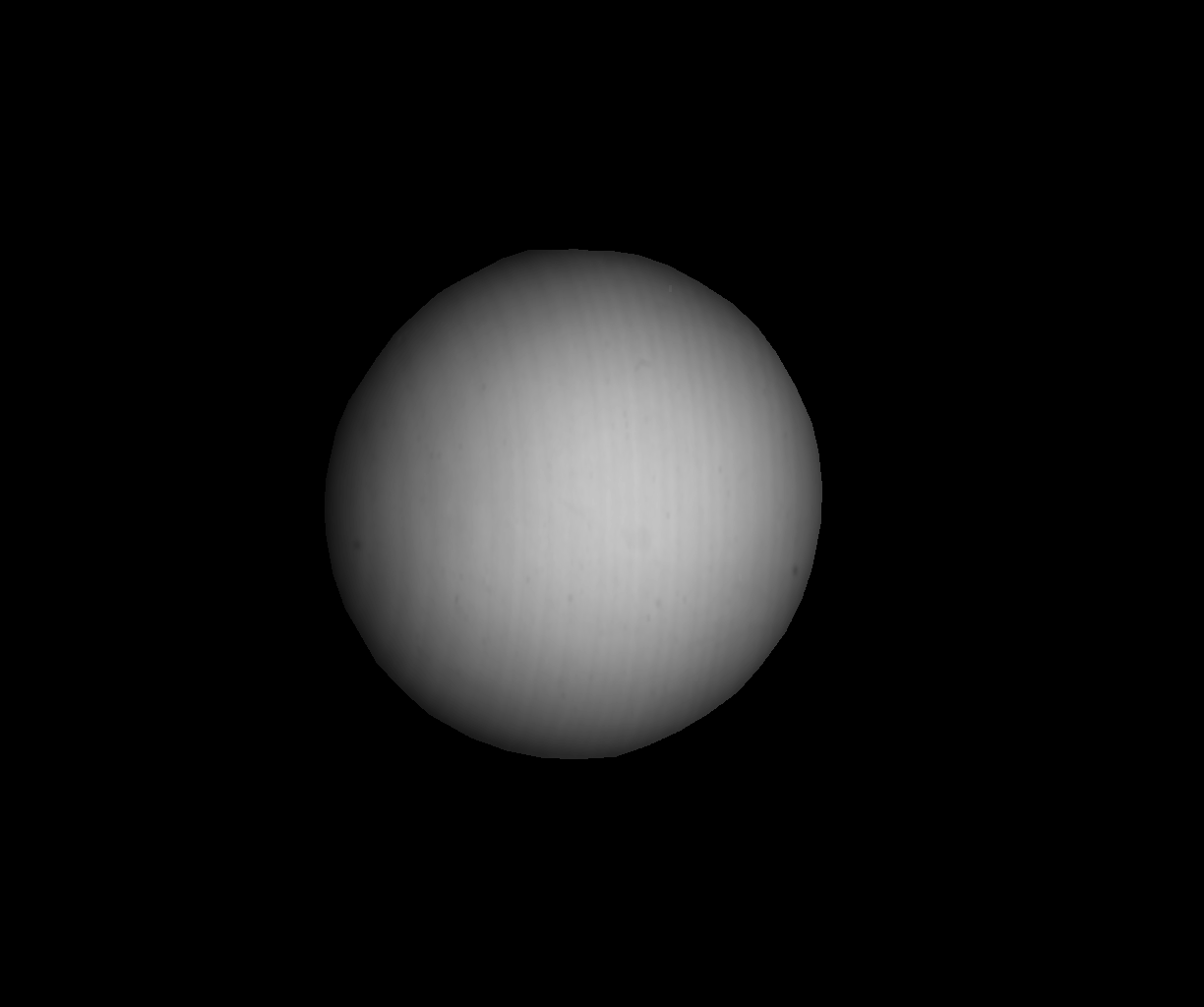}};
    \advance\vX by \vdXX
    \node[inner sep=0pt, font=\myfonth, rotate=90] (a) at (\vX,\vY) {GT};
    \advance\vX by \vdXXX
    \node[inner sep=0pt] (a) at (\vX,\vY) {\myincA{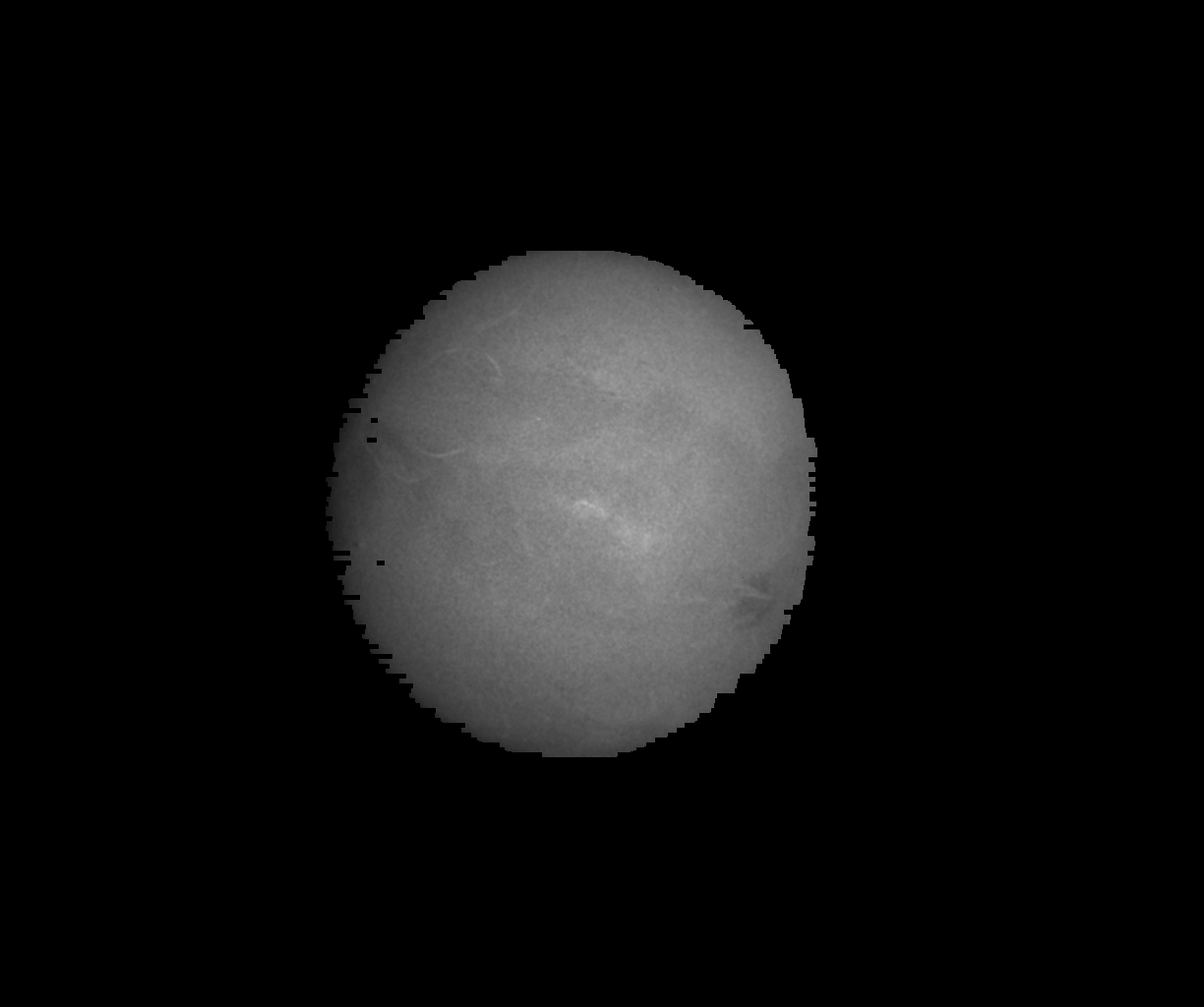}};
    \advance\vX by \vdX
    \node[inner sep=0pt] (a) at (\vX,\vY) {\myincA{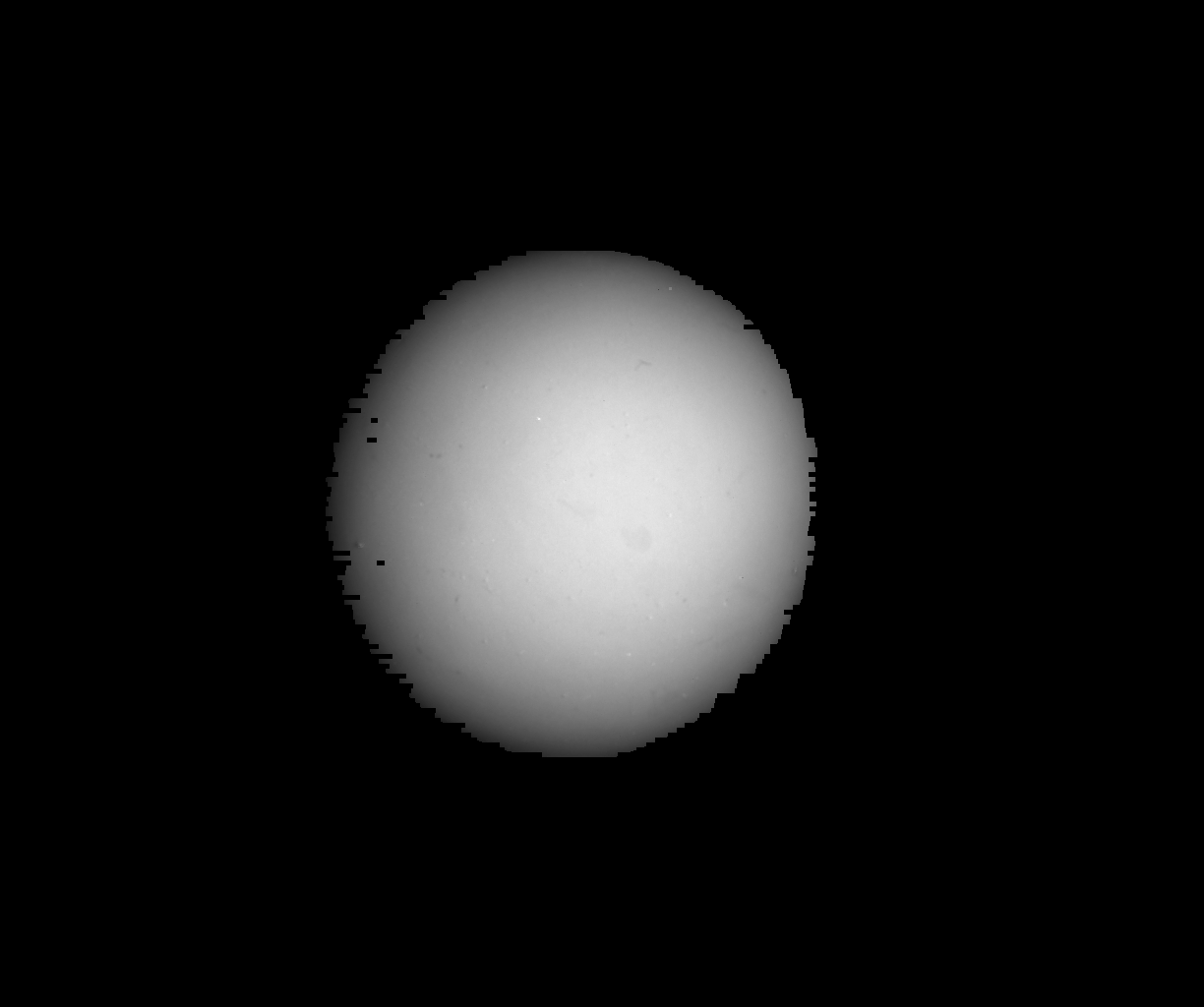}};
    \advance\vX by \vdX
    \node[inner sep=0pt] (a) at (\vX,\vY) {\myincA{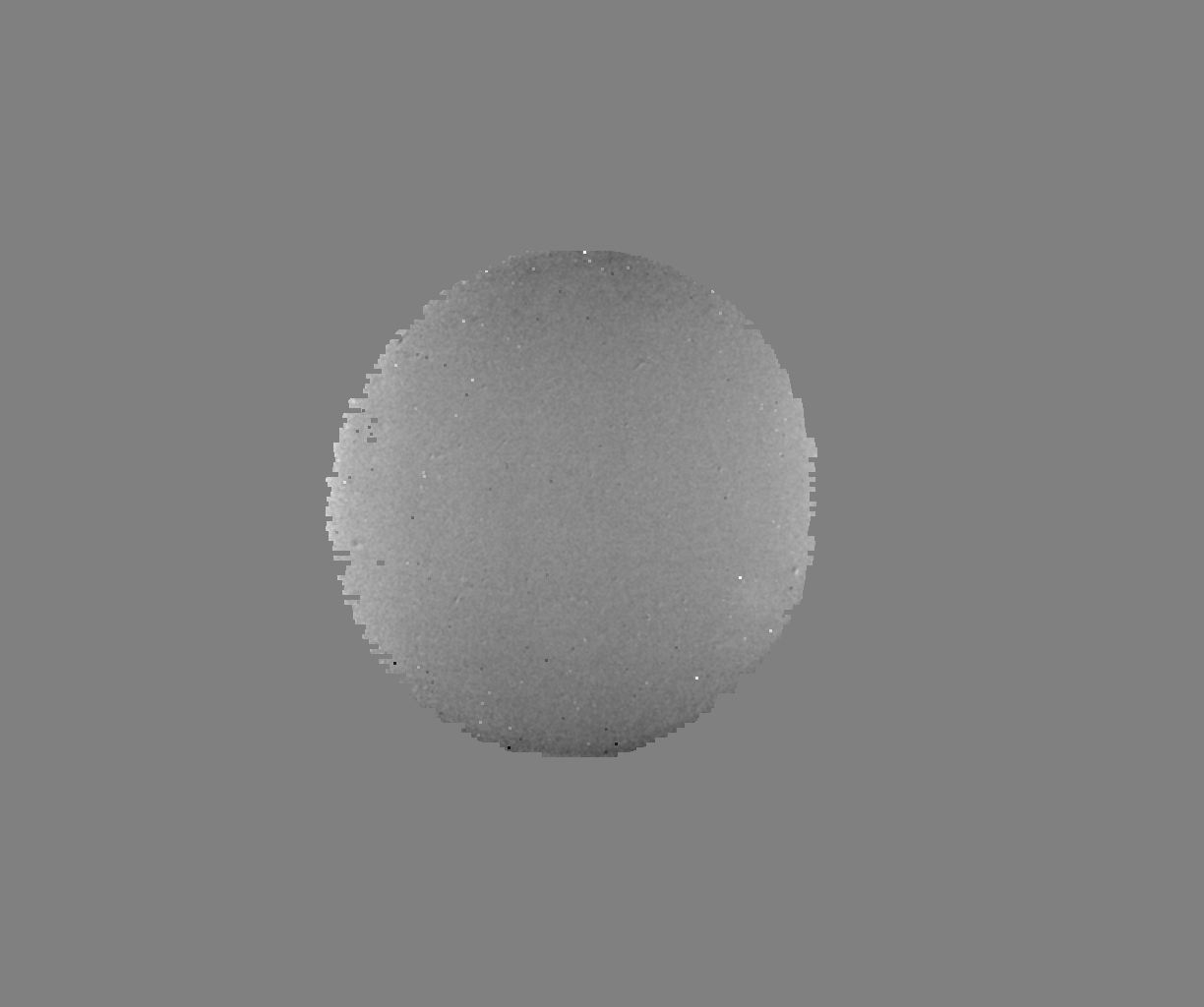}};
    \advance\vX by \vdX
    \node[inner sep=0pt] (a) at (\vX,\vY) {\myincA{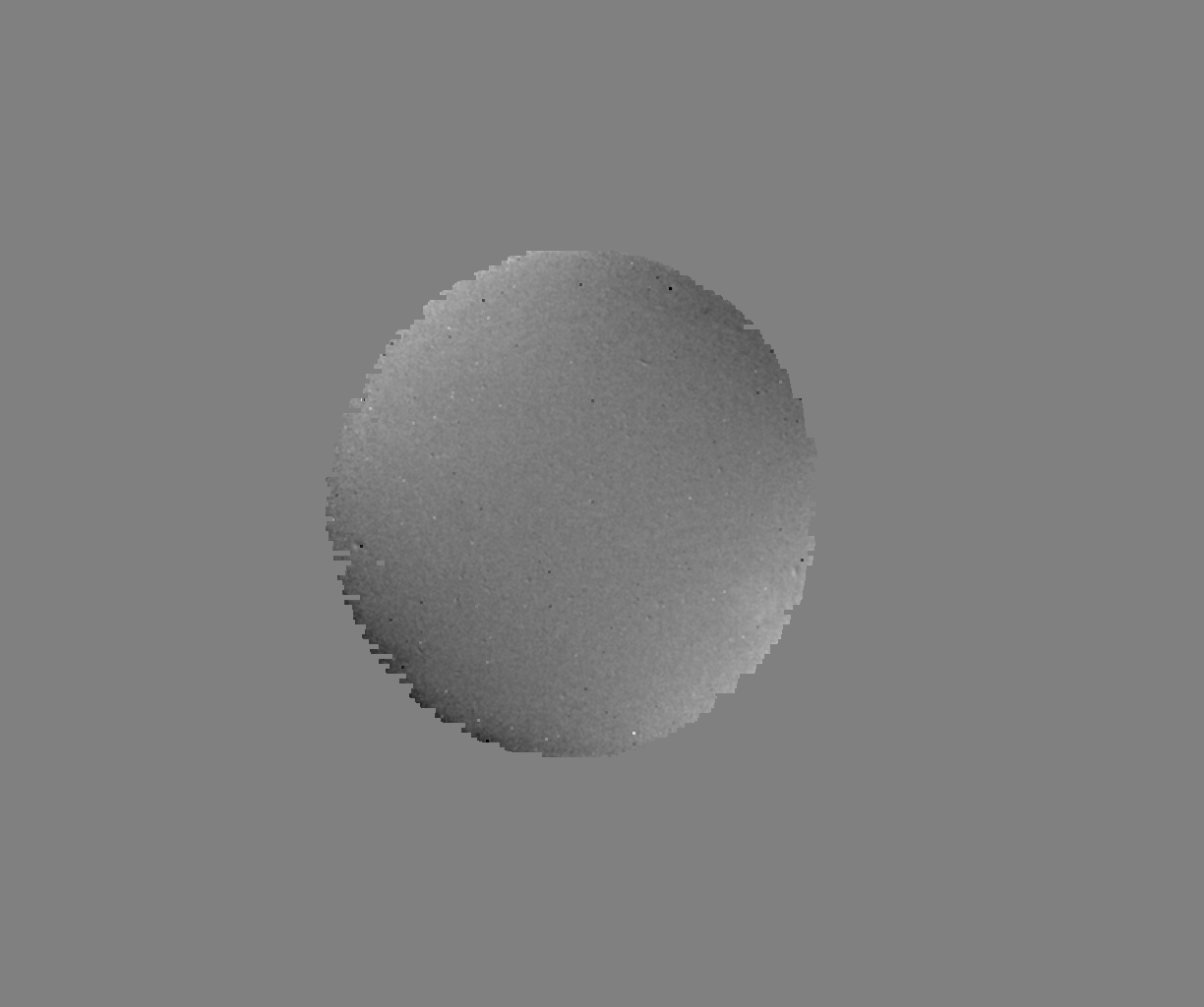}};
    \advance\vX by \vdX
    \node[inner sep=0pt] (a) at (\vX,\vY) {\myincA{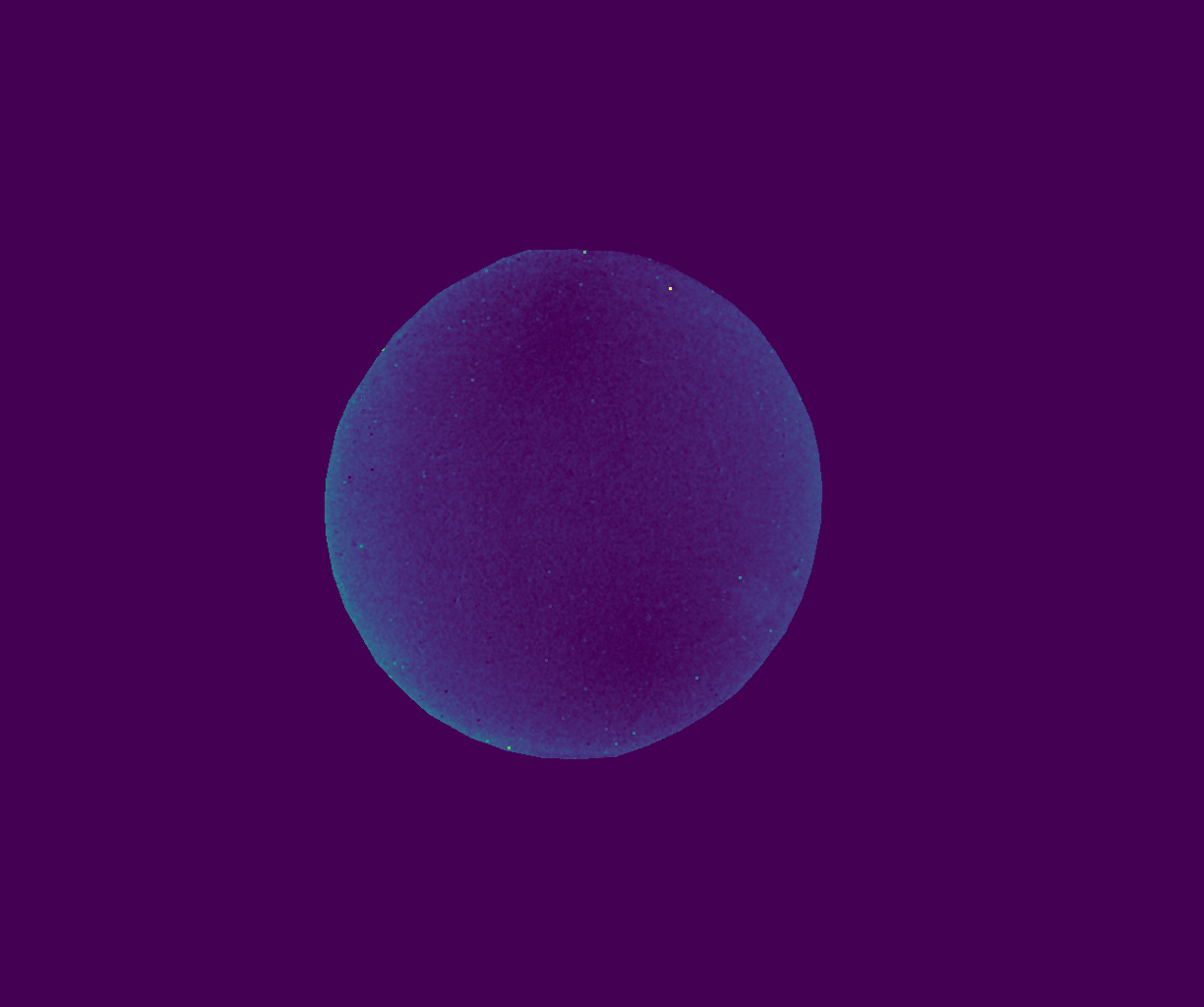}};
    \advance\vX by \vdX
    \node[inner sep=0pt] (a) at (\vX,\vY) {\myincA{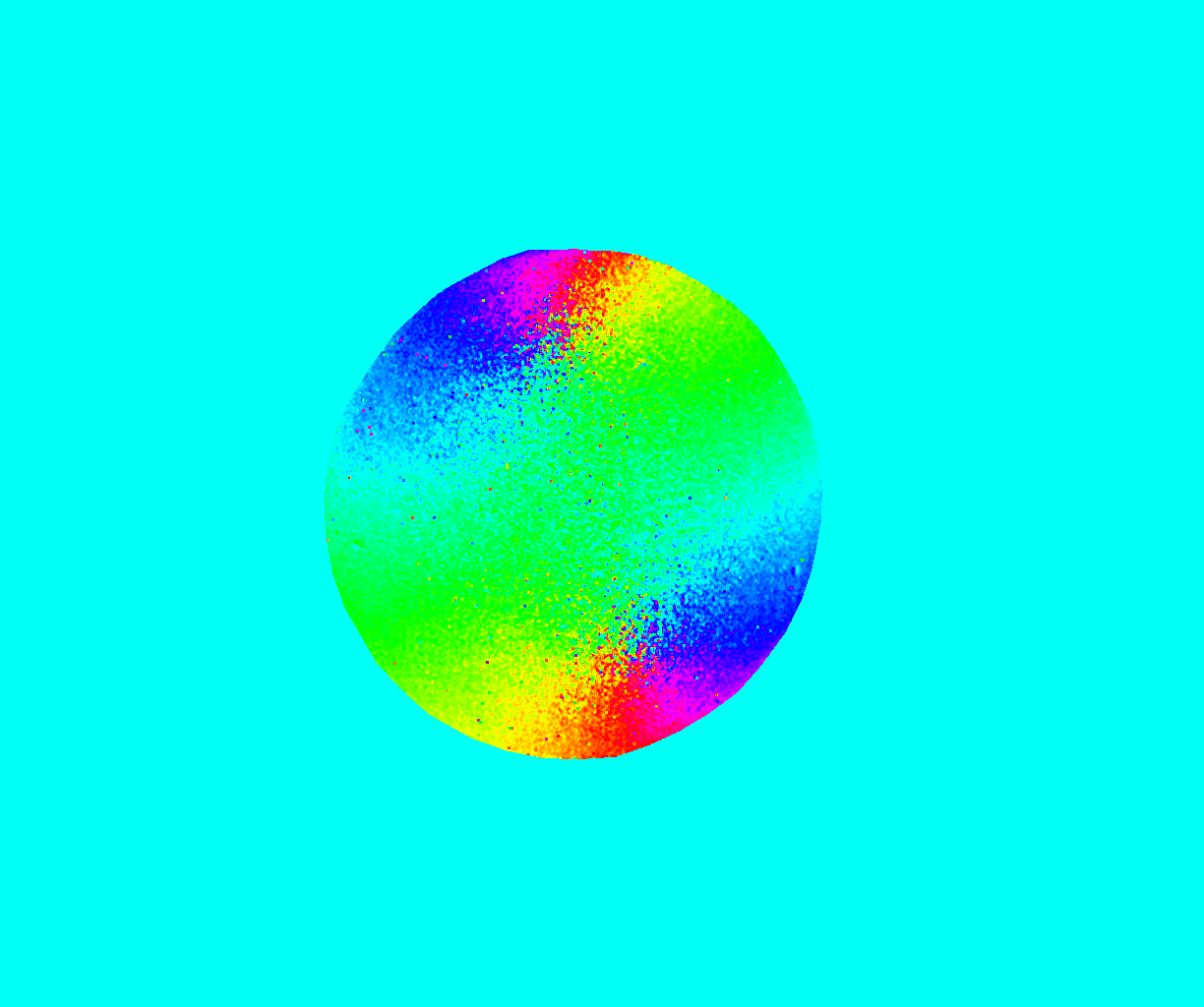}};

    \vdY = -123
    \vX = 0
    \advance\vY by \vdY
    \node[inner sep=0pt, font=\myfonth, rotate=90] (a) at (\vX,\vY) {DoLP};
    \advance\vX by \vdXXX
    \node[inner sep=0pt] (a) at (\vX,\vY) {\myincA{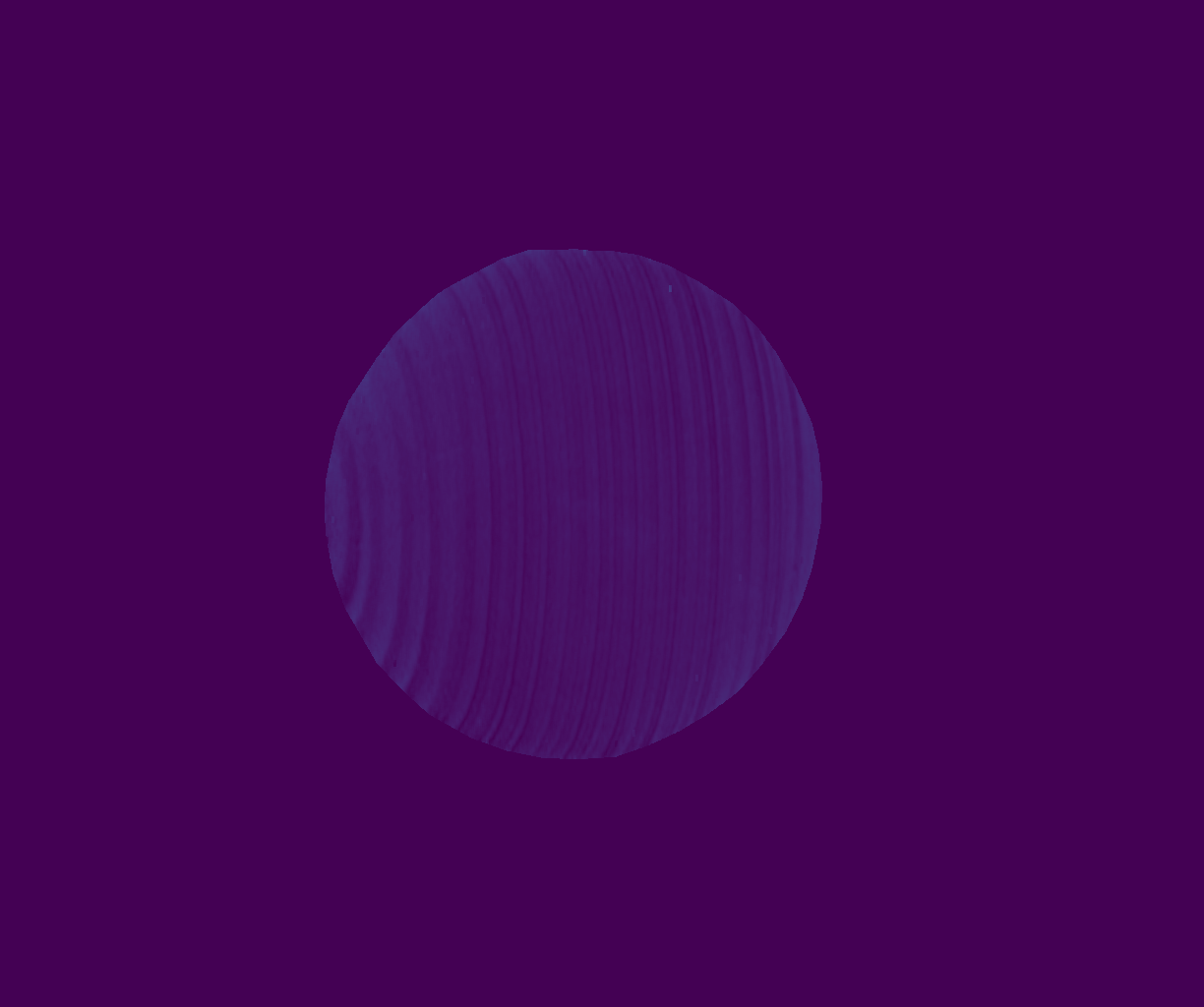}};
    \advance\vX by \vdXX
    \node[inner sep=0pt, font=\myfonth, rotate=90] (a) at (\vX,\vY) {Ours};
    \advance\vX by \vdXXX
    \node[inner sep=0pt] (a) at (\vX,\vY) {\myincA{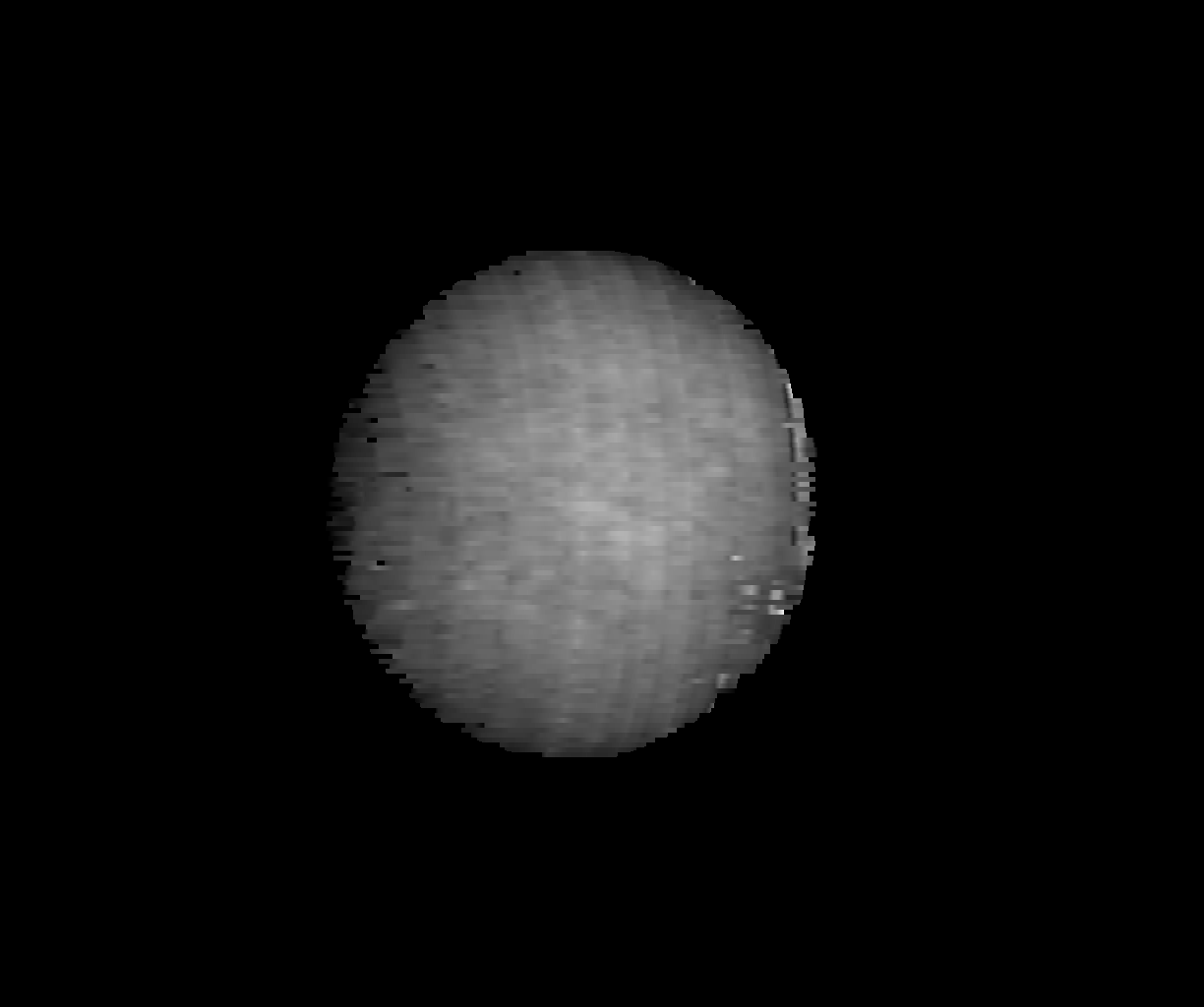}};
    \advance\vX by \vdX
    \node[inner sep=0pt] (a) at (\vX,\vY) {\myincA{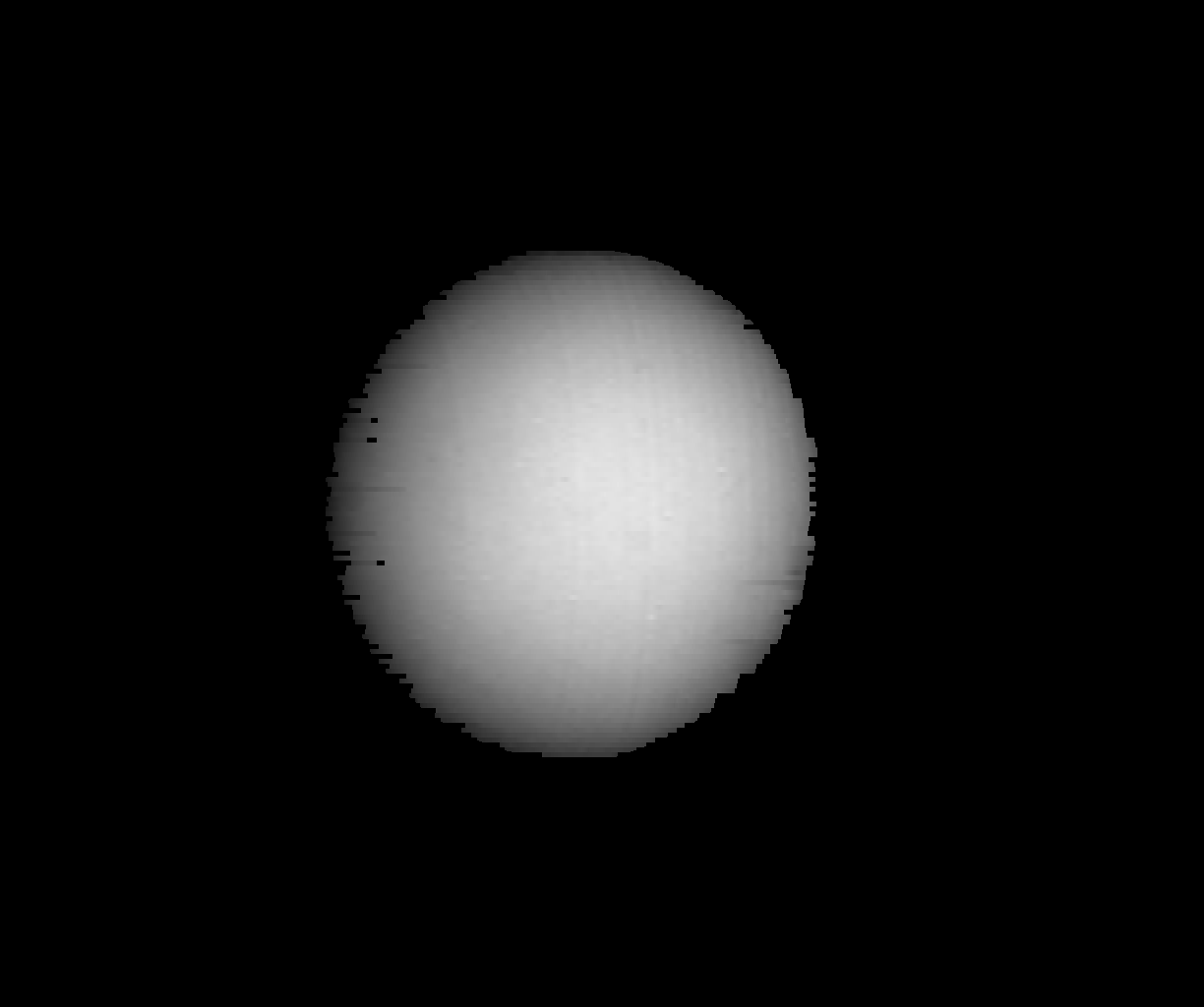}};
    \advance\vX by \vdX
    \node[inner sep=0pt] (a) at (\vX,\vY) {\myincA{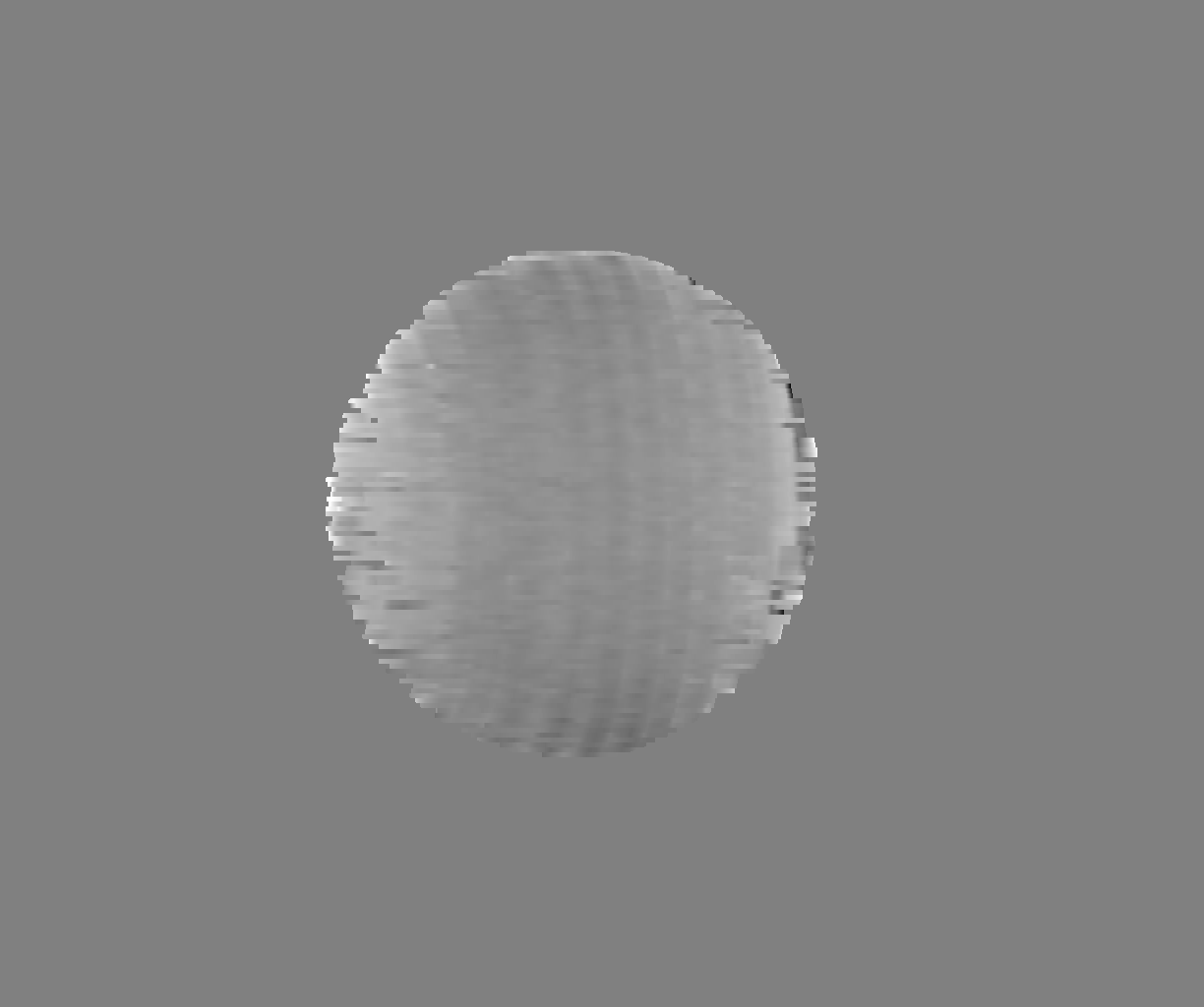}};
    \advance\vX by \vdX
    \node[inner sep=0pt] (a) at (\vX,\vY) {\myincA{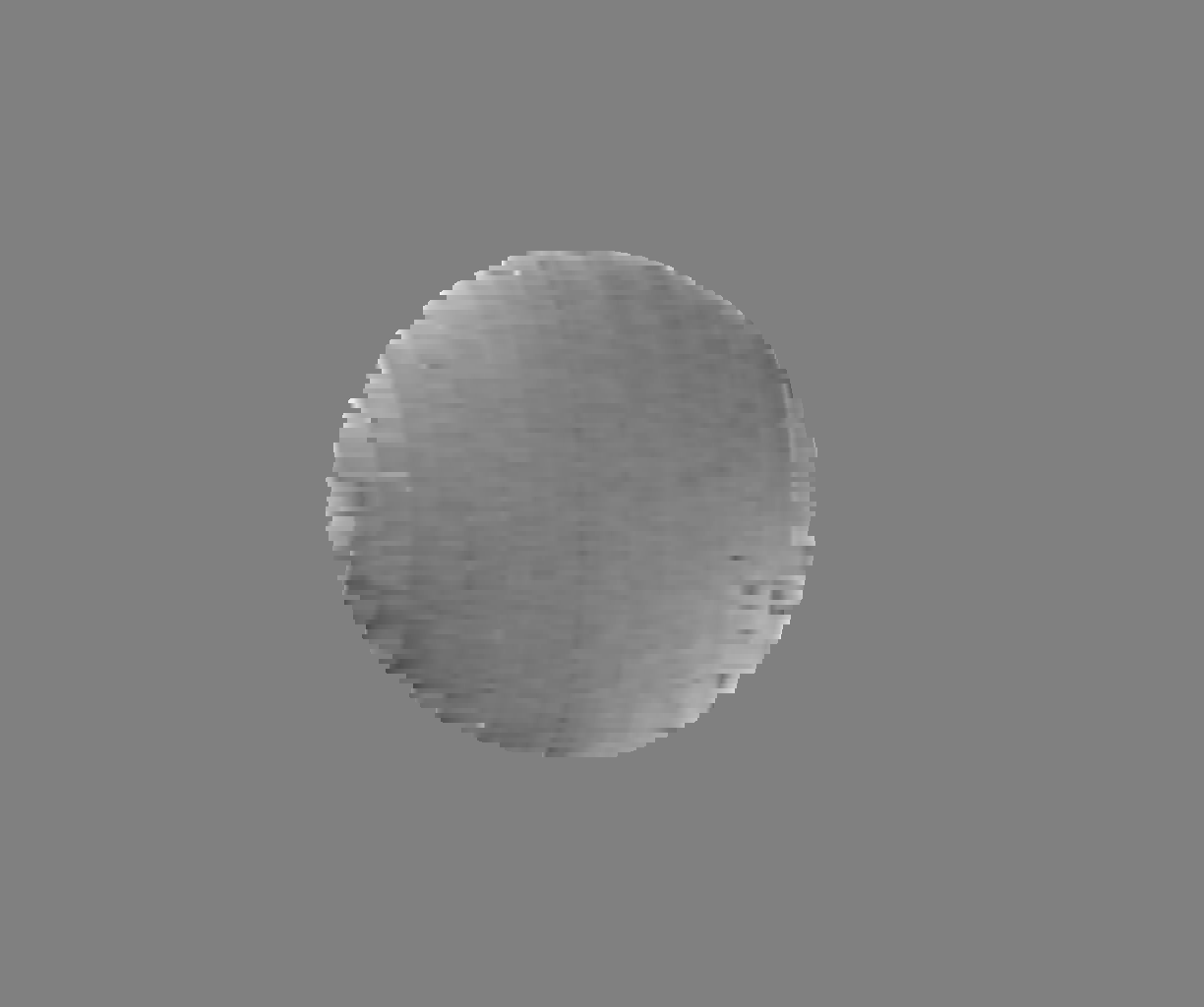}};
    \advance\vX by \vdX
    \node[inner sep=0pt] (a) at (\vX,\vY) {\myincA{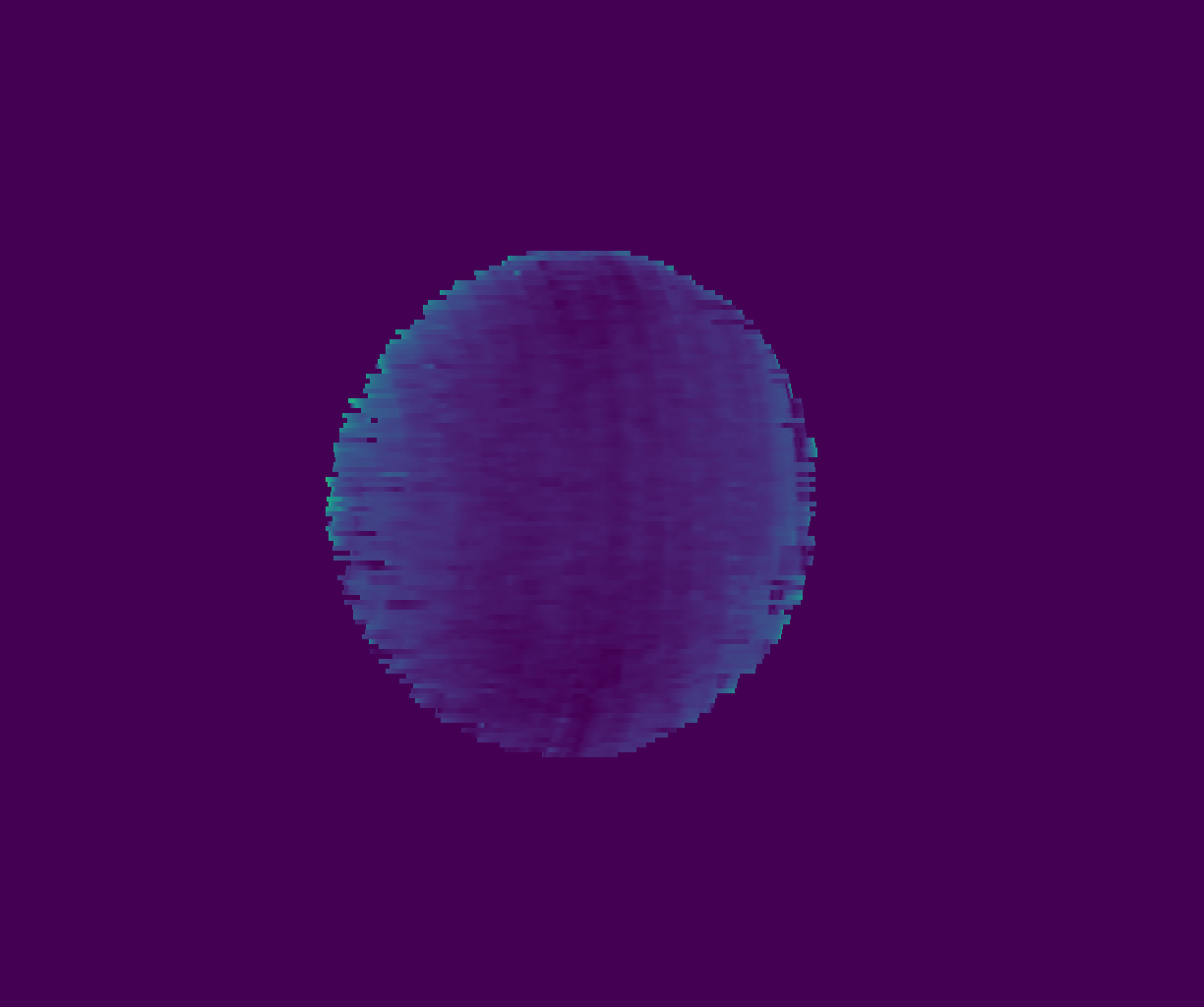}};
    \advance\vX by \vdX
    \node[inner sep=0pt] (a) at (\vX,\vY) {\myincA{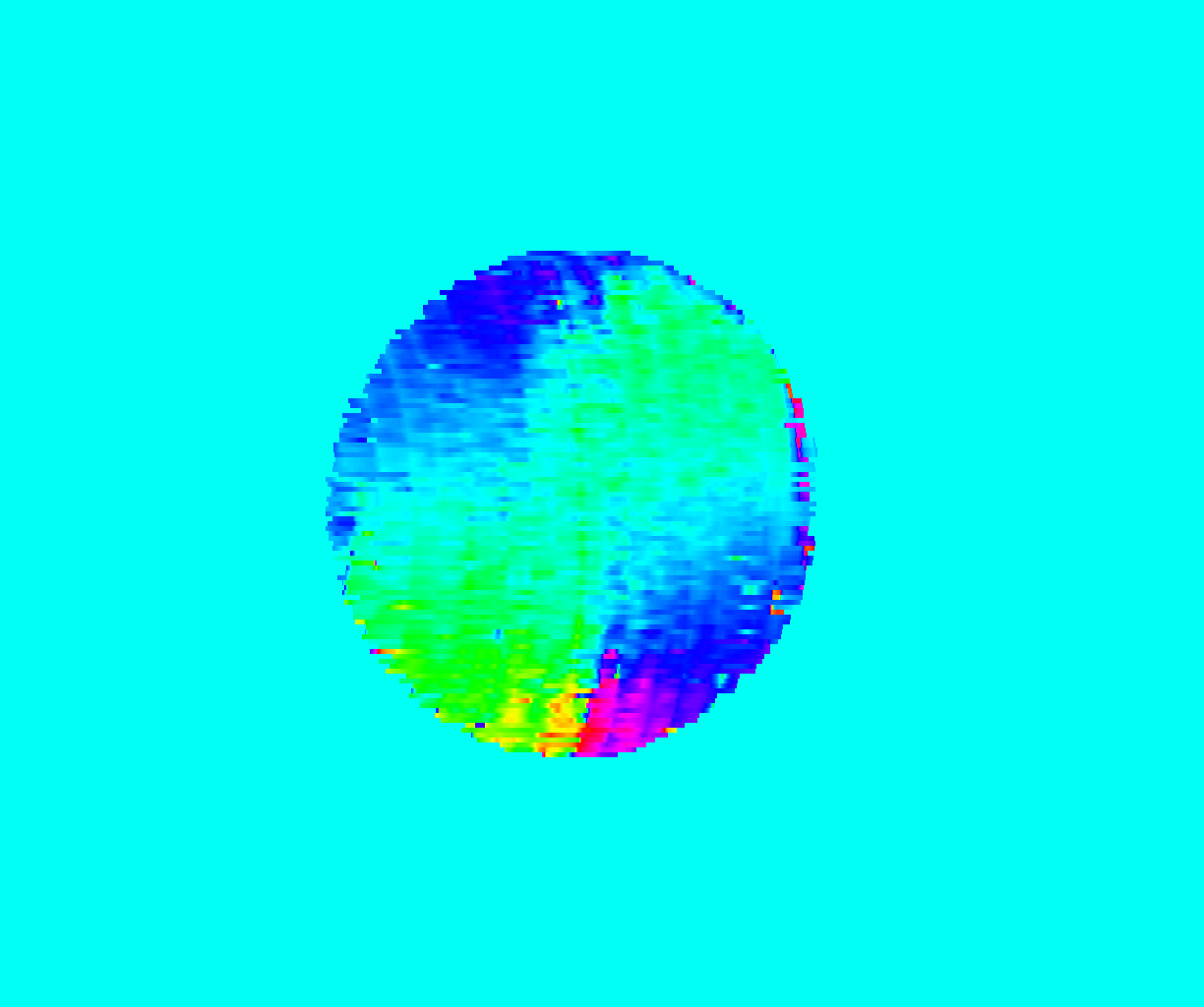}};

    \vdY = -123
    \vX = 0
    \advance\vY by \vdY
    \node[inner sep=0pt, font=\myfonth, rotate=90] (a) at (\vX,\vY) {AoLP};
    \advance\vX by \vdXXX
    \node[inner sep=0pt] (a) at (\vX,\vY) {\myincA{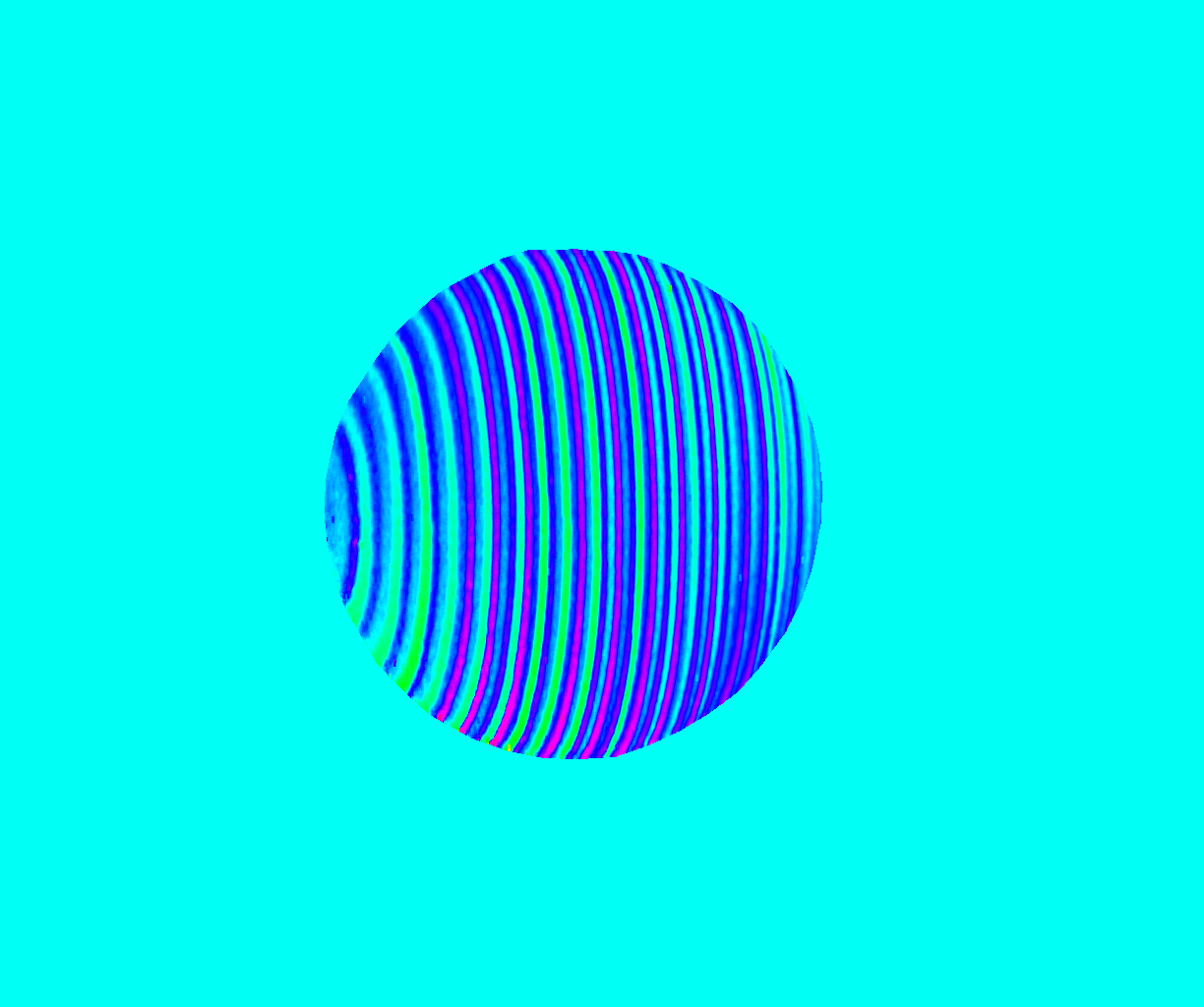}};
    \advance\vX by \vdXX
    \node[inner sep=0pt, text height=1ex,font=\myfonth, rotate=90] (a) at (\vX,\vY) { Unpol. diff.};
    \advance\vX by \vdXXX
    \node[inner sep=0pt] (a) at (\vX,\vY) {\myincA{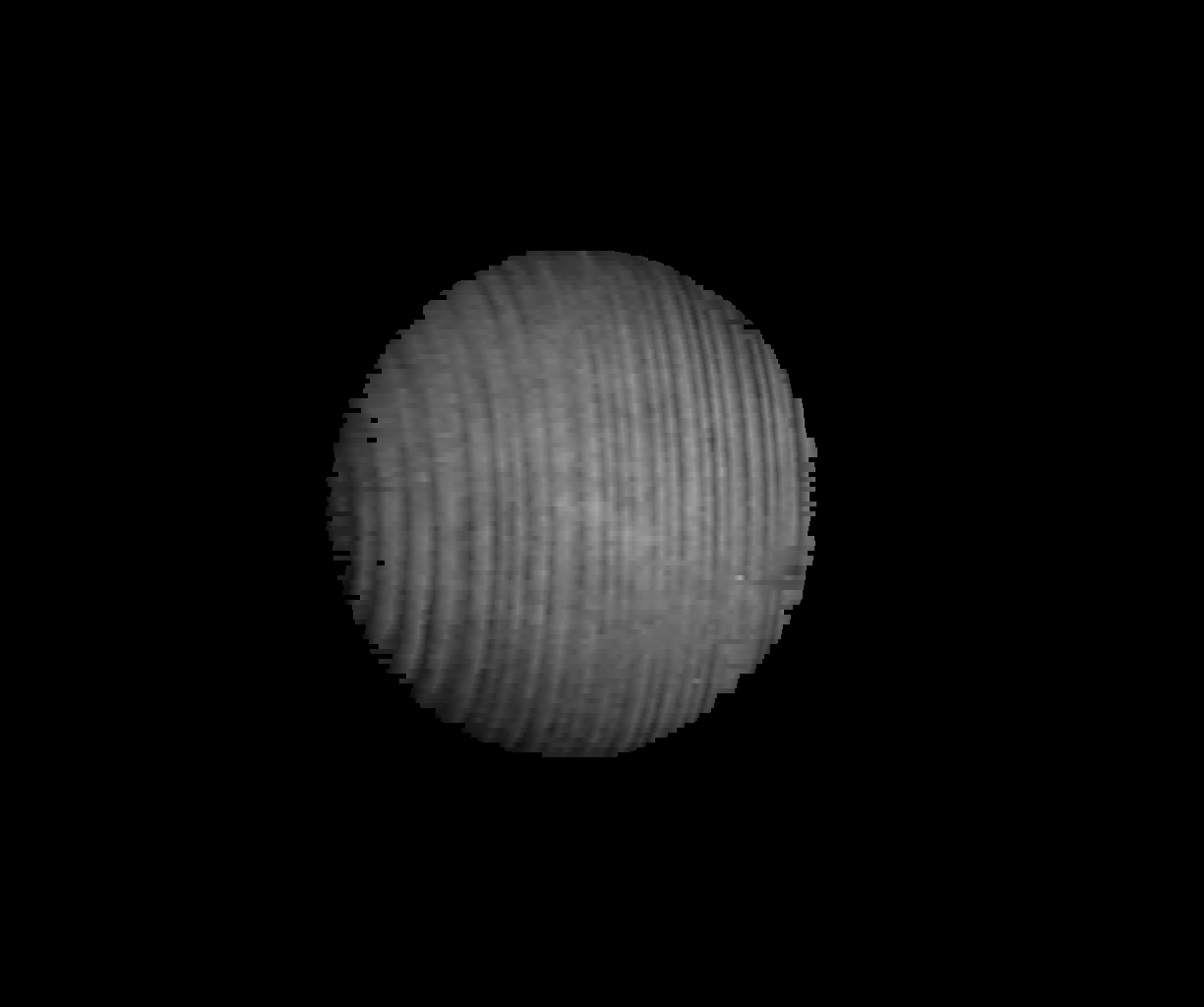}};
    \advance\vX by \vdX
    \node[inner sep=0pt] (a) at (\vX,\vY) {\myincA{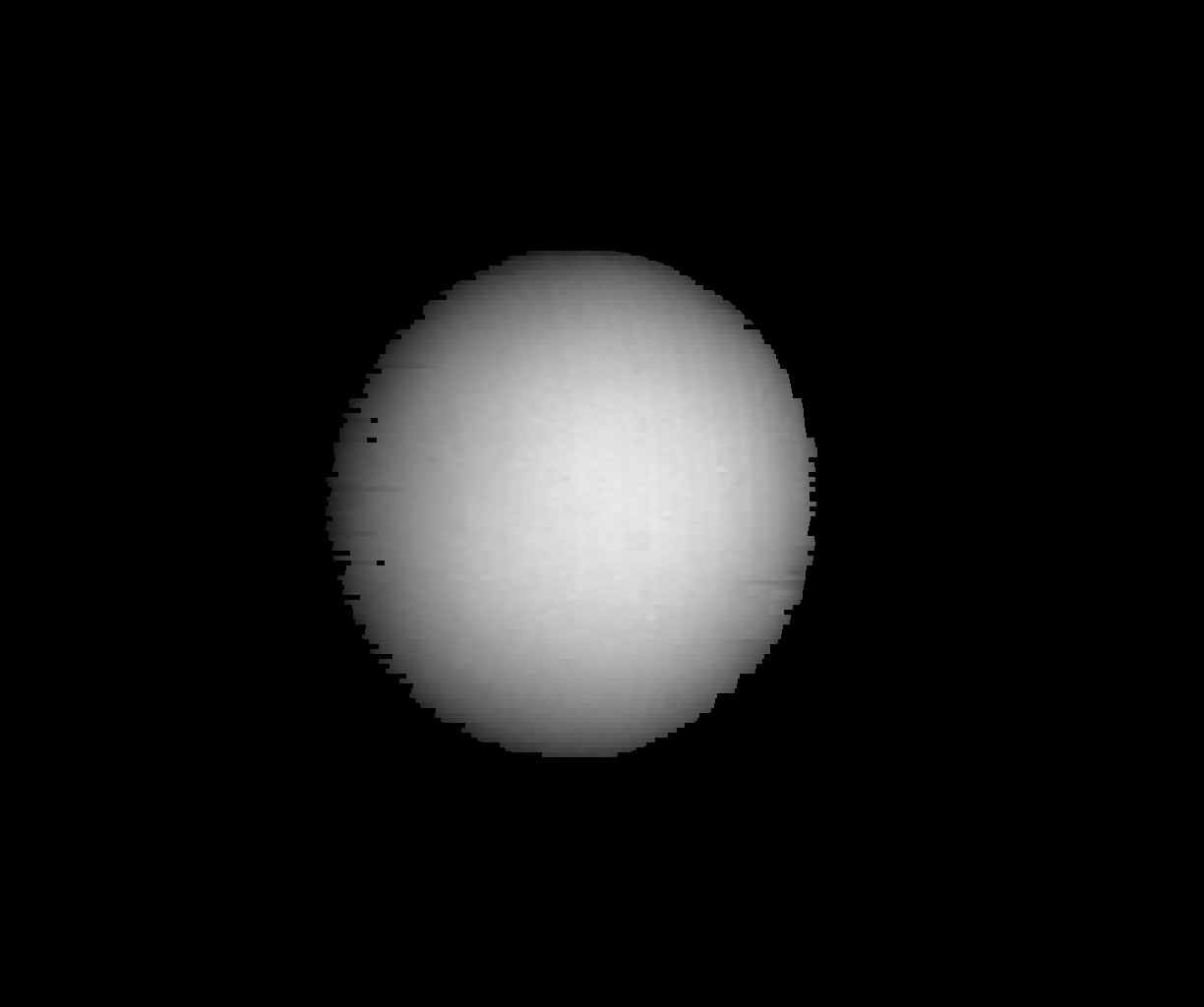}};

}
\end{tikzpicture}
    \vspace{-0.7em}
    \caption{Decomposition results of a polarimetric reflection. Assuming unpolarized diffuse reflection causes artifacts in the separated specular component. In contrast, our method accurately decomposes polarimetric reflection in a single shot.}
    \vspace{-0.5em}
    \label{fig: result decomposition}
\end{figure}

\begin{figure}[t]
    \centering
    \input{fig_tex/comparison_RGB_relighting}
    \vspace{-0.5em}
    \caption{Relighting results using shape and BRDF. The Lambertian sphere indicates the direction of light. We can relight real-world objects in a single shot by fully exploiting polarimetric reflections of the SPM pattern.}
    \vspace{-1.4em}
    \label{fig: relighting RGB}
\end{figure}

\subsection{Shape Reconstruction}
\Cref{fig: reconstructed shape RGB} shows shape reconstruction results on real objects.
We obtain the ground truth shapes by structured light with Gray code and phase-shifting patterns.
The numbers below each depth map show the mean and median of depth errors and those below each nomral map show the angular errors.
The results show that our method can measure accurate object geometry robustly from the single SPM pattern.

\subsection{Polarimetric Decomposition}
\Cref{fig: result decomposition} shows the results of the polarimetric specular and diffuse decomposition.
We compute the ground truth decomposition results from the ground truth Mueller matrix, following \cref{eq: specular and diffuse separation}.
The ground truth Mueller matrix is obtained from 26 polarimetric images while projecting different uniform polarization patterns, and by solving \cref{eq:Mueller matrix}.
Unpolarized diffuse reflection is often assumed for single-shot diffuse and specular separation~\cite{wolff2002constraining}.
This assumption, however, leads to inaccurate specular reflection as diffuse reflection is, in reality, partially polarized. 
In contrast, our method can successfully decompose a single polarimetric image into polarimetric diffuse and specular reflections by exploiting the spatially multiplexed polarization.

\begin{figure}[t]
    \centering
    \input{fig_tex/static_scene}
    \vspace{-0.6em}
    \caption{High-resolution reconstruction of static scenes by using shifted SPM patterns. The shifted SPM enables shape and reflectance sensing of detailed surface texture.}
    \label{fig: static scene}
    \vspace{-1.5em}
\end{figure}

\subsection{Relighting}

\Cref{fig: relighting RGB} shows relighting results by using the reconstructed shape and BRDF.
Polarimetric specular and diffuse decomposition enables BRDF estimation from a single polarimetric image.
The relighting results show plausible surface color texture and specular highlight rendering under novel lighting conditions. Note again that past single-shot sensing methods alter the visual texture and thus cannot achieve this from the single capture. 

\subsection{Comparison with Learning-based Methods}
We compare our method with learning-based methods that use single RGB polarimetric images.
Deep Polarization Imaging~\cite{deschaintre2021deep} (DPI) and Shape from Polarization with Distant Lighting Estimation~\cite{lyu2023shape} (SfPDLE) estimate a normal map and SVBRDF. DPI also estimates a depth map.
We align the domain of illumination with their training data.
\Cref{fig: reconstructed shape RGB,fig: relighting RGB} shows the shape reconstructions and the RGB relighting results.
The scale of the depth map of DPI is adjusted for comparison.
Learning-based methods suffer from heavy reliance on learned priors which manifests into strong biases appearing as distortions of the recovered shape. Our SPM method realizes actual sensing of accurate shape and reflectance of the actual shape without any reliance on pre-learned priors, which is critical for real-world unseen targets.

\subsection{Dynamic Surface Reconstruction}

We apply our method to a dynamic surface by capturing continuous frames and using each frame.
\Cref{fig: opening figure} shows the shape reconstruction and relighting results of a pushed squeeze toy and a fake loaf expanding after compression.
Our method can capture changes in both shape and reflectance and relight the dynamic deformable surface at each frame.

\begin{figure}[t]
    \centering
    \input{fig_tex/adaptive_resolution}
    \vspace{-0.5em}
    \caption{Adaptive resolution reconstruction of a scene with both dynamic and static frames by using shifted SPM patterns. A static clay figure is reconstructed in high resolution. A camouflage-printed rubber duck is reconstructed in low resolution when moving and in high resolution after stopping.}
    \vspace{-1.5em}
    \label{fig: adaptive resolution}
\end{figure}

\subsection{Adaptive High Resolution Sensing}

\Cref{fig: static scene} shows high-resolution reconstruction results of a static scene.
Our method for shifted patterns can capture the detailed shape and texture at the resolution of the polarimetric camera.
\Cref{fig: adaptive resolution} demonstrates the adaptive resolution reconstruction of a dynamic scene.
Our method simultaneously reconstructs both dynamic and static regions at adaptive resolutions for each region and frame.

\section{Conclusion}\vspace{-0.5em}
We introduced a novel method for direct, single-shot, simultaneous sensing of shape and reflectance that is invisible to the naked eye. This is achieved by introducing spatial polarimetric multiplexing with quantized AoLPs and a constrained de Bruijn sequence and decoding of it for polarimetric reflectance separation and BRDF recovery. We further showed that shifted SPM patterns can be leveraged for adaptive and high-resolution reconstruction of dynamic and static scenes, regardless of surface texture and materials. 
We believe our spatial polarization multiplexing opens a new avenue of applications of single-shot shape and reflectance sensing thanks to its robustness to color, texture, and materials, applicability to both static and dynamic surfaces, and invisibility to human eyes. We are hopeful that it will find use in medicine (\eg, real-time surgery assistance), VR/AR (\eg, human-object interaction), robotics (especially for human-robot collaboration), and, of course, computer vision.  One limitation we plan to overcome in our immediate future work is the requirement of absence of environmental light.
We show that our SPM can provide some cues under environmental light in the supplementary material.
Achieving the same in broad daylight will further expand the horizon of applications. 

\vspace{-8pt}
\paragraph*{Acknowledgement}
This work was in part supported by
JSPS KAKENHI 
21H04893, 
23K28110,  
24H00742, 
23KJ1367; 
and JST JPMJAP2305. 

{
    \small
    \bibliographystyle{ieeenat_fullname}
    \bibliography{main}
}

\clearpage
\setcounter{page}{1}
\maketitlesupplementary

\appendix
\def\thesection{A.\arabic{section}}
\renewcommand\thefigure{A.\arabic{figure}}
\renewcommand\thetable{A.\arabic{table}}
\renewcommand\theequation{A.\arabic{equation}}
\setcounter{figure}{0}
\setcounter{table}{0}
\setcounter{equation}{0}

\begin{figure}[t]
    \centering
    \input{fig_tex_supp/de_Bruijn_graph}
    \caption{The diagram of the graph that generates the de Bruijn sequence. Each node represents a substring of $\ndb-1$ length. Each path represents the transition of the substrings of $\ndb$ length. The right table shows the possible symbols for each position. We impose constraints on the de Bruijn sequence to prevent adjacent stripes with close AoLP values.}
    \label{fig: de Bruijn graph}
\end{figure}

\begin{figure}[t]
    \centering
    \input{fig_tex_supp/AoLP_modulation}
    \vspace{-0.5em}
    \caption{\textbf{Left:} DoLP of diffuse reflection modeled as Fresnel transmission for the zenith angle of the surface normal. The DoLP is much lower than $1.0$ for most zenith angles. \textbf{Right:} AoLP modulation for $\frac{c^d}{c^s}$. Even when diffuse reflection is dominant, modulation of incident AoLP is within only a few degrees.}
    \vspace{-0.5em}
    \label{fig:diffuse DoLP and AoLP modulation}
\end{figure}

\section{Polarization Projector}
A polarization projector can control the polarization orientation at each pixel instead of the light intensity~\cite{ichikawa2024spiders,li2024fooling}.
The key component of the polarization projector is a Spatial Light Modulator (SLM) made of a Twisted Nematic (TN) liquid crystal.
The liquid crystal shows both fluidity and anisotropy, which allows us to change its anisotropy with external force.
The TN liquid crystal has perpendicularly aligned elongated molecules on each side, resulting in helically oriented molecules through its inside.
Such helical structure guides and changes the polarization orientation of incident light.
By applying the voltage between both sides of the liquid crystal, we can disturb its helical structure and weaken its guidance of polarization orientation, which enables us to control the rotation of polarization.

The SLM has a two-dimensional array of TN liquid crystals placed between electrodes.
By controlling the voltage between the electrodes of each cell, we can rotate the polarization orientation of incident polarized light at each pixel independently and project the polarization pattern.
Since the range of AoLP rotation is from $0^\circ$ to $90^\circ$ continuously, we can realize complex polarization patterns with the polarization projector.

\begin{figure}[t]
    \centering
    \resizebox{1.01\linewidth}{!}{
\begin{tikzpicture}[x=0.001\linewidth,y=0.001\linewidth,every node/.style={inner sep=0pt, text depth=0pt}]
{
    \newcount\vX
    \newcount\vY
    \newcount\vdX
    \newcount\vdXhalf
    \newcount\vdY
    \vX = 0
    \vY = 0
    \vdX = 245
    \def\mysz{0.250\linewidth}
    
    \def\myfont{\footnotesize}
    \node[inner sep=0pt, font=\myfont] (a) at (\vX,\vY) {Image};
    \advance\vX by \vdX
    \node[inner sep=0pt, font=\myfont] (a) at (\vX,\vY) {AoLP};
    \advance\vX by \vdX
    \node[inner sep=0pt, font=\myfont] (a) at (\vX,\vY) {GT depth};
    \advance\vX by \vdX
    \node[inner sep=0pt, font=\myfont] (a) at (\vX,\vY) {Ours};

    \vdY = -140
    \vX = 0
    \advance\vY by \vdY
    \def\myincA#1{\resizebox{\mysz}{!}{\adjincludegraphics[Clip={{0.0\width} {0.0\width} {0.0\width} {0.0\width}}]{#1}}}
    \node[inner sep=0pt] (a) at (\vX,\vY) {\resizebox{\mysz}{!}{\myincA{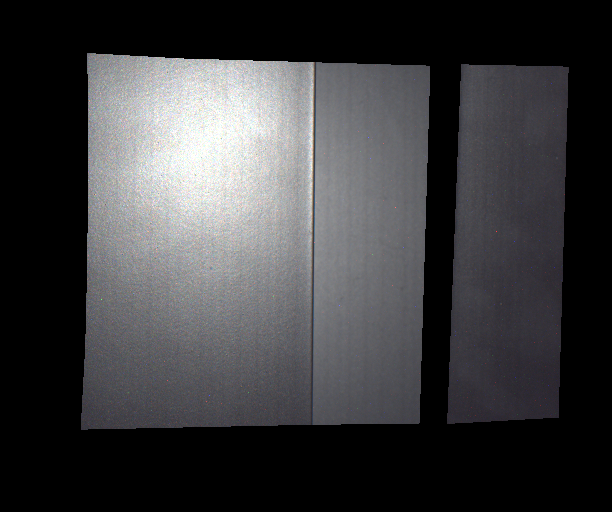}}};
    \advance\vX by \vdX
    \node[inner sep=0pt] (a) at (\vX,\vY) {\resizebox{\mysz}{!}{\myincA{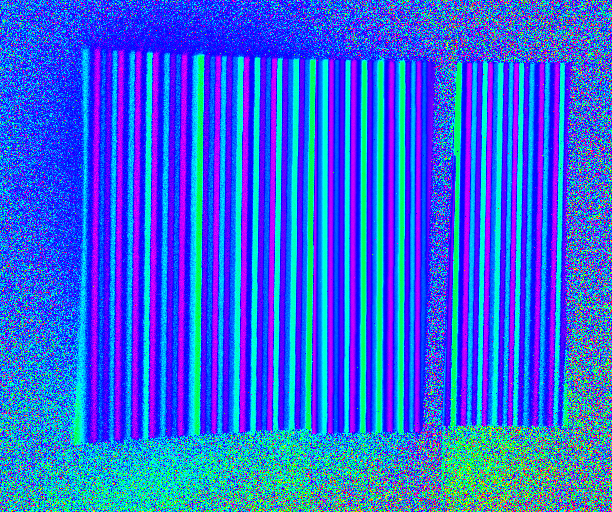}}};
    \advance\vX by \vdX
    \node[inner sep=0pt] (a) at (\vX,\vY) {\resizebox{\mysz}{!}{\myincA{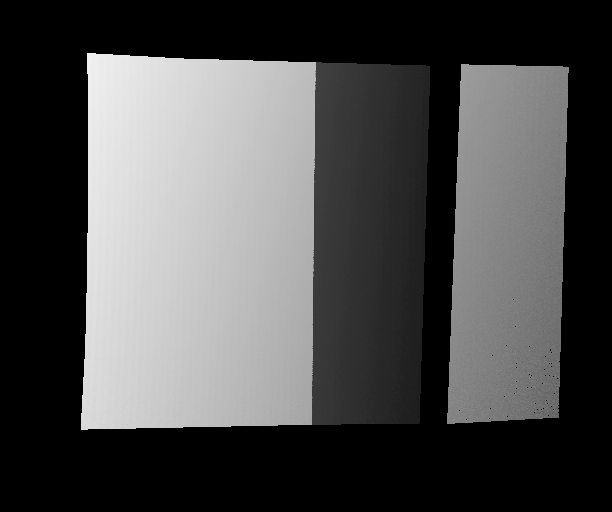}}};
    \advance\vX by \vdX
    \node[inner sep=0pt] (a) at (\vX,\vY) {\resizebox{\mysz}{!}{\myincA{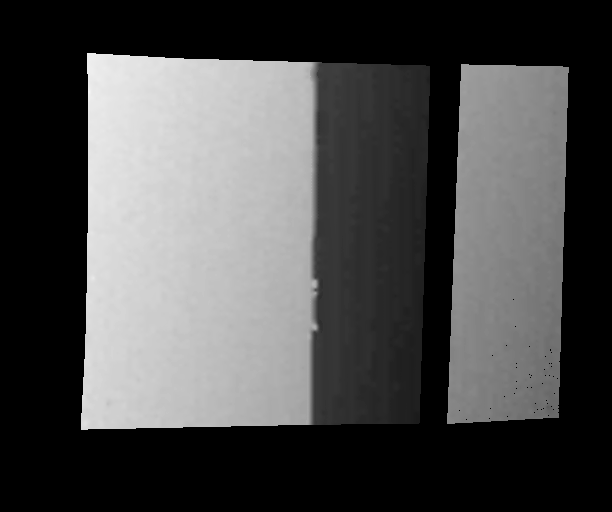}}};

}
\end{tikzpicture}
}
    \vspace{-1.0em}
    \caption{Depth reconstruction results of scenes with occlusion. Our method can accurately handle occlusion.}
    \vspace{-0.5em}
    \label{fig: occlusion}
\end{figure}

\begin{figure}[t]
    \centering
    \resizebox{1.01\linewidth}{!}{
\begin{tikzpicture}[x=0.001\linewidth,y=0.001\linewidth,every node/.style={inner sep=0pt, text depth=0pt}]
{
    \newcount\vX
    \newcount\vY
    \newcount\vdX
    \newcount\vdY
    \vX = 0
    \vY = 0
    \vdX = 195
    \def\mysz{0.200\linewidth}

    \def\myfont{\footnotesize}
    \node[inner sep=0pt, font=\myfont] (a) at (\vX,\vY) {Ambient light};
    \advance\vX by \vdX
    \node[inner sep=0pt, font=\myfont] (a) at (\vX,\vY) {AoLP};
    \advance\vX by \vdX
    \node[inner sep=0pt, font=\myfont] (a) at (\vX,\vY) {Depth};
    \advance\vX by \vdX
    \node[inner sep=0pt, font=\myfont] (a) at (\vX,\vY) {Relighting};

    \vdY = -120
    \vX = 0

    \advance\vY by \vdY
    \def\myincA#1{\resizebox{\mysz}{!}{\adjincludegraphics[Clip={{0.2\width} {0.15\width} {0.3\width} {0.2\width}}]{#1}}}
    \def\myincObjA#1{\resizebox{\mysz}{!}{\adjincludegraphics[Clip={{0.0\width} {0.03\width} {0.0\width} {0.0\width}}]{#1}}}
    \def\myincObjB#1{\resizebox{\mysz}{!}{\adjincludegraphics[Clip={{0.0\width} {0.03\width} {0.0\width} {0.0\width}}]{#1}}}

    \node[inner sep=0pt] (a) at (\vX,\vY) {\myincObjA{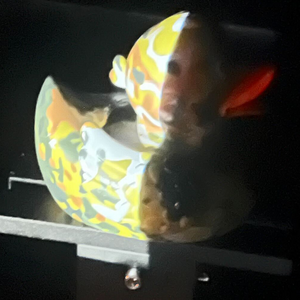}};
    \advance\vX by \vdX
    \node[inner sep=0pt] (a) at (\vX,\vY) {\myincA{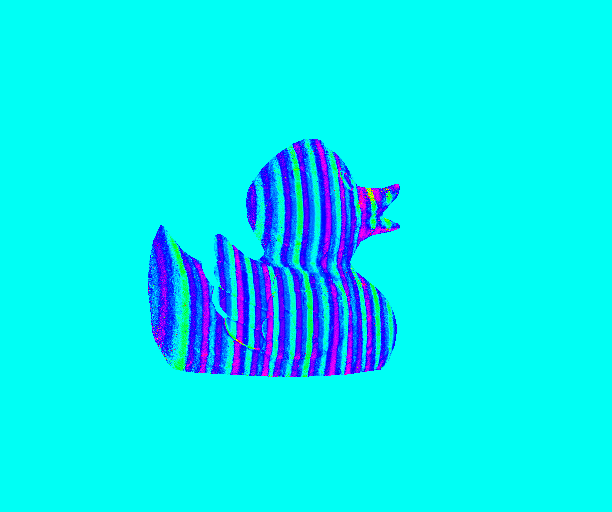}};
    \advance\vX by \vdX
    \node[inner sep=0pt] (a) at (\vX,\vY) {\myincA{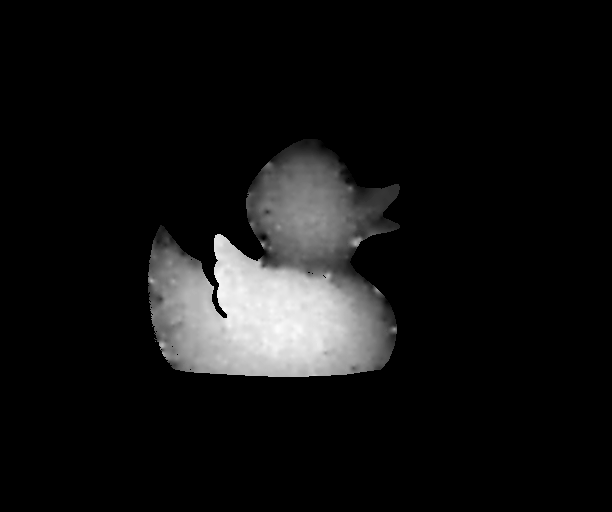}};
    \advance\vX by \vdX
    \node[inner sep=0pt] (a) at (\vX,\vY) {\myincA{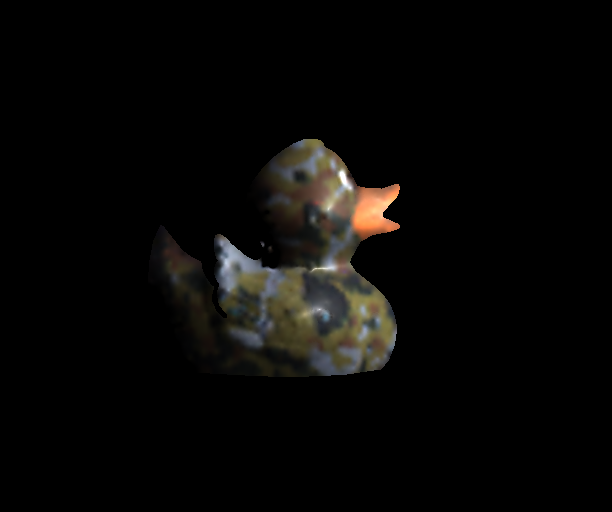}};

    \vdY = -190
    \vX = 0
    \advance\vY by \vdY
    \node[inner sep=0pt] (a) at (\vX,\vY) {\myincObjA{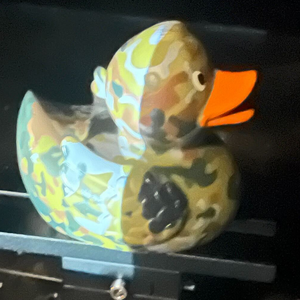}};
    \advance\vX by \vdX
    \node[inner sep=0pt] (a) at (\vX,\vY) {\myincA{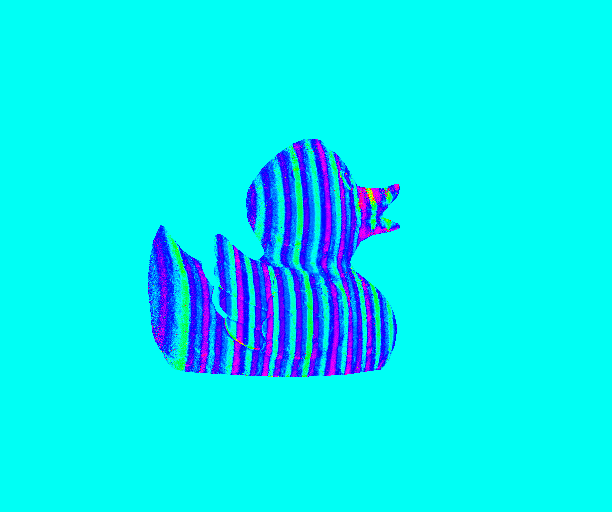}};
    \advance\vX by \vdX
    \node[inner sep=0pt] (a) at (\vX,\vY) {\myincA{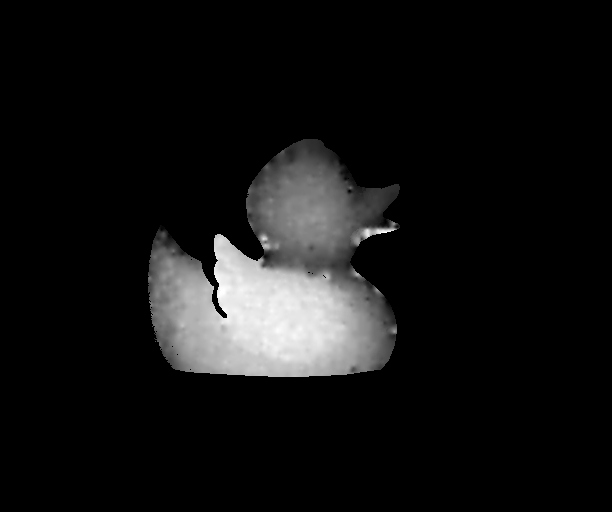}};
    \advance\vX by \vdX
    \node[inner sep=0pt] (a) at (\vX,\vY) {\myincA{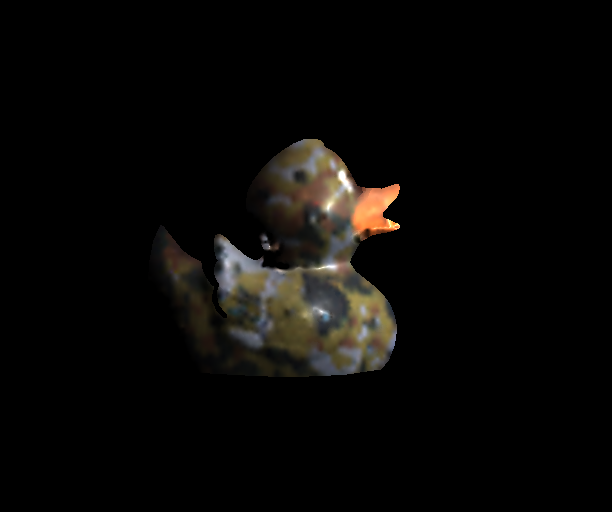}};
    
    \vdY = -190
    \vX = 0
    \advance\vY by \vdY
    \node[inner sep=0pt] (a) at (\vX,\vY) {\myincObjB{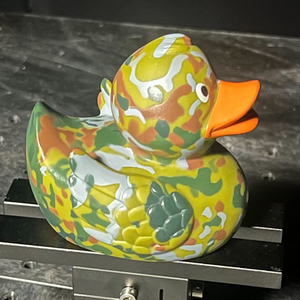}};
    \advance\vX by \vdX
    \node[inner sep=0pt] (a) at (\vX,\vY) {\myincA{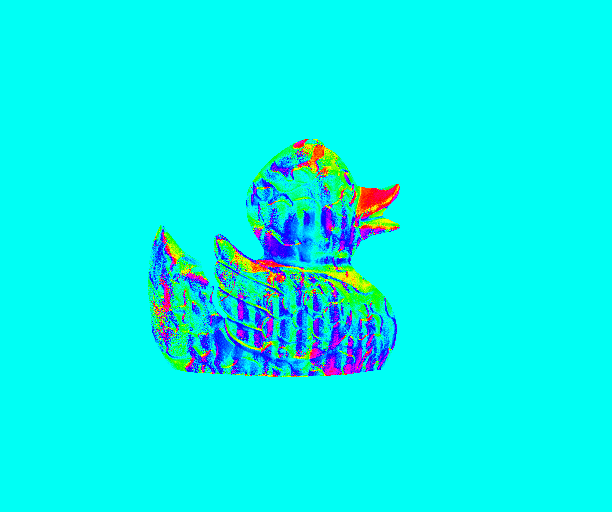}};
    \advance\vX by \vdX
    \node[inner sep=0pt] (a) at (\vX,\vY) {\myincA{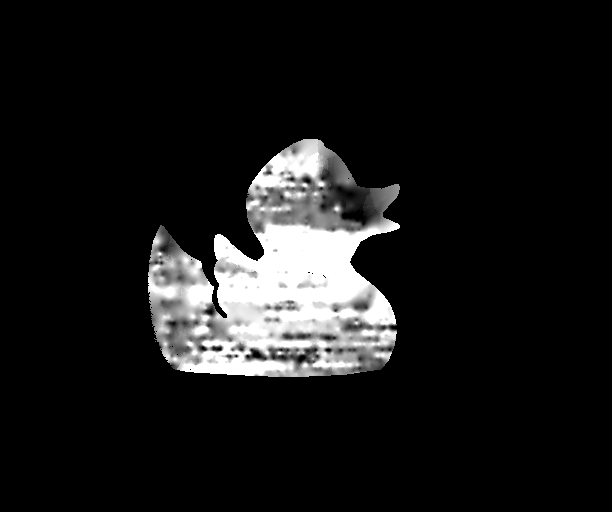}};
    \advance\vX by \vdX
    \node[inner sep=0pt] (a) at (\vX,\vY) {\myincA{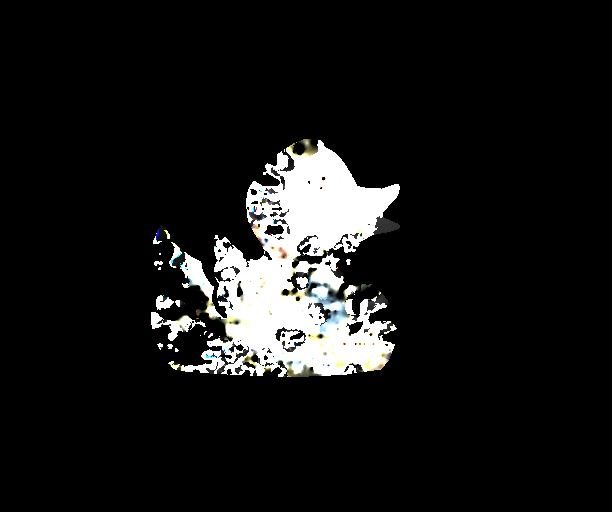}};

}
\end{tikzpicture}
}
    \vspace{-1.0em}
    \caption{Observed AoLP and reconstruction results under different ambient light. The leftmost image is a reference image to show the intensity of ambient light compared with the projector light projected onto the left side of the object.}
    \vspace{-0.5em}
    \label{fig: ambient light result}
\end{figure}

\begin{figure*}[t]
    \centering
    \input{fig_tex_supp/additional_shape_result}
    \caption{Additional shape reconstruction results. The numbers below each map indicate the mean and median of depth errors in millimeters and angular errors in degrees.}
    \label{fig: additional shape RGB}
\end{figure*}

\section{Constrained de Bruijn Sequence}

The original de Bruijn sequence is generated as the Eulerian path of the oriented graph whose nodes represent all possible strings of length $\ndb-1$ and edges represent the substrings of length $\ndb$~\cite{fredricksen1982survey}.
In this graph, each edge connects two nodes whose strings partially overlap to represent a substring of length $\ndb$.
Since we can concatenate one of $\kdb$ symbols to the front or back of each node to make substrings, the in-degree and out-degree of each node are $\kdb$, which indicates this graph has an Eulerian path.

As mentioned in the main paper, we impose two constraints for the de Bruijn sequence.
First, AoLP values of three sequential elements must be different for polarimetric decomposition.
Second, the differences between AoLP values of adjacent elements must be more than one quantization step for robust stripe detection.
By denoting $\kdb$ symbols as $0, 1, \ldots, \kdb-1$ and monotonically assigning each quantized AoLP value to the symbols, our two constraints on the de Bruijn sequence become $p_j \neq p_{j-2}, p_{j-1}, p_{j-1}\pm 1$ for all $j$, where $p_j$ is the symbol of the $j$-th element.

As shown in \cref{fig: de Bruijn graph}, we modify the oriented graph by removing the nodes and edges representing the substrings that violate our constraints.
When $\ndb \ge 3$ and $\kdb \ge 5$, we can concatenate one of the $\kdb-4$ symbols to the front or back of each remaining node.
Since the in-degree and out-degree of each node are $\kdb - 4$ in the modified graph, it has an Eulerian path.
The graph has $\kdb(\kdb-3)(\kdb-4)^{\ndb-3}$ nodes.
The number of length-$\ndb$ substrings in the generated sequence is equal to the number of edges, $\kdb(\kdb-3)(\kdb-4)^{\ndb-2}$.
We find the Eulerian path of the modified graph by Hierholzer's algorithm to generate the constrained de Bruijn sequence.
Our polarization pattern is constructed by assigning quantized AoLP values to the generated sequence of symbols.

\begin{figure*}[t]
    \centering
    \input{fig_tex_supp/additional_relighting_result}
    \caption{Additional relighting results. The Lambertian sphere at the bottom of the figure indicates the direction of light.}
    \label{fig: additional relighting RGB}
\end{figure*}

\section{AoLP Modulation by Diffuse Reflection}

Polarization of diffuse reflection is modeled with Fresnel transmission on the surface~\cite{Baek2018SimultaneousAO,ichikawa2023fresnel}.
In the left graph of \cref{fig:diffuse DoLP and AoLP modulation}, we plot the DoLP of diffuse reflection modeled as Fresnel transmission~\cite{Baek2018SimultaneousAO,ichikawa2023fresnel} for the zenith angle of the surface normal when the refractive index is $1.5$.
The diffuse DoLP $\rho^d_{L}$ is much lower than $1.0$ for all zenith angles.
In particular, when the zenith angle is less than $45^\circ$, the diffuse DoLP is $\rho^d_L < 0.05$.

In the right graph of \cref{fig:diffuse DoLP and AoLP modulation}, we show the AoLP modulation for $\frac{c^d}{c^s}$ 
when $\frac{\rho^d_L}{\rho^i_L} = 0.05$ and $\phi_L^i=0$.
Even when the diffuse reflection $c^d$ is dominant, the AoLP modulation is within only a few degrees.

\section{Shape Reconstruction with Occlusion}
\Cref{fig: occlusion} shows shape reconstruction results of scenes with occlusion. Decoding by dynamic programming can successfully handle the occlusion.
Occlusions for the camera are omitted as not-detected symbols during dynamic programming.
Occlusions for the projector are not detected as symbols or they are corrected as misdetections.

\section{Additional Experimental Results}

\Cref{fig: additional shape RGB,fig: additional relighting RGB} show additional experimental results of shape reconstruction and relighting on various objects of different shapes and materials.
Our method can accurately reconstruct both shape and reflectance from a single polarimetric image, compared with learning-based methods.

\section{Limitation}
\Cref{fig: ambient light result} experimentally shows the effect of ambient light on our method.
While we can reconstruct shape and reflectance under weak ambient light, strong ambient light compared with projector light affects the observed AoLP image, leading to failures in decoding.
A naive approach to handling ambient light is to use a more powerful light source.
Another approach is to exploit the modulated AoLP pattern.
Even when the ambient light is stronger than the projector light, a stripe AoLP pattern remains in the AoLP image.
Using this modulated polarization pattern as the guide for inverse rendering under natural illumination is one of our future directions.

Even though quantization makes detection of the projected AoLP robust to the AoLP modulation by diffuse reflection, there are two cases where the correct decoding is difficult. The first case is when the object surface shows very narrow specular highlights. In this case, $c^s=0$ on most of the surface regions and the reflected AoLP is largely modulated. The second case is when the zenith angle of the surface normal is close to the grazing angle, which causes large diffuse DoLP $\rho^d_L$.

Our method does not consider inter-reflection on the object surface. Inter-reflection disturbs the encoded polarization state of direct reflection. An object with a complex shape may cause errors in decoding and reflectance recovery due to inter-reflection.

\begin{figure}[t]
    \centering
    \resizebox{1.01\linewidth}{!}{
\begin{tikzpicture}[x=0.001\linewidth,y=0.001\linewidth,every node/.style={inner sep=0pt, text depth=0pt}]
{
    \newcount\vX
    \newcount\vY
    \newcount\vdX
    \newcount\vdXhalf
    \newcount\vdY
    \vX = 0
    \vY = 0
    \vdX = 164
    \def\mysz{0.166\linewidth}

    \def\myfont{\footnotesize}
    \node[inner sep=0pt, font=\myfont] (a) at (\vX,\vY) {Object};
    \advance\vX by \vdX
    \node[inner sep=0pt, font=\myfont] (a) at (\vX,\vY) {Image};
    \advance\vX by \vdX
    \node[inner sep=0pt, font=\myfont] (a) at (\vX,\vY) {AoLP};
    \advance\vX by \vdX
    \node[inner sep=0pt, font=\myfont] (a) at (\vX,\vY) {Object};
    \advance\vX by \vdX
    \node[inner sep=0pt, font=\myfont] (a) at (\vX,\vY) {Image};
    \advance\vX by \vdX
    \node[inner sep=0pt, font=\myfont] (a) at (\vX,\vY) {AoLP};

    \vdY = -110
    \vX = 0
    \advance\vY by \vdY
    \def\myincA#1{\resizebox{\mysz}{!}{\adjincludegraphics[Clip={{0.3\width} {0.15\width} {0.25\width} {0.275\width}}]{#1}}}
    \node[inner sep=0pt] (a) at (\vX,\vY) {\resizebox{\mysz}{!}{\myincA{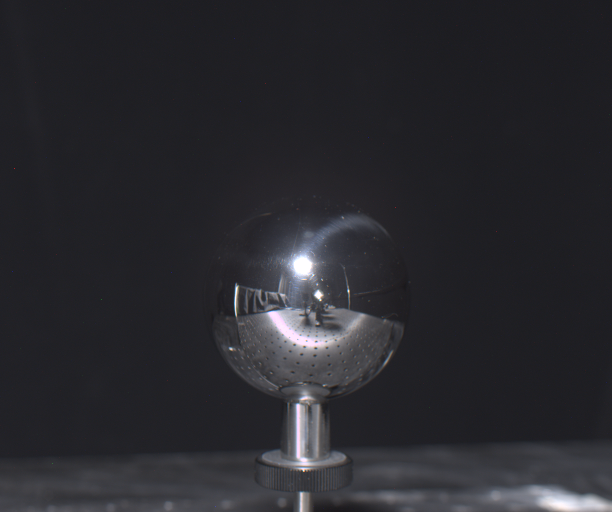}}};
    \advance\vX by \vdX
    \node[inner sep=0pt] (a) at (\vX,\vY) {\resizebox{\mysz}{!}{\myincA{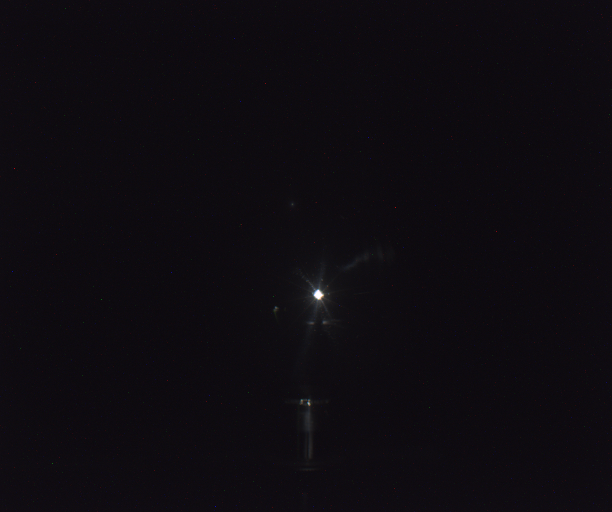}}};
    \advance\vX by \vdX
    \node[inner sep=0pt] (a) at (\vX,\vY) {\resizebox{\mysz}{!}{\myincA{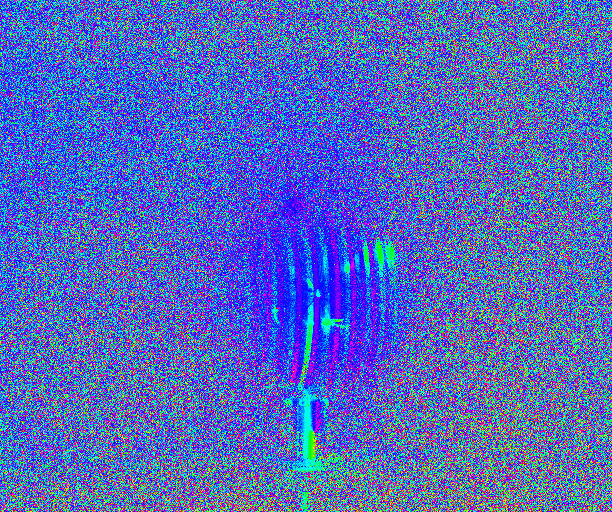}}};
    \advance\vX by \vdX

    \def\myincA#1{\resizebox{\mysz}{!}{\adjincludegraphics[Clip={{0.15\width} {0.15\width} {0.1\width} {0.0\width}}]{#1}}}
    \node[inner sep=0pt] (a) at (\vX,\vY) {\resizebox{\mysz}{!}{\myincA{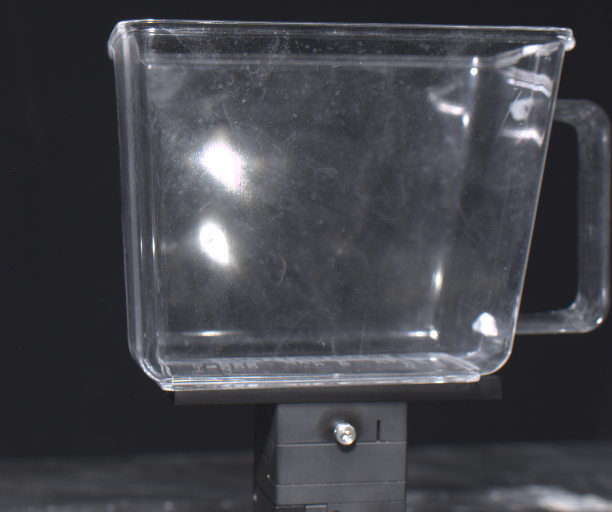}}};
    \advance\vX by \vdX
    \node[inner sep=0pt] (a) at (\vX,\vY) {\resizebox{\mysz}{!}{\myincA{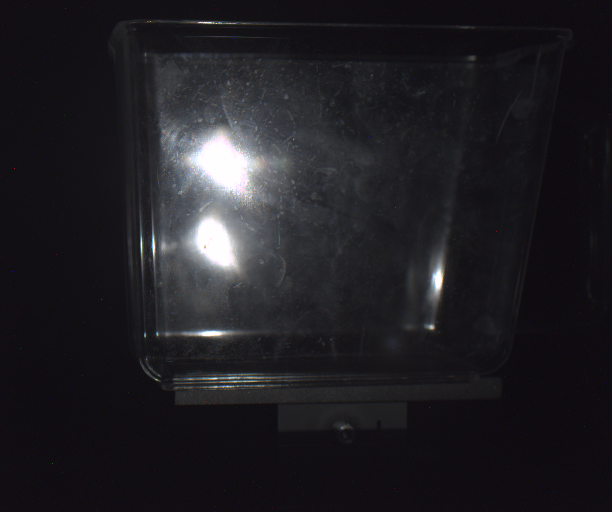}}};
    \advance\vX by \vdX
    \node[inner sep=0pt] (a) at (\vX,\vY) {\resizebox{\mysz}{!}{\myincA{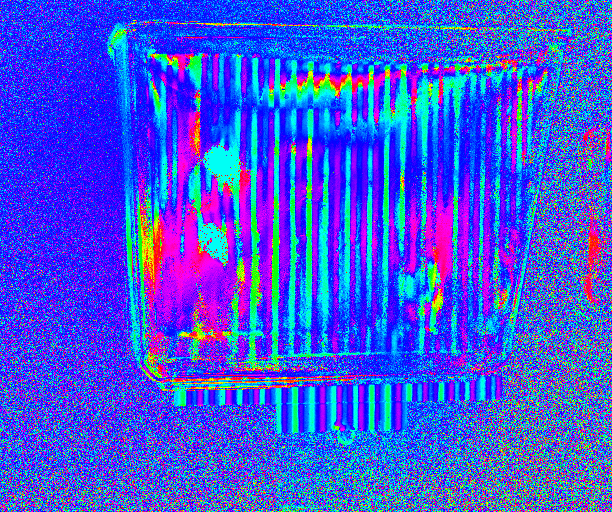}}};

}
\end{tikzpicture}
}
    \vspace{-1.0em}
    \caption{Highly specular and transparent materials can be challenging for SPM reconstruction. The reflected polarization pattern is too noisy in the large regions.}
    \vspace{-0.5em}
    \label{fig: specular limitation}
\end{figure}

Similar to many structured light methods, highly specular surfaces and transparent objects cause challenges. As shown in \cref{fig: specular limitation}, the reflected light on a highly specular surface is observed only in a small region with a specific surface normal, which prevents us from observing the reflected polarization pattern. Transparent surface shows a distorted polarization pattern due to transmitted light after inter-reflection within the object.

\begin{figure}[t]
    \centering
    \resizebox{1.01\linewidth}{!}{
\begin{tikzpicture}[x=0.001\linewidth,y=0.001\linewidth,every node/.style={inner sep=0pt, text depth=0pt}]
{
    \newcount\vX
    \newcount\vY
    \newcount\vdX
    \newcount\vdY
    \vX = 0
    \vY = 0
    \vdX = 197
    \def\mysz{0.199\linewidth}

    \newcount\vdXerror
    \newcount\vdYerror
    \def\myerror#1{\footnotesize \textcolor{white}{#1}}
    \def\showerror#1{
        \advance\vX by \vdXerror
        \advance\vY by \vdYerror
        \node[inner sep=0pt] (a) at (\vX,\vY) {\myerror{#1}};
        \advance\vX by -\vdXerror
        \advance\vY by -\vdYerror
    }

    \def\myfont{\footnotesize}
    \node[inner sep=0pt, font=\myfont] (a) at (\vX,\vY) {Image};
    \advance\vX by \vdX
    \node[inner sep=0pt, font=\myfont] (a) at (\vX,\vY) {AoLP};
    \advance\vX by \vdX
    \node[inner sep=0pt, font=\myfont] (a) at (\vX,\vY) {GT depth};
    \advance\vX by \vdX
    \node[inner sep=0pt, font=\myfont] (a) at (\vX,\vY) {Depth};
    \advance\vX by \vdX
    \node[inner sep=0pt, font=\myfont] (a) at (\vX,\vY) {Relighting};

    \vdY = -120
    \vX = 0
    \vdXerror = 0
    \vdYerror = - 62

    \advance\vY by \vdY
    \def\myincA#1{\resizebox{\mysz}{!}{\adjincludegraphics[Clip={{0.35\width} {0.32\width} {0.25\width} {0.15\width}}]{#1}}}

    \node[inner sep=0pt] (a) at (\vX,\vY) {\myincA{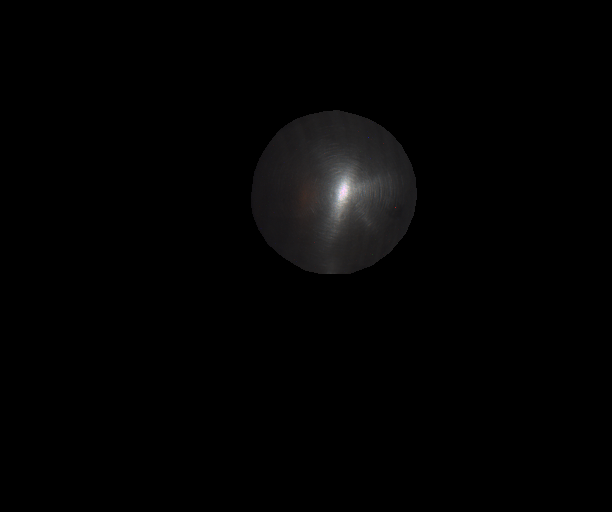}};
    \advance\vX by \vdX
    \node[inner sep=0pt] (a) at (\vX,\vY) {\myincA{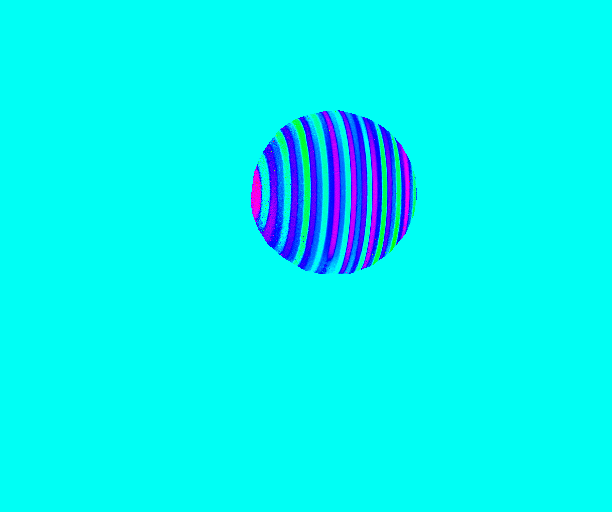}};
    \advance\vX by \vdX
    \node[inner sep=0pt] (a) at (\vX,\vY) {\myincA{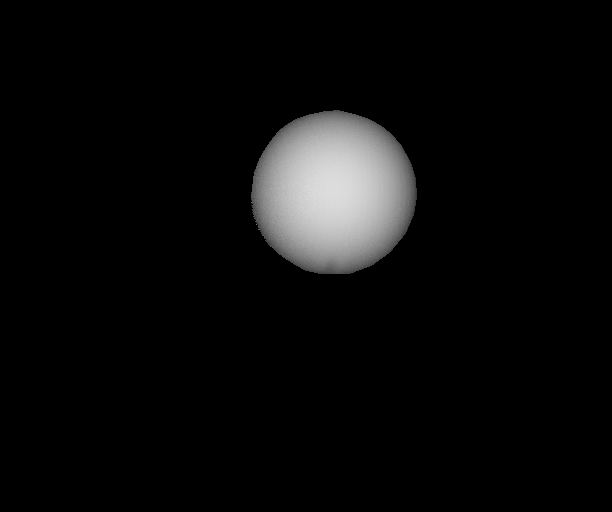}};
    \advance\vX by \vdX
    \node[inner sep=0pt] (a) at (\vX,\vY){\myincA{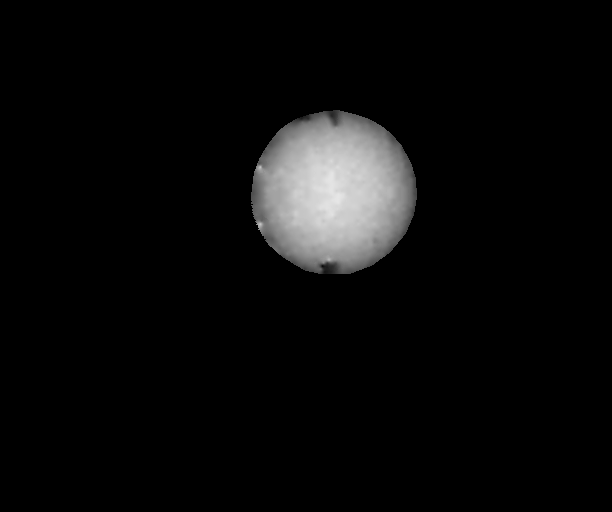}};
    \showerror{0.75/0.51}
    \advance\vX by \vdX
    \node[inner sep=0pt] (a) at (\vX,\vY){\myincA{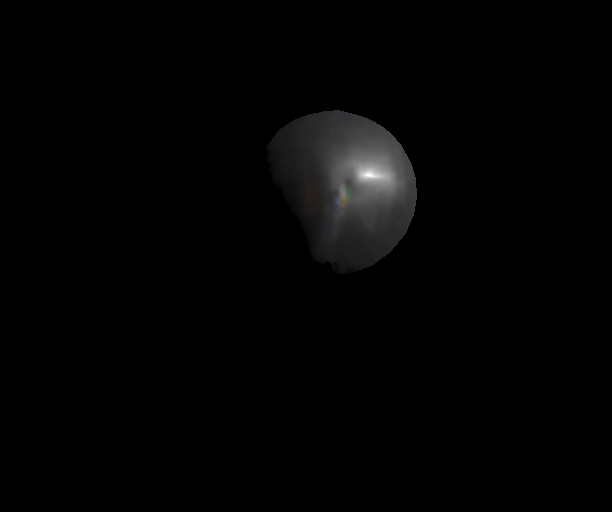}};

}
\end{tikzpicture}
}
    \vspace{-1.0em}
    \caption{Reconstruction results of a metallic surface. The reconstructed reflectance cannot represent photometric properties on a metallic surface different from a dielectric surface, such as color shift~\cite{cook1982reflectance}.}
    \vspace{-0.5em}
    \label{fig: metal result}
\end{figure}

Since the FMBRDF model~\cite{ichikawa2023fresnel} assumes a dielectric surface, our BRDF reconstruction method is limited to such materials.
\Cref{fig: metal result} shows shape reconstruction and relighting results of a rough metallic surface. Since the AoLP pattern is retained before and after reflection, our method can decode this pattern correctly and reconstruct the object shape.
Although relighting results capture the surface roughness, the reconstructed reflectance cannot correctly represent color shift on the metallic surface~\cite{cook1982reflectance}.
The metallic surface also shows a different polarimetric behavior from a dielectric surface. Phase delay on the metallic surface causes circular polarization~\cite{baek2020image}.
Combining a polarimetric reflectance model of a conductor~\cite{li2025neisf++} with our SPM to analyze these radiometric and polarimetric behaviors will allow for reflectance recovery of metallic surfaces in the future.

\begin{figure}[t]
    \centering
    \resizebox{1.01\linewidth}{!}{
\begin{tikzpicture}[x=0.001\linewidth,y=0.001\linewidth,every node/.style={inner sep=0pt, text depth=0pt}]
{
    \newcount\vX
    \newcount\vY
    \newcount\vdX
    \newcount\vdXhalf
    \newcount\vdY
    \vX = 0
    \vY = 0
    \vdX = 325
    \vdXhalf = 163
    \def\mysz{0.330\linewidth}

    \def\myfont{\footnotesize}
    \node[inner sep=0pt, font=\myfont] (a) at (\vX,\vY) {AoLP};
    \advance\vX by \vdX
    \node[inner sep=0pt, font=\myfont] (a) at (\vX,\vY) {Blurred AoLP};
    \advance\vX by \vdX
    \node[inner sep=0pt, font=\myfont] (a) at (\vX,\vY) {Focused AoLP};

    \vdY = -160
    \vX = 0
    \advance\vY by \vdY
    \def\myincA#1{\resizebox{\mysz}{!}{\adjincludegraphics[Clip={{0.08\width} {0.1\width} {0.03\width} {0.05\width}}]{#1}}}
    \node[inner sep=0pt] (a) at (\vX,\vY) {\resizebox{\mysz}{!}{\myincA{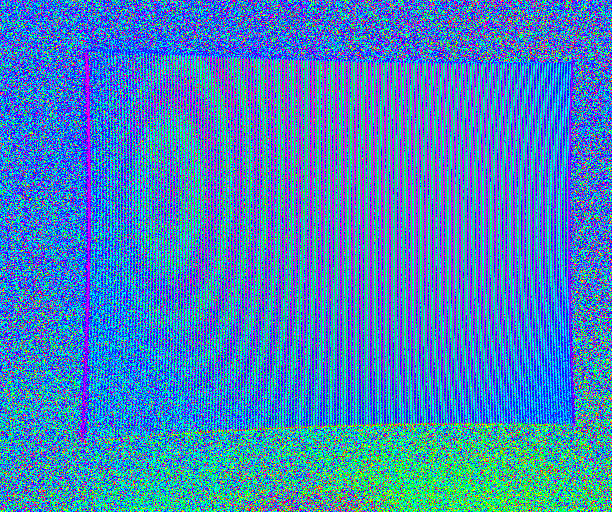}}};
    \advance\vX by \vdX

    \def\myincA#1{\resizebox{\mysz}{!}{\adjincludegraphics[Clip={{0.01\width} {0.1\width} {0.1\width} {0.05\width}}]{#1}}}
    \node[inner sep=0pt] (a) at (\vX,\vY) {\resizebox{\mysz}{!}{\myincA{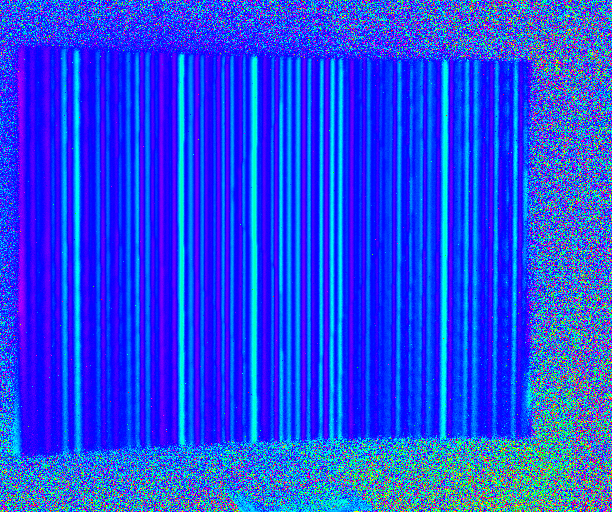}}};
    \advance\vX by \vdX

    \def\myincA#1{\resizebox{\mysz}{!}{\adjincludegraphics[Clip={{0.08\width} {0.1\width} {0.03\width} {0.05\width}}]{#1}}}
    \node[inner sep=0pt] (a) at (\vX,\vY) {\resizebox{\mysz}{!}{\myincA{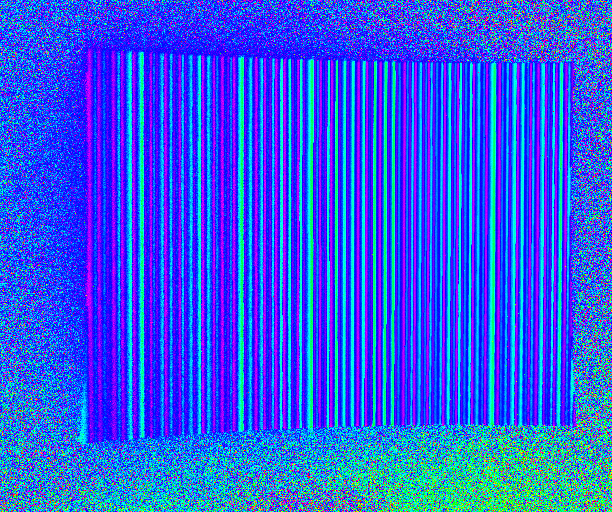}}};

    \vdY = -160
    \vX = 0
    \advance\vY by \vdY
    \def\myfont{\footnotesize}
    \node[inner sep=0pt, font=\myfont] (a) at (\vX,\vY) {Aliasing};
    \advance\vX by \vdX
    \advance\vX by \vdXhalf
    \node[inner sep=0pt, font=\myfont] (a) at (\vX,\vY) {Blur};

}
\end{tikzpicture}
}
    \vspace{-1.0em}
    \caption{The limitation of reconstruction resolution. Narrow stripe width causes aliasing and is susceptible to blur of the polarization projector.}
    \vspace{-0.5em}
    \label{fig: resolution limitation}
\end{figure}

The resolution of single-shot reconstruction is limited by two factors. First, aliasing occurs between the polarization pattern and a sensor array of a polarimetric camera. To capture the correct polarization states, a set of pixels on the polarimetric camera with four on-chip polarization filters needs to observe the same polarization states. Second, the blur of the polarization projector changes incident polarization states at the edge of the stripe pattern. If the stripe width is narrower than the blur, the AoLP values of each line may be modulated. \Cref{fig: resolution limitation} shows the effect of aliasing and blur. To avoid these effects, the stripe width must be sufficiently wide. We empirically set 12 pixels as the width of the stripe for our current hardware.
This limited resolution of reconstruction loses a detailed surface texture.
Advances in hardware (\eg, liquid crystal spatial light modulator) will increase the reconstruction resolution and mitigate this problem.

\end{document}